\documentclass[conference]{IEEEtran}

\usepackage{cite}
\usepackage{amsmath,amssymb,amsfonts}
\usepackage{algorithmic}
\usepackage{graphicx}
\usepackage{textcomp}
\usepackage{xcolor}
\usepackage{comment}

\usepackage{multirow}
\usepackage{subcaption} 
\usepackage{booktabs}
\usepackage{url}
\def\BibTeX{{\rm B\kern-.05em{\sc i\kern-.025em b}\kern-.08em
    T\kern-.1667em\lower.7ex\hbox{E}\kern-.125emX}}
\begin{document}

\newcommand{\appintseg}{A}
\newcommand{\appoutsourcing}{B}
\newcommand{\apprelatedwork}{C}
\newcommand{\appeffciency}{D}
\newcommand{\appjointmask}{E}
\newcommand{\appimplementationdetails}{F}
\newcommand{\appresults}{G}
\newcommand{\appskipclick}{H}
\newcommand{\appparallelfpn}{I}
\newcommand{\appqualitative}{J}

\title{MMMS: Multi-Modal Multi-Surface Interactive Segmentation}


\author{

\IEEEauthorblockN{Robin Schön}
\IEEEauthorblockA{\textit{Fakultät für Angewandte Informatik} \\
\textit{University of Augsburg}\\
Augsburg, Germany \\
robin.schoen@uni-a.de}
\and
\IEEEauthorblockN{Julian Lorenz}
\IEEEauthorblockA{\textit{Fakultät für Angewandte Informatik} \\
\textit{University of Augsburg}\\
Augsburg, Germany \\
julian.lorenz@uni-a.de}
\and
\IEEEauthorblockN{Katja Ludwig}
\IEEEauthorblockA{\textit{Fakultät für Angewandte Informatik} \\
\textit{University of Augsburg}\\
Augsburg, Germany \\
katja.ludwig@uni-a.de}
\and
\IEEEauthorblockN{Daniel Kienle}
\IEEEauthorblockA{\textit{Fakultät für Angewandte Informatik} \\
\textit{University of Augsburg}\\
Augsburg, Germany \\
daniel.kienzle@uni-a.de}
\and
\IEEEauthorblockN{Rainer Lienhart}
\IEEEauthorblockA{\textit{Fakultät für Angewandte Informatik} \\
\textit{University of Augsburg}\\
Augsburg, Germany \\
rainer.lienhart@uni-a.de}

}

\maketitle

\begin{abstract}
In this paper, we present a method to interactively create segmentation masks on the basis of user clicks. 
We pay particular attention to the segmentation of multiple surfaces that are simultaneously present in the same image. 
Since these surfaces may be heavily entangled and adjacent, we also present a novel extended evaluation metric that accounts for the challenges of this scenario.
Additionally, the presented method is able to use multi-modal inputs to facilitate the segmentation task.
At the center of this method is a network architecture which takes as input an RGB image, a number of non-RGB modalities, an erroneous mask, and encoded clicks. Based on this input, the network predicts an improved segmentation mask.  
We design our architecture such that it adheres to two conditions: (1) The RGB backbone is only available as a black-box. (2) To reduce the response time, we want our model to integrate the interaction-specific information after the image feature extraction and the multi-modal fusion.
We refer to the overall task as \underline{m}ulti-\underline{m}odal \underline{m}ulti-\underline{s}urface interactive segmentation (MMMS). 
We are able to show the effectiveness of our multi-modal fusion strategy. Using additional modalities, our system reduces the NoC@90 by up to 1.28 clicks per surface on average on DeLiVER and up to 1.19 on MFNet. On top of this, we are able to show that our RGB-only baseline achieves competitive, and in some cases even superior performance when tested in a classical, single-mask interactive segmentation scenario. 
\end{abstract}


\section{Introduction}
Segmentation tasks constitute some of the most important tasks in computer vision. The most prominent are instance segmentation \cite{li2023mask} and semantic segmentation \cite{xie2021segformer, kienzle2024segformerpp}. 
To train a segmentation model, we need large amounts of annotated segmentation data. 

However, ground truth segmentation masks are hard to obtain. This led to the development of click-based interactive segmentation systems \cite{ritm2022, kirillov2023segment, chen2022focalclick}. Therein, the user places clicks on the image to indicate which object surface they want to segment. The system then combines these clicks with the image to produce a high-quality segmentation mask. 

\begin{figure}[t]
\centering
\setlength{\tabcolsep}{0.5pt}
\newcommand{\qwidth}{0.16} 
\begin{tabular}{ccc}

\includegraphics[width=\qwidth\textwidth]{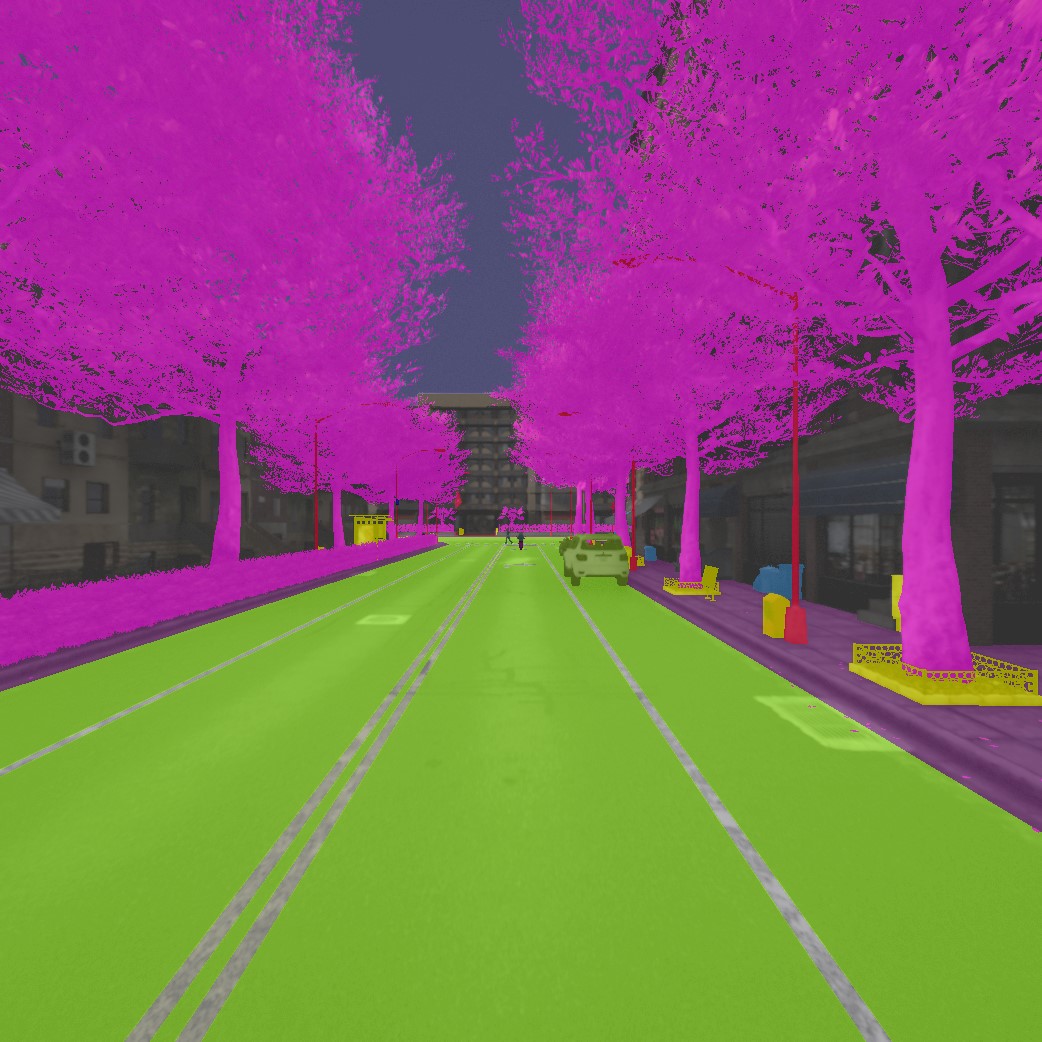} & 
\includegraphics[width=\qwidth\textwidth]{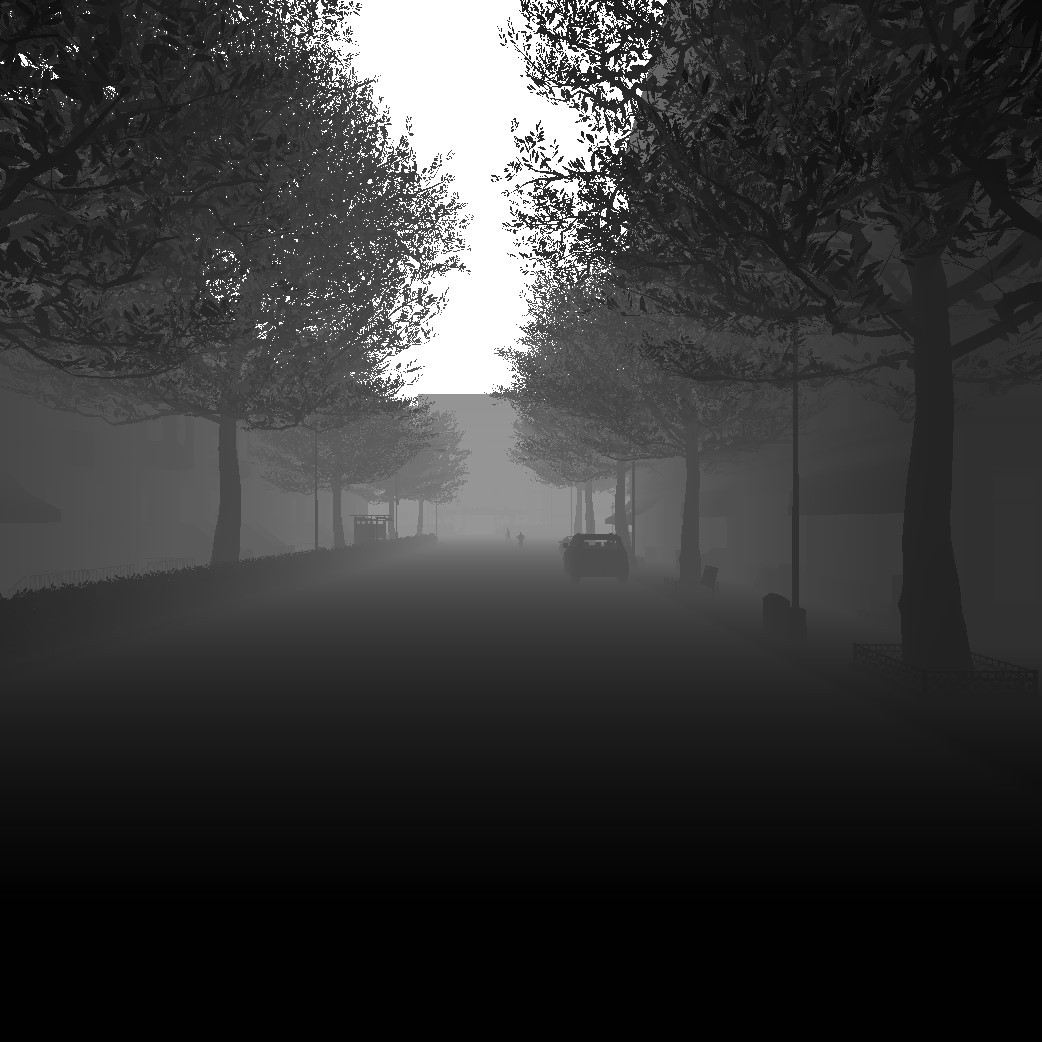} &    
\includegraphics[width=\qwidth\textwidth]{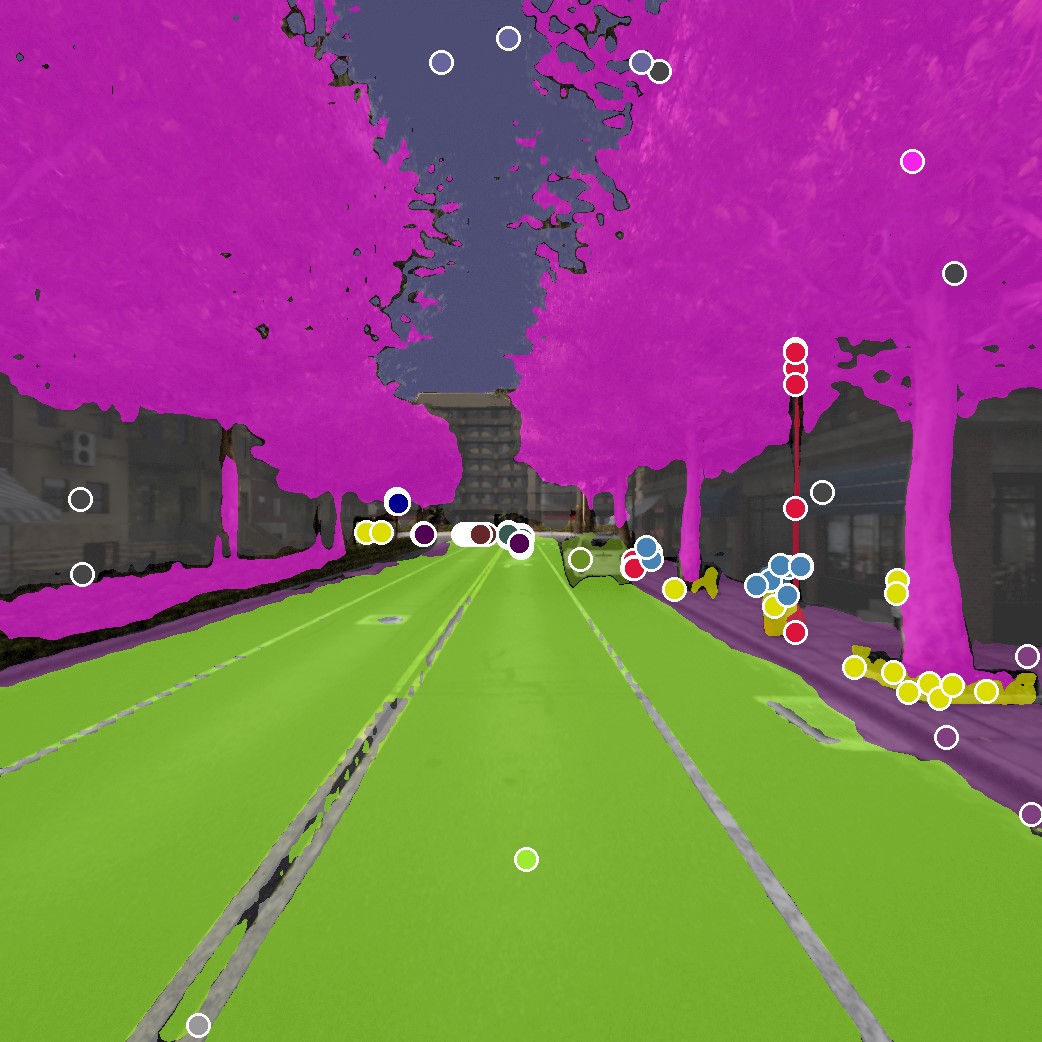}  \\

\includegraphics[width=\qwidth\textwidth]{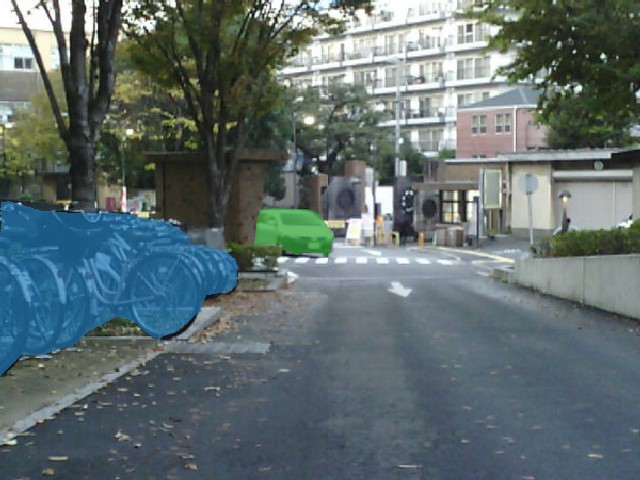} & 
\includegraphics[width=\qwidth\textwidth]{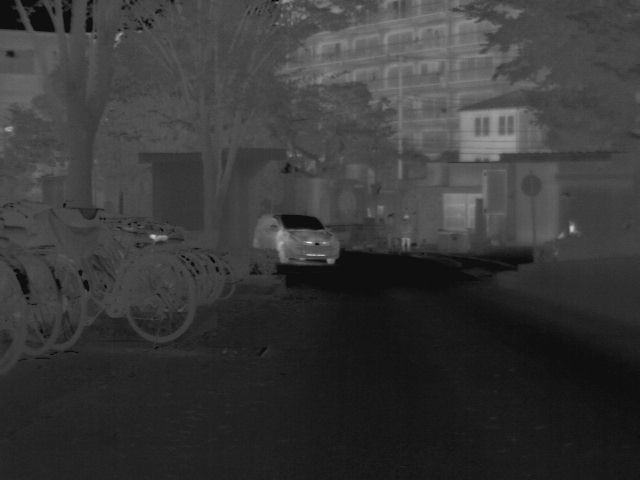} & 
\includegraphics[width=\qwidth\textwidth]{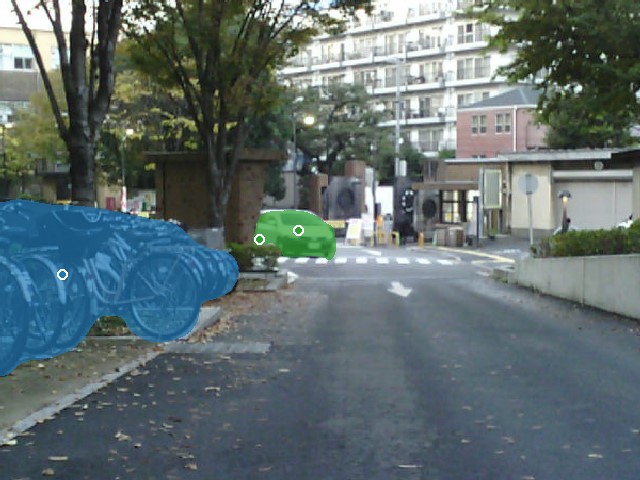}
\end{tabular}
\caption{Qualitative examples from DeLiVER (upper images) and MFNet (lower images). The left column shows the ground truth, the middle column shows the additional modality, and the right column shows the prediction. We only plot positive points for easier identification. More qualitative examples can be found in the supplementary material section \appqualitative{}. } 
\label{fig:qualitative_paper} 

\end{figure}

Most of the interactive segmentation literature only considers the segmentation process of each mask in isolation.
Nevertheless, there are situations in which we have multiple adjacent surfaces in the same image (e.g. Fig. \ref{fig:qualitative_paper}).

In such cases, false positive pixels in the annotations can cause conflicts between masks (see Fig. \ref{fig:revisiting}).  
A user then has to revisit some masks for correction. 
Previous evaluation procedures for interactive segmentation ignore this problem. 
For this reason, we propose a novel evaluation procedure that takes this problem into account.

When it comes to segmentation tasks, we often have the opportunity to benefit from additional modalities instead of just RGB images. For example, we might have depth maps or thermal images. These additional modalities are usually collected by a specific sensor at the same time as the images. 
Thus, by the time we annotate our images with segmentation masks, we have access to the non-RGB modalities. 
For this reason, we may leverage the other modalities to ease the annotation of our data. 
In this paper, we propose a network architecture that can leverage an arbitrary number of modalities for interactive segmentation.
We are able to show that that our multi-modal fusion strategy supports the interactive segmentation process.

On top of this, we design our multi-modal segmentation strategy in a way that allows for the RGB backbone to be a black-box. In particular, this allows us to use RGB foundation models which are outsourced to an external foundation model provider. This scenario is discussed in greater detail in section \appoutsourcing{} of the supplementary material.

In summary, we tackle the problem of \underline{m}ulti-\underline{m}odal \underline{m}ulti-\underline{s}urface interactive segmentation (MMMS). \footnote{We will release the code upon publication: \url{https://github.com/Schorob/mmms}}
Our contributions can be summarized as follows: 
\begin{itemize}
    \item We present an \textbf{asymmetric multi-modal fusion strategy} which assumes the RGB backbone to be an untrainable black-box while all other parts of the network are trainable. We show that our fusion strategy successfully leverages multi-modal data for interactive segmentation.
    \item We introduce \textbf{new metrics for interactive segmentation of multiple adjacent surfaces} in the same image. We adapt the evaluation mechanism to account for challenges in this scenario that have previously been ignored by evaluation mechanisms for classical interactive segmentation.
    \item We provide an experimental evaluation that demonstrates the \textbf{performances improvements caused by our strategy of multi-modal fusion} on various datasets with a wide range of non-RGB modalities. 
\end{itemize}

\section{Related Work}
\subsection{Interactive Segmentation} 
While there are other modes of interaction \cite{lowes2023interactive,chen2022focalclick,zhang2020interactive}, this paper only focuses on click-based interactive segmentation \cite{ritm2022,sofiiuk2020f}. A detailed introduction to this problem can be found in the supplementary material  section \appintseg{}. 
Most modern interactive segmentation systems are based on a neural network that takes the encoded clicks and the image as input, and tries to predict a high-quality mask \cite{ritm2022, Liu_2023_ICCV, chen2022focalclick, kirillov2023segment, ke2024segment}. As the task formulation revolves around iteratively improving a mask, recent systems \cite{ritm2022, Liu_2023_ICCV} use the previous, imperfect mask as additional input. 
The bulk of these works require rerunning the entire network after each click, incurring slow response times. Similar to previous works \cite{kirillov2023segment,huang2023interformer,liu2024rethinking,schoen2025skipclick}, we design our network such that only a small part of the architecture has to be rerun after each click. 
Notably, there is also previous work on interactive segmentation on images with multiple surfaces \cite{agustsson2019interactive,andriluka2020efficient,rana2023dynamite,li2024learning}. However, come to find their interaction modes rather arbitrary and constraining with respect to  architectural choices. In contrast to previous methods, we offer an extension of the standard interactive segmentation problem.

\subsection{Multi-Modal Segmentation}
Earlier work on multi-modal segmentation has mostly dealt with a single additional input modality. Examples for such architectures are FuseNet \cite{fusenet2016accv}, MFNet \cite{ha2017mfnet}, RTFNet \cite{sun2019rtfnet}, SegMiF \cite{liu2023segmif} and CMX \cite{zhang2023cmx}. 
More recent methods use multiple additional modalities at once. TokenFusion \cite{wang2022tokenfusion} combines tokens from a variable number of modalities. MCubeSNet \cite{Liang_2022_CVPR} concatenates the features from different modalities before feeding them to the decoder. 
CMNeXt \cite{zhang2023delivering} introduces a mechanism to select tokens from multiple non-RGB modalities, which are afterwards fused with the main RGB modality. Both, HRFuser \cite{broedermann2023hrfuser} and CAFuser \cite{broedermann2024cafuser} use windowed cross-attention to combine multiple modalities. 
GeminiFusion \cite{jia2024geminifusion} applies a pixel-wise fusion mechanism between the attention mechanism and the MLP of the backbone.
StitchFusion \cite{li2024stitchfusion} uses adapter modules which are inserted into a frozen backbone.
All of the aforementioned methods assume access to the RGB backbone during training. However, our method will allow for a black-box RGB backbone. 
We are not the first to use non-RGB information for the purpose of click-guided interactive segmentation. The methods presented in \cite{Schon_2023_CVPR} and \cite{wang2025orderawareinteractivesegmentation} are both based on using pseudo-depth maps generated by a pretrained monocular depth estimation (MDE) network. 
In contrast to these methods, we want to use real and arbitrary additional modalities.

\section{Multi-Surface Interactive Segmentation}
\label{sec:multi_surface_metrics}

In this section, we discuss an extension to the classical interactive segmentation problem. We adress challenges that occur when segmenting multiple masks in the same image. We recommend that any reader unfamiliar with the standard interactive segmentation problem and the NoC metric may first read section \appintseg{} in the supplementary material. The standard NoC metric considers each mask in isolation.

Suppose we want to interactively create segmentation masks for multiple surfaces in the same image using a regular interactive segmentation system. 
The most straightforward way of doing this would be to annotate these masks independently. Let $L$ be the number of surfaces / masks we want to annotate. 
We want to create a set of segmentation masks 
\begin{equation} 
\label{eq:mask_sequence}
    \mathcal{S}_\mathbf{m} = \left\{ \mathbf{m}^{(1)}, ..., \mathbf{m}^{(i)}, ..., \mathbf{m}^{(j)},  ..., \mathbf{m}^{(L)} \right\}, 
\end{equation}
where each of those masks is annotated using the standard interactive segmentation procedure. 
We generally do not assume the user to annotate each of these masks to absolute perfection. We just expect a sufficiently high degree of quality.
This is also reflected in the automatic evaluation of interactive segmentations systems. We only continue the annotation clicks until an IoU with a pre-existing grouth truth mask of at least $\Theta_\text{IoU}$ is reached. 
These imperfections allow for conflicts by creating overlap between adjacent masks. 

The standard version of the interactive segmentation problem ignores the issue of overlap between imperfect masks. It deals with each of these masks in isolation, despite each describing a different surface on the same image.
Given the fact that these surfaces are sometimes semantically defined (e.g. \emph{grass}, \emph{sky} or \emph{water}), they may actually be heavily entangled.

In the remainder of this section, we will describe an extended version of the interactive segmentation evaluation procedure that takes this problem into account. We will also describe corresponding metrics.

\begin{figure}
    \centering
    \includegraphics[width=\linewidth]{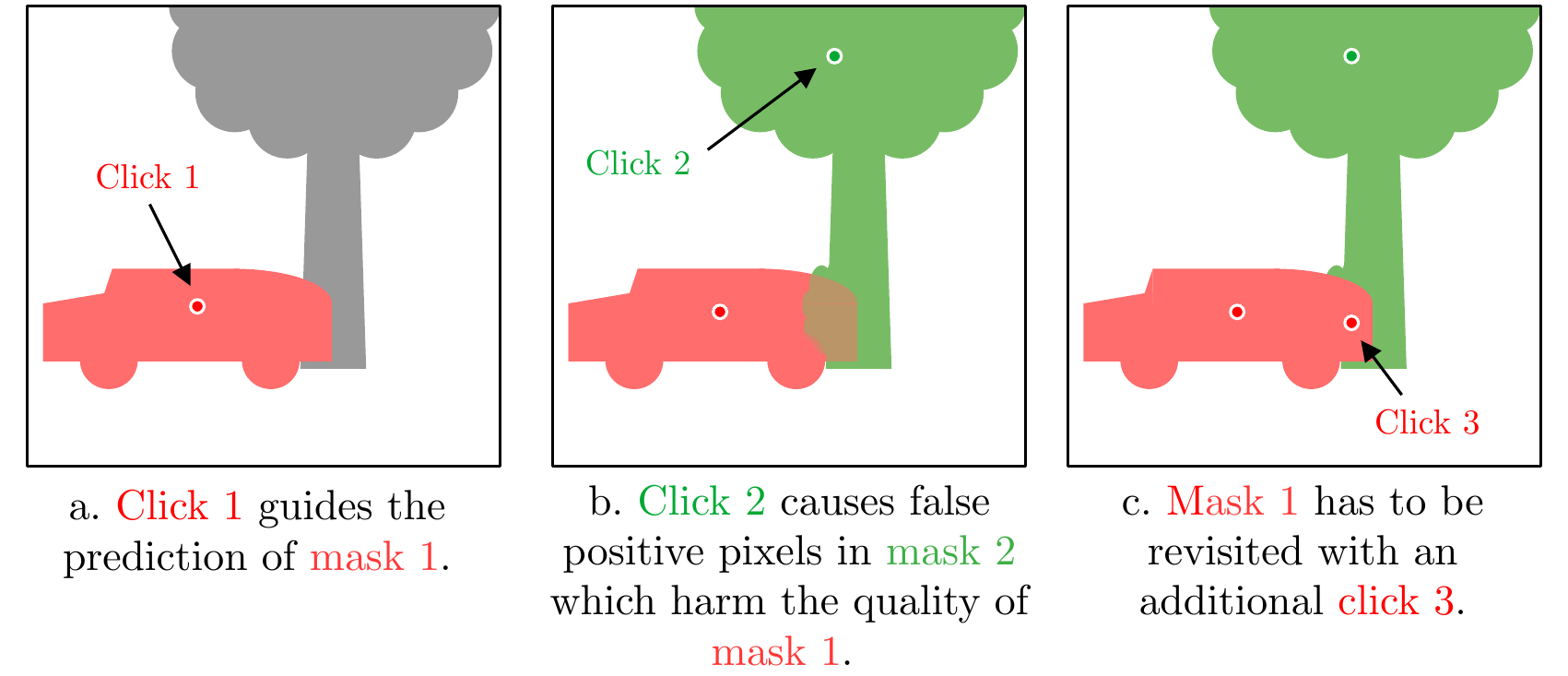}
    \caption{We first place click 1 to annotate mask 1 (a.). Afterwards, we place click 2 to annotate mask 2 (b.). False positive pixels in mask 2 harm the quality of mask 1. We then have to revisit mask 1 by placing click 3 (c.).} 

    \label{fig:revisiting}
\end{figure}

Let $\mathbf{m}^{(i)}, \mathbf{m}^{(j)} \in \mathcal{S}_\mathbf{m}$ be two masks from Eq. \ref{eq:mask_sequence}. Here, $i$ and $j$ indicate different target surfaces; not different iterations for the same target surface. Assume that $\mathbf{m}^{(j)}$ has been annotated after $\mathbf{m}^{(i)}$. In case of an overlap, the later annotated mask $\mathbf{m}^{(j)}$ overrides the earlier annotated mask $\mathbf{m}^{(i)}$. An example of such a situation can be found in Fig. \ref{fig:revisiting}. Some pixels in the overlapping area may have been false positives of $\mathbf{m}^{(j)}$ that actually belong to $\mathbf{m}^{(i)}$ (see Fig. \ref{fig:revisiting} b.). In such a case, the false positive pixels in $\mathbf{m}^{(j)}$ have harmed the quality of $\mathbf{m}^{(i)}$. We would need to revisit $\mathbf{m}^{(i)}$ (see in Fig. \ref{fig:revisiting} c.). 

In order to account for this challenge, we introduce two new metrics: \emph{NoCMS} and \emph{FRMS}. These metrics are based on two IoU-thresholds $\Theta_\text{Average-IoU}$ and $\Theta_\text{IoU}$. 
When computing these metrics, our goal is to reach a sufficiently high average IoU of all masks with their associated ground truth masks. This is the case if 
\begin{equation}
\label{eq:avg_iou}
   \frac{1}{L} \sum_{l=1}^L \text{IoU}(\mathbf{m}^{(l)}, \mathbf{m}^{(l)}_\text{GT}) \ge \Theta_\text{Average-IoU}. 
\end{equation}
If this holds true, we are done with annotating the image. If not, we pick the mask $\mathbf{m}^{(k)} \in \mathcal{S}_\mathbf{m}$ which has the smallest IoU with its corresponding ground truth. This $\mathbf{m}^{(k)}$ is then improved with the regular interactive segmentation mechanism for single surfaces. We add clicks until an IoU of at least $\Theta_\text{IoU}$ is reached. Afterwards, we test again whether the condition in Eq. \ref{eq:avg_iou} is true.
We repeat this cycle of \emph{selection and improvement} of an $\mathbf{m}^{(k)}$ until an average IoU of at least $\Theta_\text{Average-IoU}$ is reached. 
It should be noted that this evaluation requires $\Theta_\text{IoU} \ge \Theta_\text{Average-IoU}$. Otherwise, the improvements of $\mathbf{m}^{(k)}$ might not be significant enough for the average IoU to rise above $\Theta_\text{Average-IoU}$. 
The count of already executed clicks for each particular surface is \textbf{not} reset at each revisiting attempt. Instead, we accumulate the click count over different attempts with the same surface. 
Since the network might not be capable of annotating some masks, we use a maximum number of clicks $n_\text{max} = 20$ for each surface mask. 
We introduce two new metrics for \underline{m}ulti-\underline{s}urface interactive segmentation. The average number of clicks per surface in this extended scenario is called $\text{NoCMS}@(\Theta_\text{IoU}, \Theta_\text{Average-IoU})$. 
We can also measure the percentage of surfaces that could not be segmented successfully within $n_\text{max}=20$ clicks. We refer to this percentage as the failure rate $\text{FRMS}@(\Theta_\text{IoU}, \Theta_\text{Average-IoU})$. 
Additionally, section \appjointmask{} of our supplementary material discusses the creation of a joint mask for all surfaces in an image.

\section{Method\label{sec:architecture}}

\begin{figure*}
    \centering
    \includegraphics[width=0.97\linewidth]{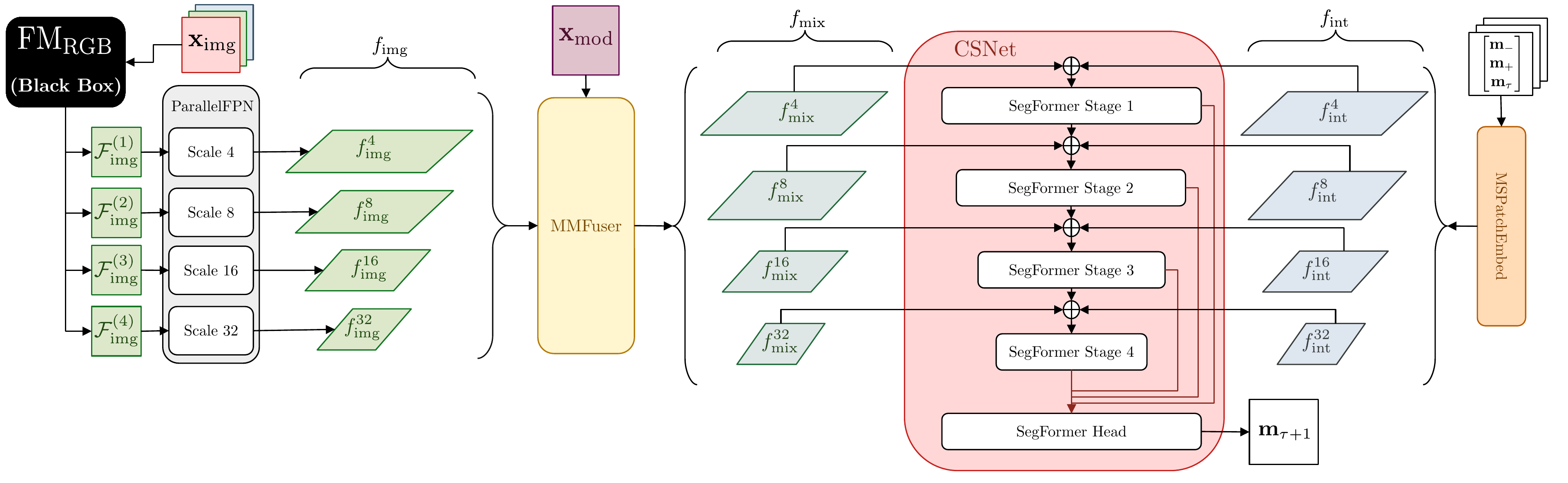}
    \caption{Overall architecture of our multi-modal interactive segmentation system. The RGB features $\mathcal{F}_\text{img}$ from the backbone $\text{FM}_\text{RGB}$ are processed by a ParallelFPN to obtain a feature pyramid $f_\text{img}$. The MMFuser integrates information from non-RGB modalities $\mathbf{x}_\text{mod}$ into the feature pyramid resulting in $f_\text{mix}$. MSPatchEmbed creates a feature pyramid $f_\text{int}$ from the interaction tensor $[\mathbf{m}_+;\mathbf{m}_-;\mathbf{m}_\tau]$. CSNet uses $f_\text{mix}$ and $f_\text{int}$ to predict the mask $\mathbf{m}_{\tau + 1}$. Apart from CSNet and the mutli-scale patch embedding (MSPatchEmbed) every part of our architecture only has to be executed once per image. $\oplus$ denotes element-wise addition.}
    \label{fig:architecture} 

\end{figure*}

In this section, we describe the architecture that we use to tackle our problem. 
We first explain the general architecture and its overall functionality. Afterwards, we explain the \emph{MMFuser} and the \emph{CSNet} in greater detail. An overview of our architecture is given in Fig. \ref{fig:architecture}.

\paragraph{Overall architecture }

We will describe the data processing in our architecture in a step-by-step fashion. 
First, the RGB image $\mathbf{x}_\text{img} \in \mathbb{R}^{H \times W \times 3}$ is processed by the RGB backbone $\text{FM}_\text{RGB}$. 
We restrict ourselves to cases where $\text{FM}_\text{RGB}$ has a ViT architecture. Let $P_\text{FM}$ and $d_\text{FM}$ be the patch size and the embedding dimension of the model, respectively. We obtain the image feature tensors 
\begin{equation}
    \mathcal{F}_\text{img} =  \left( \mathcal{F}^{(1)}_\text{img}, \mathcal{F}^{(2)}_\text{img}, \mathcal{F}^{(3)}_\text{img}, \mathcal{F}^{(4)}_\text{img} \right) = \text{FM}_\text{RGB}(\mathbf{x}_\text{img}).
\end{equation}
These image feature tensors $\mathcal{F}^{(i)}_\text{img} \in \mathbb{R}^{\frac{H}{P_\text{FM}} \times \frac{W}{P_\text{FM}} \times d_\text{FM}}$ for $i=1, 2, 3, 4$ are extracted from various intermediate layers in the network. In our case, the features are extracted after the blocks 3, 6, 9, and 12.
As $\text{FM}_\text{RGB}$ is a black-box, we still need a learnable adapter that gives our model the ability to change the representation of $\mathcal{F}_\text{img}$. We follow common practice and operate on feature pyramids to better represent multi-scale features. 
In order to obtain an adapted feature pyramid, we use a \emph{ParallelFPN} to transform $\mathcal{F}_\text{img}$ into  
\begin{equation}
    f_\text{img} = \left( f^{4}_\text{img}, f^{8}_\text{img}, f^{16}_\text{img}, f^{32}_\text{img} \right) = \text{ParallelFPN}(\mathcal{F}_\text{img}). 
\end{equation}
We have $f^i_\text{img} \in \mathbb{R}^{\frac{H}{i} \times \frac{W}{i} \times C_i}$ for $i \in \{4, 8, 16, 32\}$. The $C_i$ are the feature channel numbers of the respective tensors in our feature pyramid. Unless stated otherwise, all feature pyramids in the following text will have this shape. 
Our ParallelFPN is inspired by \cite{Liu_2023_ICCV}. A detailed description can be found in the supplementary material in section \appparallelfpn{}.

So far, we have not integrated any non-RGB information into our features. This will be done by the \emph{MMFuser}. The MMFuser receives $M$ non-RGB images $\mathbf{x}_{\text{mod}, m} \in \mathbb{R}^{H \times W \times C_m}$ for $m = 1, ..., M$ 
as well as the RGB features $f_\text{img}$ as input. 
This results in the mixed feature pyramid 
\begin{equation}
\begin{split}
    f_\text{mix} = \left( f^{4}_\text{mix}, f^{8}_\text{mix}, f^{16}_\text{mix}, f^{32}_\text{mix} \right) = \\
    \text{MMFuser}(f_\text{img}, \mathbf{x}_{\text{mod}, 1}, ..., , \mathbf{x}_{\text{mod}, M}). 
\end{split}
\end{equation}
$f_\text{mix}$ has the same tensor shapes as $f_\text{img}$. It should be noted that \emph{all computation up until this point has to be carried out only once per image}. The feature extraction is not dependent on information for interactive mask annotation. 
Due to this design choice, we do not have to rerun $\text{FM}_\text{RGB}$ and MMFuser again after each click, which improves the response time. 
The positive and negative clicks are encoded as small disks in their respective binary maps $\mathbf{m}_+$ and $\mathbf{m}_-$ as described in \cite{ritm2022}. The clicks are then concatenated with the previous mask $\mathbf{m}_\tau$, resulting in the \emph{interaction tensor} $[\mathbf{m}_+;\mathbf{m}_-;\mathbf{m}_\tau] \in \mathbb{R}^{H \times W \times 3}$. 
We want to give our model access to the interaction information at every stage of the mask prediction.
For this reason, we make use of a multi-scale patch embedding \emph{MSPatchEmbed}. MSPatchEmbed converts the interaction tensor into an interaction feature pyramid 
\begin{equation}
\begin{split}
    f_\text{int} &= \left( f^{4}_\text{int}, f^{8}_\text{int}, f^{16}_\text{int}, f^{32}_\text{int} \right) \\
    &= \text{MSPatchEmbed}([\mathbf{m}_+;\mathbf{m}_-;\mathbf{m}_\tau]). 
\end{split}
\end{equation}
The tensors in $f_\text{int}$ have the same shapes as those in $f_\text{mix}$. \linebreak 
MSPatchEmbed is implemented as different convolutions that produce the desired shapes in our feature pyramid.

The last part of our architecture is our click segmentation network \emph{CSNet}, which can be seen on the right side of Fig. \ref{fig:architecture}. CSNet combines the features in $f_\text{mix}$ and $f_\text{int}$, and produces the improved mask $\mathbf{m}_{\tau+1}$: 
\begin{equation}
    \mathbf{m}_{\tau + 1} = \text{CSNet}(f_\text{mix}, f_\text{int})
\end{equation}
We note that the two modules MSPatchEmbed and CSNet are the only two components of our network that would have to be rerun after each click. 

\begin{figure}
    \centering
    \begin{tabular}{cc}
         \includegraphics[width=0.57\linewidth]{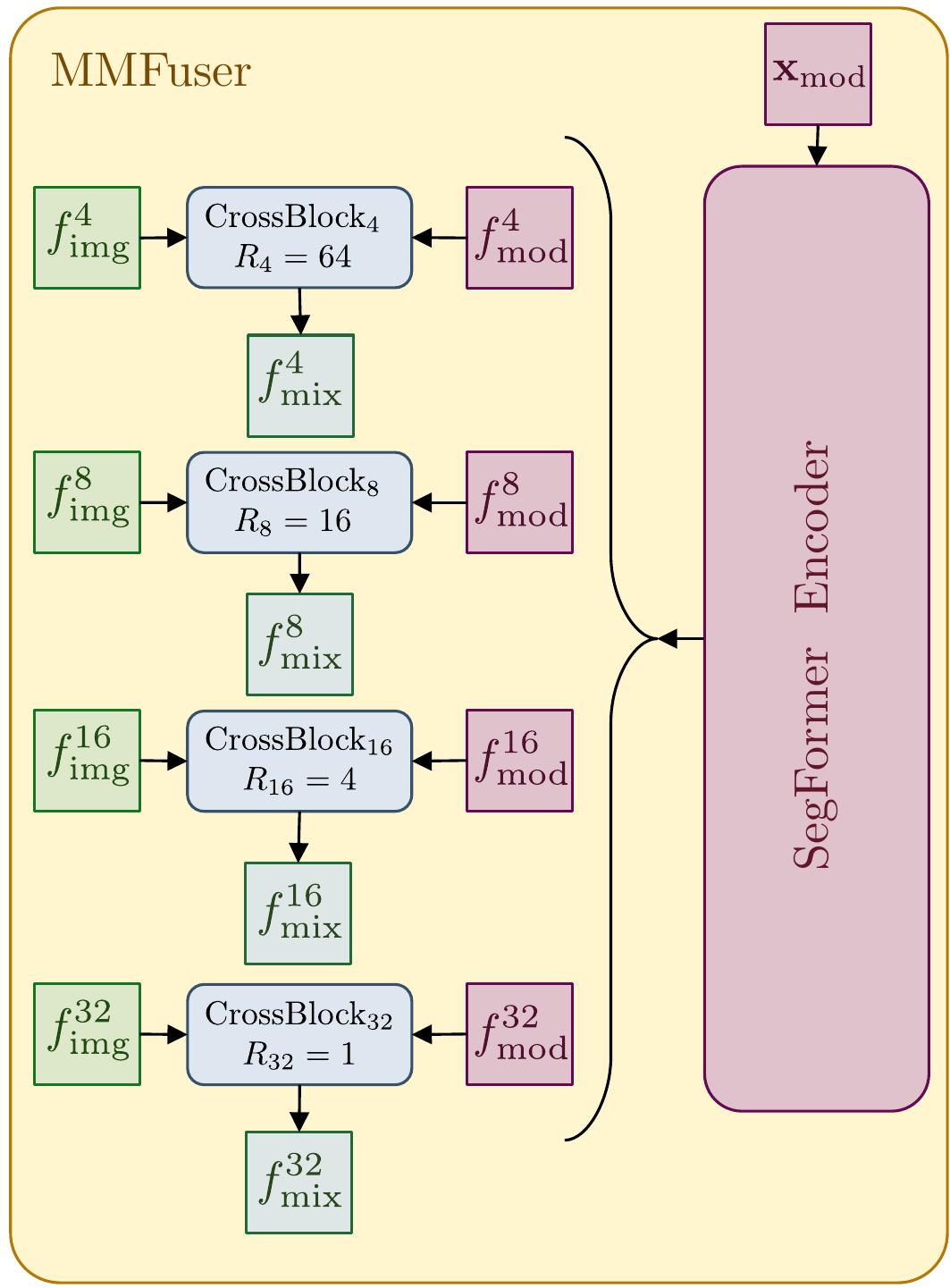} & \includegraphics[width=0.33\linewidth]{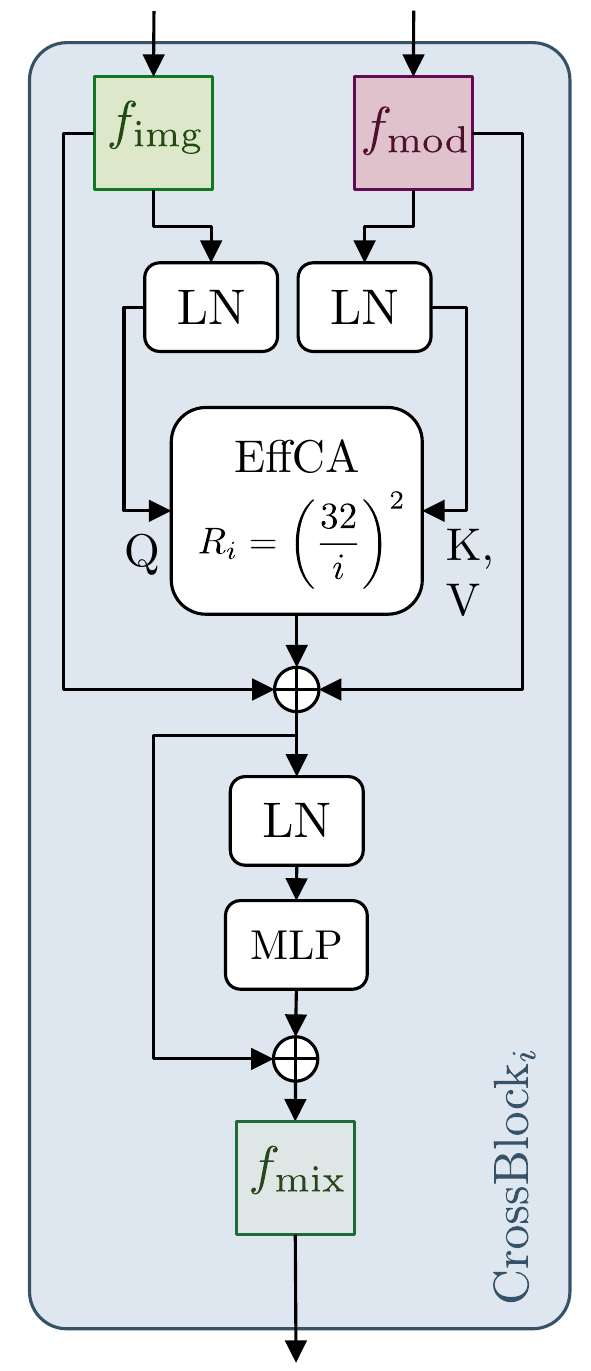} 
    \end{tabular}
    
    \caption{The MMFuser (\emph{left}) uses a SegFormer encoder to compute the features for the non-RGB modality. The feature fusion happens in the CrossBlock (\emph{right}). Similar to \cite{xie2021segformer}, the efficient cross-attention (EffCA) works with a number of keys and values that are reduced by a reduction rate $R \in \{1, 4, 16, 64\}$. In case we use multiple non-RGB modalities, we have one SegFormer encoder and one CrossBlock per modality. $\oplus$ denotes element-wise addition. } 

    \label{fig:mmfuser_crossblock}
\end{figure}

\paragraph{MMFuser }
We now describe our MMFuser. The MMFuser is used to integrate multi-modal information into the feature pyramid (Fig. \ref{fig:mmfuser_crossblock}, \emph{left}). 

First, the non-RGB modality $\mathbf{x}_\text{mod}$ is processed by a SegFormer encoder \cite{xie2021segformer}. This encoder yields a feature pyramid 
\begin{equation}
\begin{split}
    f_\text{mod} &= \left( f^{4}_\text{mod}, f^{8}_\text{mod}, f^{16}_\text{mod}, f^{32}_\text{mod} \right) \\ 
    &= \text{SegFormerEncoder}(\mathbf{x}_\text{mod}). 
\end{split}
\end{equation}
The tensors in $f_\text{mod}$ have the same shape as those in $f_\text{img}$. 
In order to mix $f_\text{img}$ and $f_\text{mod}$, we will employ a cross-attention based \emph{CrossBlock} (Fig. \ref{fig:mmfuser_crossblock}, \emph{right}) at each stage. For each $i \in \{4, 8, 16, 32\}$, we compute  
\begin{equation}
    f^i_\text{mix} = \text{CrossBlock}_i(f_\text{img}^i, f_\text{mod}^i). 
\end{equation}
We want to equip the model with the capacity to dynamically integrate useful parts from other modalities. For this reason, the first part of the CrossBlock is a cross-attention module.
To reduce the execution time, we extend the efficient self-attention of \cite{xie2021segformer} to cross-attention. We reduce the number of keys and values by a reduction rate $R_i$. The number of queries stays the same. This results in what we call \emph{efficient cross-attention EffCA}. 
Therein, the image features $f^i_\text{img}$ are used to compute the queries Q, while we compute the keys $K$ and values $V$ on the basis of $f^i_\text{mod}$. For a particular $i \in \{4, 8, 16, 32\}$ we will use a reduction rate $R_i = \left( \frac{32}{i} \right)^2$. 
Formally we have 
\begin{align}
    \hat{f}^i_\text{mix} &= \text{EffCA}(\text{LN}(f^i_\text{img}), \text{LN}(f^i_\text{mod})) + f^i_\text{img} + f^i_\text{mod} \\
    f^i_\text{mix} &= \text{MLP}(\text{LN}(\hat{f}^i_\text{mix})) + \hat{f}^i_\text{mix}, 
\end{align}
where LN represents layer normalizations. Each layer normalization has its own learnable parameters. 
In cases where we have more than one non-RGB modality, we use a separate SegFormer encoder \cite{xie2021segformer} for each of them. In addition to this, we have multiple cross-attentions, one for each modality. 

\paragraph{CSNet }
The click segmentation network \emph{CSNet} uses the interaction information $f_\text{int}$ and our feature pyramid $f_\text{mix}$ to predict a mask $\mathbf{m}_{\tau+1}$. 
Both, $f_\text{int}$ and $f_\text{mix}$ are encoded in a dense fashion.
We thus deem it natural to use existing insights about processing dense features, and employ an existing segmentation architecture. 
In our case, we will use the different stages and the prediction head of a SegFormer \cite{xie2021segformer}. The internal mechanism is depicted on the right side of Fig. \ref{fig:architecture}. 

We use the stages of the SegFormer sequentially, and make sure that the network has access to both, image features and interaction features, at each stage of the network. 
The input to the first stage only consists of $f^0_\text{mix} + f^0_\text{int}$. The input to later stages is the sum of $f^i_\text{mix}$, $f^i_\text{int}$ and the output of the previous stage. The output of all stages is then fed to the SegFormer prediction head which will predict the subsequent mask $\mathbf{m}_{\tau + 1}$.

\section{Experiments}

To show the efficacy of our our multi-modal fusion strategy, we test our system on DeLiVER \cite{zhang2023cmx, zhang2023delivering} and MFNet \cite{ha2017mfnet}. DeLiVER offers depth maps, lidar and event camera images as additional modalities, and MFNet offers thermal images. We use four different types of RGB foundation models as $\text{FM}_\text{RGB}$: DINOv2-B14 \cite{oquab2024dinov2, darcet2023vitneedreg}, MAE-B16 \cite{MaskedAutoencoders2021}, IJEPA-H16 \cite{assran2023self}, and the RGB encoder from the SAM model (SAM-B16) \cite{kirillov2023segment}. Further implementation details and qualitative examples can be found in sections \appimplementationdetails{} and \appqualitative{} of our supplementary material, respectively. Section \appresults{} of the supp. material contains further results on MCubeS \cite{Liang_2022_CVPR} and FMB \cite{liu2023segmif}. 

We use an NVIDIA V100 GPU. If we amortize the duration of the feature extraction over all clicks, our model takes 42 ms per click on average when tested on the MFNet dataset. If we only consider the response time in isolation, we measure 12 ms when averaged over 30 clicks. On the CPU (Intel\textregistered Xeon\textregistered CPU E5-2697 v4) we arrive at 206 ms per click. We thus consider our model to be real-time-capable.

\subsection{Results on DeLiVER}
The results on the DeLiVER dataset \cite{zhang2023delivering} are presented in Table \ref{tab:deliver_results}. 

\begin{table*}
\centering
\caption{Results on DeLiVER \cite{zhang2023delivering}. The leftmost column ($\text{FM}_\text{RGB}$) indicates the backbone we used. The usage of non-RGB modalities is indicated by a checkmark. In RGB-only cases the MMFuser is not present. }
\resizebox{0.9\linewidth}{!}{
\begin{tabular}{c|ccc|cccc|c|c}
    \toprule
     $\text{FM}_\text{RGB}$ & Depth & Lidar & Event & NoC@60 $\downarrow$ & NoC@70 $\downarrow$ & NoC@80 $\downarrow$ & NoC@90 $\downarrow$ & $\text{NoCMS}@(80,70)$ $\downarrow$ & $\text{FRMS}@(80,70)$ $\downarrow$ \\
     \midrule
     \multirow{8}{*}{DINOv2-B14} &  &  &  & 10.92 & 13.05 & 15.10 & 16.99 & 15.58 & 66.43 \\ 
      & \checkmark &  &  & 9.83 & \textbf{11.68} & \textbf{13.80} & \textbf{16.07} & \underline{14.55} & \underline{60.89} \\ 
      &  & \checkmark &  & 10.85 & 12.96 & 15.03 & 16.92 & 15.48 & 64.97 \\ 
      &  &  & \checkmark & 10.86 & 12.97 & 15.01 & 16.91 & 15.47 & 65.33 \\ 
      &  & \checkmark & \checkmark & 10.77 & 12.88 & 14.99 & 16.90 & 15.46 & 65.18 \\ 
      & \checkmark &  & \checkmark & \textbf{9.78} & \textbf{11.68} & \textbf{13.80} & \underline{16.08} & \textbf{14.50} & \textbf{60.19} \\ 
      & \checkmark & \checkmark &  & 9.82 & \textbf{11.68} & \underline{13.82} & 16.10 & 14.59 & 61.24 \\ 
      & \checkmark & \checkmark & \checkmark & \underline{9.79} & \underline{11.73} & \underline{13.82} & 16.09 & 14.59 & 61.06 \\ 
      \hline 
      \multirow{2}{*}{MAE-B16} &  &  &  & 11.88 & 13.87 & 15.68 & 17.33 & 16.11 & 69.50 \\ 
      & \checkmark &  &  & 10.16 & 11.97 & 13.96 & 16.13 & 14.72 & 61.85 \\ 
      \hline 
      \multirow{2}{*}{SAM-B16} &  &  &  & 11.63 & 13.59 & 15.47 & 17.19 & 15.93 & 68.39 \\ 
      & \checkmark &  &  & 10.11 & 11.89 & 13.94 & 16.14 & 14.67 & 61.61 \\ 
      \hline 
      \multirow{2}{*}{IJEPA-H16} &  &  &  & 11.97 & 14.05 & 15.89 & 17.48 & 16.34 & 68.86 \\ 
      & \checkmark &  &  & 10.31 & 12.07 & 14.01 & 16.20 & 14.85 & 62.15 \\ 
      \bottomrule
\end{tabular} 
}
\label{tab:deliver_results}
\end{table*}

The first thing to be noted is that we encounter very complex non-contiguous shapes. 
For this reason, we also look at lower IoU thresholds with $\Theta_\text{IoU} \in \{60, 70, 80, 90\}$. 
For the DINOv2-B14 we test all possible combinations of the available modalities. Using only a single non-RGB modality our fusion method works best with depth maps, causing a reduction in the NoC@80 of 1.3 clicks on average. 

The DeLiVER dataset \cite{zhang2023delivering} contains intentionally placed perturbations in lidar and event data. These modalities are thus less reliable.
This is reflected in our method performance, such as a NoC@90 of 16.07 for depth vs. 16.10 for depth and lidar. However, our method still causes improvements over the RGB-only case. 

For the new multi-surface interactive segmentation metrics (see Sec. \ref{sec:multi_surface_metrics}) we use $\Theta_\text{IoU} = 80$ and $\Theta_\text{Average-Iou} = 70$. 
The multi-modal fusion works best with a combination of depth and event data with a NoCMS@(80, 70) of 14.50 and a FRMS@(80, 70) of 60.19. 

We use depth to demonstrate the effectiveness of our fusion strategy for other backbones. 
Fig. \ref{fig:qualitative_paper} shows qualitative examples.

\subsection{Results on MFNet}
MFNet \cite{ha2017mfnet} offers thermal images as an additional modality. The general performance results can be seen in Table \ref{tab:mfnet_results}.

\begin{table}
    \centering
    
    \caption{Results on MFNet \cite{ha2017mfnet}. The leftmost column ($\text{FM}_\text{RGB}$) indicates the backbone we used. The usage of thermal images is indicated by a checkmark. In RGB-only cases the MMFuser is not present. }
    \resizebox{0.95\linewidth}{!}{
    \begin{tabular}{c|c|cc|c}
    \toprule
    $\text{FM}_\text{RGB}$ & Thermal & NoC@80 $\downarrow$ & NoC@90 $\downarrow$ & $\text{NoCMS}@(80,70)$ $\downarrow$ \\
    \midrule
    \multirow{2}{*}{DINOv2-B14} &  & 10.05 & 15.37 & 11.30 \\
    & \checkmark & \textbf{8.66} & \textbf{14.88} & \textbf{9.88} \\
    \hline
    \multirow{2}{*}{MAE-B16} &  & 11.28 & 16.14 & 12.40 \\
    & \checkmark & 9.39 & 15.38 & \underline{10.58} \\
    \hline
    \multirow{2}{*}{SAM-B16} &  & 11.01 & 15.93 & 12.30 \\
    & \checkmark & \underline{9.26} &  \underline{15.31} & 10.91 \\
    \hline
    \multirow{2}{*}{IJEPA-H16} &  & 12.06 & 16.81 & 13.45 \\
    & \checkmark & 9.79 & 15.62 & 11.35 \\
    \bottomrule
    \end{tabular}
    }
    
    \label{tab:mfnet_results}
\end{table}

The results in Table \ref{tab:mfnet_results} corroborate our observations on DeLiVER. When looking at RGB-only results, DINOv2-B14 delivers the best performance with a NoCMS@(80,70) of 11.30. This points to the model's great capability to perform geometrically intricate segmentation tasks. 
In all cases, our model successfully integrates information from thermal images. The most significant improvements can be observed when using IJEPA, reducing the NoC@90 by 1.19 and the NoCMS@(80, 70) by 2.10. However, it should be noted that thermal images are not sufficient to bridge the pre-exising performance gap between DINOv2 and IJEPA. DINOv2 using only RGB images (NoC@90 of 15.37) still outperforms the IJEPA-based model with thermal images (NoC@90 of 15.62).

\subsection{RGB-Only Baseline Performance}

We demonstrate the efficacy of our RGB-only baseline (see Table \ref{tab:comparisonmulti}). The baseline results from the architecture described in Section \ref{sec:architecture} if we remove the MMFuser. We only compare ourselves to models with similar conditions for a fair comparison. This especially means using a ViT-B as the backbone and the COCO+LVIS dataset \cite{ritm2022,lin2014microsoft,gupta2019lvis} as training data. The only exception are the SAM-based models in Table \ref{tab:comparison_davis_hqseg} and \ref{tab:comparison_other}. 
Most importantly, we only compare ourselves to late fusion models. We do so, because late fusion generally imposes a more difficult condition by sacrificing performance for the goal of improved efficiency, as described by \cite{schoen2025skipclick}.  We test on DAVIS \cite{perazzi2016davis}, HQSeg-44k \cite{ke2024segment}, GrabCut \cite{rother2004grabcut}, SBD \cite{hariharan2011sbd}, Berkeley \cite{martin2001berkeley, guinness2010berkeleytest} and SHSeg \cite{schoen2025skipclick}. 

\begin{table}
    \centering
    
    \caption{A comparison with other methods.\label{tab:comparisonmulti}}
    \begin{subtable}{0.45\textwidth}
    
    \centering
    \caption{Comparison of our baseline with other models on DAVIS and HQSeg-44k. The metric is always the NoC@90 (lower is better). Results for other methods taken from \cite{schoen2025skipclick} and \cite{huang2024hrsamefficientinteractivesegmentation}. } 
    \resizebox{0.65\textwidth}{!}{
    \begin{tabular}{l|c|c}
        \toprule
        \textbf{Architecture} & DAVIS & HQSeg-44k \\
        \midrule
        SAM \cite{kirillov2023segment} & \underline{5.14} & 7.46 \\
        MobileSAM \cite{mobile_sam} & 5.83 & 8.70 \\
        HQ-SAM \cite{ke2024segment} & 5.26 & 6.49 \\
        SegNext \cite{liu2024rethinking} & 5.34 & 7.18 \\
        Interformer \cite{huang2023interformer} & 5.45 & 7.17 \\
        HR-SAM \cite{huang2024hrsamefficientinteractivesegmentation} & 5.48 & 7.66 \\
        HR-SAM++ \cite{huang2024hrsamefficientinteractivesegmentation} & 5.41 & 7.47 \\
        SkipClick \cite{schoen2025skipclick} & \textbf{4.94} & \textbf{6.00} \\
        \hline
        Our Baseline & \underline{5.14} & \underline{6.30} \\
        \bottomrule
    \end{tabular}
    }
    \label{tab:comparison_davis_hqseg}
    \end{subtable}

    \begin{subtable}{0.45\textwidth}
    
    \centering
    \caption{Comparison of our baseline on further datasets with the NoC@90 metric (lower is better). Our model outperforms others on SBD and SHSeg. Results for other methods taken from \cite{huang2023interformer} and \cite{schoen2025skipclick}.}
    \resizebox{0.9\textwidth}{!}{
    \begin{tabular}{l|c|c|c|c}
        \toprule
        \textbf{Architecture} & GrabCut & SBD & Berkeley & SHSeg \\
        \midrule
        SAM \cite{kirillov2023segment} & - & - & - & 7.46 \\ 
        HQ-SAM \cite{ke2024segment} & - & - & - & 14.29 \\ 
        Interformer \cite{huang2023interformer} & \underline{1.50} & 6.34 & 3.14 & - \\
        SkipClick \cite{schoen2025skipclick} & \textbf{1.44} & \underline{6.18} & \textbf{2.45} & \underline{2.52} \\ 
        \hline
        Our Baseline & 1.54 &\textbf{ 6.10} & \underline{2.75} & \textbf{2.51} \\
        \bottomrule
    \end{tabular}
    }
    
    \label{tab:comparison_other}
    \end{subtable}

    \begin{subtable}{0.45\textwidth}
    \centering
    \caption{Comparison of our method with SkipClick on DeLiVER. In order to use depth in SkipClick, we integrated the MMFuser module as described in section \appskipclick{} in the supplementary material. }
    \resizebox{\linewidth}{!}{
    \begin{tabular}{l|c|c|c}
    \toprule
    \textbf{Architecture} & NoC@80 $\downarrow$ & NoC@90 $\downarrow$ & NoCMS@(80, 70) $\downarrow$ \\
    \midrule
    SkipClick + Depth & 14.54 & 16.53 & 14.90 \\ 
    Ours + Depth & \textbf{13.80} & \textbf{16.07} & \textbf{14.55} \\ 
    \bottomrule
    \end{tabular}
    }
    \label{tab:comparison_skipclick}
    \end{subtable}

\end{table}

The results can be seen in Tables \ref{tab:comparison_davis_hqseg} and \ref{tab:comparison_other}. We only compare ourselves to the other methods on datasets for which results are available. Our model generally produces competitive results and even outperforms most of the SAM-based methods. We are on-par with SAM on DAVIS. This is remarkable, since all of these methods made use of the large SA-1B dataset \cite{kirillov2023segment} with 1.1 billion masks at some point in their training. 
On SBD and SHSeg our model even outperforms all other methods. 
In general, however, SkipClick delivers the best performance and is thus the only method with better results on most of the datasets.

We also want to see how our method performs in comparison to SkipClick if both are given multi-modal information.
For this reason, we apply our multi-modal fusion strategy to SkipClick by integrating the MMFuser module. A description on how we do this can be found in section \appskipclick{} in the supplementary material.  
In Table \ref{tab:comparison_skipclick} we carry out the comparison on the DeLiVER dataset with depth maps as the additional modality. Our model outperforms SkipClick in the multi-modal setting. We thus assume our model to give us the best performance in a multi-modal setting.

\section{Conclusion}
In this paper, we introduced a system for interactive segmentation with multiple image-like input modalities. 
Our system is constructed to adhere to certain constraints. 
We want our model to run fast and we want the image feature computation to be completely decoupled from the integration of interactions. For this reason we opt for a late fusion model. 

On top of this, we introduce a metric that considers interactive segmentation with multiple, possibly adjacent surfaces per image. 
We propose a multi-modal feature fusion strategy that is capable of dealing with a black-box RGB backbone. We are able to show that this fusion strategy incurs improvements for multiple RGB backbones on multiple datasets for different non-RGB modalities. 
Finally, we are able to show that our RGB-only baseline offers competitive, and in some cases even superior performance for regular interactive segmentation tasks. 
We even outperform the currently best model in a multi-modal setting, when integrating our multi-modal fusion strategy into its architecture.

\bibliography{references}{}

\begin{thebibliography}{10}

\bibitem{li2023mask}
Feng Li, Hao Zhang, Huaizhe Xu, Shilong Liu, Lei Zhang, Lionel~M Ni, and
  Heung-Yeung Shum.
\newblock Mask dino: Towards a unified transformer-based framework for object
  detection and segmentation.
\newblock In {\em Proceedings of the IEEE/CVF conference on computer vision and
  pattern recognition}, pages 3041--3050, 2023.

\bibitem{xie2021segformer}
Enze Xie, Wenhai Wang, Zhiding Yu, Anima Anandkumar, Jose~M Alvarez, and Ping
  Luo.
\newblock Segformer: Simple and efficient design for semantic segmentation with
  transformers.
\newblock In {\em Neural Information Processing Systems (NeurIPS)}, 2021.

\bibitem{kienzle2024segformerpp}
Daniel Kienzle, Marco Kantonis, Robin Sch{\"o}n, and Rainer Lienhart.
\newblock Segformer++: Efficient token-merging strategies for high-resolution
  semantic segmentation.
\newblock {\em IEEE International Conference on Multimedia Information
  Processing and Retrieval (MIPR)}, 2024.

\bibitem{ritm2022}
Konstantin Sofiiuk, Ilya~A Petrov, and Anton Konushin.
\newblock Reviving iterative training with mask guidance for interactive
  segmentation.
\newblock In {\em 2022 IEEE International Conference on Image Processing
  (ICIP)}, pages 3141--3145. IEEE, 2022.

\bibitem{kirillov2023segment}
Alexander Kirillov, Eric Mintun, Nikhila Ravi, Hanzi Mao, Chloe Rolland, Laura
  Gustafson, Tete Xiao, Spencer Whitehead, Alexander~C Berg, Wan-Yen Lo, et~al.
\newblock Segment anything.
\newblock In {\em Proceedings of the IEEE/CVF international conference on
  computer vision}, pages 4015--4026, 2023.

\bibitem{chen2022focalclick}
Xi~Chen, Zhiyan Zhao, Yilei Zhang, Manni Duan, Donglian Qi, and Hengshuang
  Zhao.
\newblock Focalclick: Towards practical interactive image segmentation.
\newblock In {\em Proceedings of the IEEE/CVF Conference on Computer Vision and
  Pattern Recognition}, pages 1300--1309, 2022.

\bibitem{lowes2023interactive}
Mathias~M. Lowes, Jakob~L. Christensen, Bj{\o}rn~Schreblowski Hansen,
  Morten~Rieger Hannemose, Anders~B. Dahl, and Vedrana Dahl.
\newblock Interactive scribble segmentation.
\newblock In {\em Proceedings of the Northern Lights Deep Learning Workshop},
  volume~4, 2023.

\bibitem{zhang2020interactive}
Shiyin Zhang, Jun~Hao Liew, Yunchao Wei, Shikui Wei, and Yao Zhao.
\newblock Interactive object segmentation with inside-outside guidance.
\newblock In {\em Proceedings of the IEEE/CVF conference on computer vision and
  pattern recognition}, pages 12234--12244, 2020.

\bibitem{sofiiuk2020f}
Konstantin Sofiiuk, Ilia Petrov, Olga Barinova, and Anton Konushin.
\newblock f-brs: Rethinking backpropagating refinement for interactive
  segmentation.
\newblock In {\em Proceedings of the IEEE/CVF Conference on Computer Vision and
  Pattern Recognition}, pages 8623--8632, 2020.

\bibitem{Liu_2023_ICCV}
Qin Liu, Zhenlin Xu, Gedas Bertasius, and Marc Niethammer.
\newblock Simpleclick: Interactive image segmentation with simple vision
  transformers.
\newblock In {\em Proceedings of the IEEE/CVF International Conference on
  Computer Vision (ICCV)}, pages 22290--22300, October 2023.

\bibitem{ke2024segment}
Lei Ke, Mingqiao Ye, Martin Danelljan, Yu-Wing Tai, Chi-Keung Tang, Fisher Yu,
  et~al.
\newblock Segment anything in high quality.
\newblock {\em Advances in Neural Information Processing Systems}, 36, 2024.

\bibitem{huang2023interformer}
You Huang, Hao Yang, Ke~Sun, Shengchuan Zhang, Liujuan Cao, Guannan Jiang, and
  Rongrong Ji.
\newblock Interformer: Real-time interactive image segmentation.
\newblock In {\em Proceedings of the IEEE/CVF International Conference on
  Computer Vision}, pages 22301--22311, 2023.

\bibitem{liu2024rethinking}
Qin Liu, Jaemin Cho, Mohit Bansal, and Marc Niethammer.
\newblock Rethinking interactive image segmentation with low latency high
  quality and diverse prompts.
\newblock In {\em Proceedings of the IEEE/CVF Conference on Computer Vision and
  Pattern Recognition}, pages 3773--3782, 2024.

\bibitem{schoen2025skipclick}
Robin Sch\"on, Julian Lorenz, Daniel Kienzle, and Rainer Lienhart.
\newblock Skipclick: Combining quick responses and low-level features for
  interactive segmentation in winter sports contexts.
\newblock In {\em Proceedings of the Winter Conference on Applications of
  Computer Vision (WACV) Workshops}, pages 1247--1256, February 2025.

\bibitem{agustsson2019interactive}
Eirikur Agustsson, Jasper~RR Uijlings, and Vittorio Ferrari.
\newblock Interactive full image segmentation by considering all regions
  jointly.
\newblock In {\em Proceedings of the IEEE/CVF Conference on Computer Vision and
  Pattern Recognition}, pages 11622--11631, 2019.

\bibitem{andriluka2020efficient}
Mykhaylo Andriluka, Stefano Pellegrini, Stefan Popov, and Vittorio Ferrari.
\newblock Efficient full image interactive segmentation by leveraging
  within-image appearance similarity.
\newblock {\em arXiv preprint arXiv:2007.08173}, 2020.

\bibitem{rana2023dynamite}
Amit~Kumar Rana, Sabarinath Mahadevan, Alexander Hermans, and Bastian Leibe.
\newblock Dynamite: Dynamic query bootstrapping for multi-object interactive
  segmentation transformer.
\newblock In {\em Proceedings of the IEEE/CVF International Conference on
  Computer Vision}, pages 1043--1052, 2023.

\bibitem{li2024learning}
Kun Li, Hao Cheng, George Vosselman, and Michael~Ying Yang.
\newblock Learning from exemplars for interactive image segmentation.
\newblock {\em arXiv preprint arXiv:2406.11472}, 2024.

\bibitem{fusenet2016accv}
C.~Hazirbas, L.~Ma, C.~Domokos, and D.~Cremers.
\newblock Fusenet: incorporating depth into semantic segmentation via
  fusion-based cnn architecture.
\newblock In {\em Asian Conference on Computer Vision}, November 2016.

\bibitem{ha2017mfnet}
Qishen Ha, Kohei Watanabe, Takumi Karasawa, Yoshitaka Ushiku, and Tatsuya
  Harada.
\newblock Mfnet: Towards real-time semantic segmentation for autonomous
  vehicles with multi-spectral scenes.
\newblock In {\em 2017 IEEE/RSJ International Conference on Intelligent Robots
  and Systems (IROS)}, pages 5108--5115. IEEE, 2017.

\bibitem{sun2019rtfnet}
Yuxiang Sun, Weixun Zuo, and Ming Liu.
\newblock {RTFNet: RGB-Thermal Fusion Network for Semantic Segmentation of
  Urban Scenes}.
\newblock {\em {IEEE Robotics and Automation Letters}}, 4(3):2576--2583, July
  2019.

\bibitem{liu2023segmif}
Jinyuan Liu, Zhu Liu, Guanyao Wu, Long Ma, Risheng Liu, Wei Zhong, Zhongxuan
  Luo, and Xin Fan.
\newblock Multi-interactive feature learning and a full-time multi-modality
  benchmark for image fusion and segmentation.
\newblock In {\em International Conference on Computer Vision}, 2023.

\bibitem{zhang2023cmx}
Jiaming Zhang, Huayao Liu, Kailun Yang, Xinxin Hu, Ruiping Liu, and Rainer
  Stiefelhagen.
\newblock Cmx: Cross-modal fusion for rgb-x semantic segmentation with
  transformers.
\newblock {\em IEEE Transactions on Intelligent Transportation Systems}, 2023.

\bibitem{wang2022tokenfusion}
Yikai Wang, Xinghao Chen, Lele Cao, Wenbing Huang, Fuchun Sun, and Yunhe Wang.
\newblock Multimodal token fusion for vision transformers.
\newblock In {\em IEEE Conference on Computer Vision and Pattern Recognition
  (CVPR)}, 2022.

\bibitem{Liang_2022_CVPR}
Yupeng Liang, Ryosuke Wakaki, Shohei Nobuhara, and Ko~Nishino.
\newblock Multimodal material segmentation.
\newblock In {\em Proceedings of the IEEE/CVF Conference on Computer Vision and
  Pattern Recognition (CVPR)}, pages 19800--19808, June 2022.

\bibitem{zhang2023delivering}
Jiaming Zhang, Ruiping Liu, Hao Shi, Kailun Yang, Simon Rei{\ss}, Kunyu Peng,
  Haodong Fu, Kaiwei Wang, and Rainer Stiefelhagen.
\newblock Delivering arbitrary-modal semantic segmentation.
\newblock In {\em CVPR}, 2023.

\bibitem{broedermann2023hrfuser}
Tim Broedermann, Christos Sakaridis, Dengxin Dai, and Luc Van~Gool.
\newblock Hrfuser: A multi-resolution sensor fusion architecture for 2d object
  detection.
\newblock In {\em IEEE International Conference on Intelligent Transportation
  Systems (ITSC)}, 2023.

\bibitem{broedermann2024cafuser}
Tim Br{\"o}dermann, Christos Sakaridis, Yuqian Fu, and Luc Van~Gool.
\newblock Cafuser: Condition-aware multimodal fusion for robust semantic
  perception of driving scenes.
\newblock {\em IEEE Robotics and Automation Letters}, 2025.

\bibitem{jia2024geminifusion}
Ding Jia, Jianyuan Guo, Kai Han, Han Wu, Chao Zhang, Chang Xu, and Xinghao
  Chen.
\newblock {G}emini{F}usion: Efficient pixel-wise multimodal fusion for vision
  transformer.
\newblock In Ruslan Salakhutdinov, Zico Kolter, Katherine Heller, Adrian
  Weller, Nuria Oliver, Jonathan Scarlett, and Felix Berkenkamp, editors, {\em
  Proceedings of the 41st International Conference on Machine Learning}, volume
  235 of {\em Proceedings of Machine Learning Research}, pages 21753--21767.
  PMLR, 21--27 Jul 2024.

\bibitem{li2024stitchfusion}
Bingyu Li, Da~Zhang, Zhiyuan Zhao, Junyu Gao, and Xuelong Li.
\newblock Stitchfusion: Weaving any visual modalities to enhance multimodal
  semantic segmentation.
\newblock {\em arXiv preprint arXiv:2408.01343}, 2024.

\bibitem{Schon_2023_CVPR}
Robin Sch\"on, Katja Ludwig, and Rainer Lienhart.
\newblock Impact of pseudo depth on open world object segmentation with minimal
  user guidance.
\newblock In {\em Proceedings of the IEEE/CVF Conference on Computer Vision and
  Pattern Recognition (CVPR) Workshops}, pages 4809--4819, June 2023.

\bibitem{wang2025orderawareinteractivesegmentation}
Bin Wang, Anwesa Choudhuri, Meng Zheng, Zhongpai Gao, Benjamin Planche, Andong
  Deng, Qin Liu, Terrence Chen, Ulas Bagci, and Ziyan Wu.
\newblock Order-aware interactive segmentation, 2025.

\bibitem{oquab2024dinov2}
Maxime Oquab, Timothée Darcet, Theo Moutakanni, Huy~V. Vo, Marc Szafraniec,
  Vasil Khalidov, Pierre Fernandez, Daniel Haziza, Francisco Massa, Alaaeldin
  El-Nouby, Russell Howes, Po-Yao Huang, Hu~Xu, Vasu Sharma, Shang-Wen Li,
  Wojciech Galuba, Mike Rabbat, Mido Assran, Nicolas Ballas, Gabriel Synnaeve,
  Ishan Misra, Herve Jegou, Julien Mairal, Patrick Labatut, Armand Joulin, and
  Piotr Bojanowski.
\newblock Dinov2: Learning robust visual features without supervision.
\newblock {\em Transactions on Machine Learning Research Journal}, pages 1--31,
  2024.

\bibitem{darcet2023vitneedreg}
Timothée Darcet, Maxime Oquab, Julien Mairal, and Piotr Bojanowski.
\newblock Vision transformers need registers, 2023.

\bibitem{MaskedAutoencoders2021}
Kaiming He, Xinlei Chen, Saining Xie, Yanghao Li, Piotr Dollar, and Ross
  Girshick.
\newblock Masked autoencoders are scalable vision learners.
\newblock In {\em 2022 IEEE/CVF Conference on Computer Vision and Pattern
  Recognition (CVPR)}, pages 15979--15988. IEEE Computer Society, 2022.

\bibitem{assran2023self}
Mahmoud Assran, Quentin Duval, Ishan Misra, Piotr Bojanowski, Pascal Vincent,
  Michael Rabbat, Yann LeCun, and Nicolas Ballas.
\newblock Self-supervised learning from images with a joint-embedding
  predictive architecture.
\newblock In {\em Proceedings of the IEEE/CVF Conference on Computer Vision and
  Pattern Recognition}, pages 15619--15629, 2023.

\bibitem{lin2014microsoft}
Tsung-Yi Lin, Michael Maire, Serge Belongie, James Hays, Pietro Perona, Deva
  Ramanan, Piotr Doll{\'a}r, and C~Lawrence Zitnick.
\newblock Microsoft coco: Common objects in context.
\newblock In {\em Computer vision--ECCV 2014: 13th European conference, zurich,
  Switzerland, September 6-12, 2014, proceedings, part v 13}, pages 740--755.
  Springer, 2014.

\bibitem{gupta2019lvis}
Agrim Gupta, Piotr Dollar, and Ross Girshick.
\newblock {LVIS}: A dataset for large vocabulary instance segmentation.
\newblock In {\em Proceedings of the {IEEE} Conference on Computer Vision and
  Pattern Recognition}, 2019.

\bibitem{perazzi2016davis}
Federico Perazzi, Jordi Pont-Tuset, Brian McWilliams, Luc~Van Gool, Markus
  Gross, and Alexander Sorkine-Hornung.
\newblock A benchmark dataset and evaluation methodology for video object
  segmentation.
\newblock In {\em The IEEE Conference on Computer Vision and Pattern
  Recognition (CVPR)}, 2016.

\bibitem{rother2004grabcut}
Carsten Rother, Vladimir Kolmogorov, and Andrew Blake.
\newblock "grabcut": interactive foreground extraction using iterated graph
  cuts.
\newblock {\em ACM Trans. Graph.}, 23(3):309–314, August 2004.

\bibitem{hariharan2011sbd}
Bharath Hariharan, Pablo Arbeláez, Lubomir Bourdev, Subhransu Maji, and
  Jitendra Malik.
\newblock Semantic contours from inverse detectors.
\newblock In {\em 2011 International Conference on Computer Vision}, pages
  991--998, 2011.

\bibitem{martin2001berkeley}
D.~Martin, C.~Fowlkes, D.~Tal, and J.~Malik.
\newblock A database of human segmented natural images and its application to
  evaluating segmentation algorithms and measuring ecological statistics.
\newblock In {\em Proceedings Eighth IEEE International Conference on Computer
  Vision. ICCV 2001}, volume~2, pages 416--423 vol.2, 2001.

\bibitem{guinness2010berkeleytest}
Kevin McGuinness and Noel~E. O'Connor.
\newblock A comparative evaluation of interactive segmentation algorithms.
\newblock {\em Pattern Recogn.}, 43(2):434–444, February 2010.

\bibitem{huang2024hrsamefficientinteractivesegmentation}
You Huang, Wenbin Lai, Jiayi Ji, Liujuan Cao, Shengchuan Zhang, and Rongrong
  Ji.
\newblock Hrsam: Efficient interactive segmentation in high-resolution images,
  2024.

\bibitem{mobile_sam}
Chaoning Zhang, Dongshen Han, Yu~Qiao, Jung~Uk Kim, Sung-Ho Bae, Seungkyu Lee,
  and Choong~Seon Hong.
\newblock Faster segment anything: Towards lightweight sam for mobile
  applications.
\newblock {\em arXiv preprint arXiv:2306.14289}, 2023.

\bibitem{Schn2024AdaptingTS}
Robin Sch{\"o}n, Julian Lorenz, Katja Ludwig, and Rainer Lienhart.
\newblock Adapting the segment anything model during usage in novel situations.
\newblock {\em 2024 IEEE/CVF Conference on Computer Vision and Pattern
  Recognition Workshops (CVPRW)}, pages 3616--3626, 2024.

\bibitem{gan2023maas}
Wensheng Gan, Shicheng Wan, and Philip~S. Yu.
\newblock { Model-as-a-Service (MaaS): A Survey }.
\newblock In {\em 2023 IEEE International Conference on Big Data (BigData)},
  pages 4636--4645, Los Alamitos, CA, USA, December 2023. IEEE Computer
  Society.

\bibitem{achiam2023gpt}
Josh Achiam, Steven Adler, Sandhini Agarwal, Lama Ahmad, Ilge Akkaya,
  Florencia~Leoni Aleman, Diogo Almeida, Janko Altenschmidt, Sam Altman,
  Shyamal Anadkat, et~al.
\newblock Gpt-4 technical report.
\newblock {\em arXiv preprint arXiv:2303.08774}, 2023.

\bibitem{betker2023improving}
James Betker, Gabriel Goh, Li~Jing, Tim Brooks, Jianfeng Wang, Linjie Li, Long
  Ouyang, Juntang Zhuang, Joyce Lee, Yufei Guo, et~al.
\newblock Improving image generation with better captions.
\newblock {\em Computer Science. https://cdn. openai. com/papers/dall-e-3.
  pdf}, 2(3):8, 2023.

\bibitem{sun2022bbt}
Tianxiang Sun, Yunfan Shao, Hong Qian, Xuanjing Huang, and Xipeng Qiu.
\newblock Black-box tuning for language-model-as-a-service.
\newblock In {\em Proceedings of {ICML}}, 2022.

\bibitem{amazonnova}
Amazon.
\newblock User guide for amazon nova.
\newblock Accessed: 09-07-2025,
  https://docs.aws.amazon.com/pdfs/nova/latest/userguide/nova-ug.pdf.

\bibitem{liu2021swin}
Ze~Liu, Yutong Lin, Yue Cao, Han Hu, Yixuan Wei, Zheng Zhang, Stephen Lin, and
  Baining Guo.
\newblock Swin transformer: Hierarchical vision transformer using shifted
  windows.
\newblock In {\em Proceedings of the IEEE/CVF international conference on
  computer vision}, pages 10012--10022, 2021.

\bibitem{hu2024interactive}
Kangpeng Hu, Quansen Sun, Yinghui Sun, and Tao Wang.
\newblock Interactive segmentation by considering first-click intentional
  ambiguity.
\newblock In {\em Proceedings of the 32nd ACM International Conference on
  Multimedia}, pages 4823--4831, 2024.

\bibitem{yan2024piclick}
Cilin Yan, Haochen Wang, Jie Liu, Xiaolong Jiang, Yao Hu, Xu~Tang, Guoliang
  Kang, and Efstratios Gavves.
\newblock Piclick: Picking the desired mask from multiple candidates in
  click-based interactive segmentation.
\newblock {\em Neurocomputing}, 599:128083, 2024.

\bibitem{kingma2017adammethodstochasticoptimization}
Diederik~P. Kingma and Jimmy Ba.
\newblock Adam: A method for stochastic optimization, 2017.

\bibitem{dosovitskiyimage}
Alexey Dosovitskiy, Lucas Beyer, Alexander Kolesnikov, Dirk Weissenborn,
  Xiaohua Zhai, Thomas Unterthiner, Mostafa Dehghani, Matthias Minderer, Georg
  Heigold, Sylvain Gelly, et~al.
\newblock An image is worth 16x16 words: Transformers for image recognition at
  scale.
\newblock In {\em International Conference on Learning Representations}, 2021.

\bibitem{hendrycks2016gelu}
Dan Hendrycks and Kevin Gimpel.
\newblock Gaussian error linear units (gelus).
\newblock {\em arXiv preprint arXiv:1606.08415}, 2016.

\end{thebibliography}
\bibliographystyle{unsrt}

\newpage
\appendix

\subsection{Introduction to the Interactive Segmentation Problem}
\label{sec:intseg_intro}
In this section, we will introduce the interactive segmentation problem by describing the inputs and outputs of our interactive segmentation model. Afterwards we will describe the NoC (\underline{N}umber \underline{o}f \underline{C}licks) metric. The NoC is generally used to measure the performance of interactive segmentation systems based on clicks. 

\subsubsection{Interactive Segmentation Task}
In the standard interactive segmentation scenario, we assume to have an RGB image $\mathbf{x}_\text{img} \in \mathbb{R}^{H \times W \times 3}$. This image contains an object whose surface we want to segment. 
More precisely, we want to create a mask $\mathbf{m} \in \{0, 1\}^{H \times W}$ with $m_{i, j} = 1$ if and only if the pixel $(i, j)$ belongs to the desired surface, while $m_{i, j} = 0$ otherwise.
In order to create such a mask, we will leverage a neural network $N_\text{IntSeg}$. $N_\text{IntSeg}$ has been trained to predict high-quality masks by using a certain form of user guidance. In this paper, the only form of user interaction are iterative clicks. 
The user will repeatedly carry out the following steps: 
\begin{enumerate}
    \item The user is shown the currently estimated mask $\mathbf{m}_\tau$, with $\tau$ being the index of the current round. In the initial round, we will not have a current estimate of the mask. Therefore we define $\mathbf{m}_0$ to only consist of $0$s (background only). 
    \item In case the user judges the mask to be of insufficient quality, they place a corrective click $\mathbf{p}_{\tau} = (i_{\tau}, j_{\tau}, l_\tau)$ on a falsely annotated position. Such a click consists of a coordinate $(i_{\tau}, j_{\tau}) \in \{1, ..., H\} \times \{1, ..., W\}$ and a label $l_\tau \in \{+, -\}$. If the user places the click with the left mouse button, they indicate that the position belongs to the foreground ($l_\tau = +$), while the right mouse button indicates a background position ($l_\tau = -$). 
    \item The network predicts a new mask 
    \begin{equation}
        \mathbf{m}_{\tau+1} = N_\text{IntSeg}(\mathbf{x}_\text{img}, \mathbf{p}_{0:\tau}, \mathbf{m}_\tau), 
    \end{equation}
    where $\mathbf{p}_{0:\tau}$ are all so far accumulated clicks and $\mathbf{m}_\tau$ is the previous faulty mask.
\end{enumerate}
These steps are carried out repeatedly until the user arrives at a resulting mask $\mathbf{m}_\text{Res}$ of sufficient quality. 

\subsubsection{NoC Metric}
The mode of interaction from the previous paragraph strongly insinuates that the quality of the mask can be easily increased by adding more click annotations. 
If the user were to just continue clicking long enough, the mask could likely reach an arbitrary level of quality. 
The most commonly used metric to measure the performance of interactive segmentation systems follows this intuition. 
The $\text{NoC}@\Theta_\text{IoU}$ metric measures the minimal \underline{N}umber \underline{o}f \underline{C}licks that is necessary to create a mask of sufficient quality.
We consider a predicted mask $\mathbf{m}_\tau$ to be of sufficient quality if its IoU with an existing ground truth mask $\mathbf{m}_\text{GT}$ reaches or surpasses a given threshold $\Theta_\textbf{IoU}$. 
Formally, this is the case if  
\begin{equation}
    \text{IoU}(\mathbf{m}_\tau, \mathbf{m}_\text{GT}) = \frac{|\mathbf{m}_\tau \cap \mathbf{m}_\text{GT}|}{|\mathbf{m}_\tau \cup \mathbf{m}_\text{GT}|} \cdot100 \ge \Theta_\text{IoU}. 
\end{equation} 
It should be noted that we follow common practice \cite{ritm2022, huang2024hrsamefficientinteractivesegmentation, kirillov2023segment, schoen2025skipclick} and express the IoU in percentage points in a range of $[0, 100] \subset \mathbb{R}$ instead of a ratio in the range of $[0, 1] \subset \mathbb{R}$. 

It remains possible that the network is not capable of segmenting a surface with a satisfying degree of quality at all. In order to account for such cases, the number of clicks is usually capped to a value of $n_\text{max} = 20$. If the number of clicks exceeds $n_\text{max}$, we consider the attempt to segment the object as a failure and use $n_\text{max}$ as a surrogate value in computing the average NoC on a dataset.  
In some works the authors decide to either measure the number of failures \cite{ritm2022, liu2024rethinking} or the percentage of failures (i.e. the failure rate) \cite{Schn2024AdaptingTS}. 
As we do not have an actual human user at our disposal during the testing process, the selection of the click placement is simulated automatically. We follow the simulation strategy described by \cite{ritm2022}.

\subsection{Oursourcing Subtasks for Interactive Segmentation}
\subsubsection{Scenario Discussion}

We now discuss the possibility to outsource subtasks of interactive segmentation and why it is beneficial to be able to deal with a black-box RGB backbone. Recently, a number of companies have trained large foundation models that are deliberately kept private \cite{gan2023maas, achiam2023gpt, betker2023improving, sun2022bbt}. This phenomenon also extends to image processing backbones being offered as a service. A real-world example of this is Amazon Nova \cite{amazonnova}. We assume this trend to continue, leading to private foundation models with increasing performance being offered as a service. 
As we have no direct access to them, these models can only be used as a black-box. Training or altering such a model is not possible. 

\begin{figure*}
    \centering
    \includegraphics[width=0.85\textwidth]{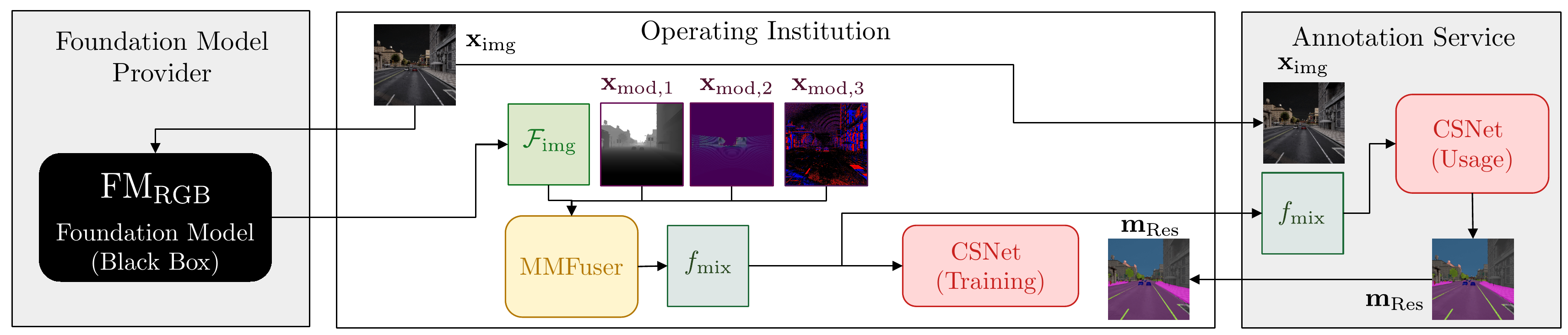}
    \caption{A scenario for interactive segmentation where the RGB foundation model and the the annotation task have been outsourced.
    In this scenario, we are the \emph{operating institution} and will act on behalf of our interests. 
    The RGB foundation model $\text{FM}_\text{RGB}$ has been outsourced to an external \emph{foundation model provider}. This makes $\text{FM}_\text{RGB}$ a black-box. We first use $\text{FM}_\text{RGB}$ to extract features $\mathcal{F}_\text{img}$ from RGB images $\mathbf{x}_\text{img}$. Afterwards, we integrate the information from non-RGB modalities $\mathbf{x}_{\text{mod}, i}$ using our MMFuser. From MMFuser, we obtain a mixed feature tensor $f_\text{mix}$. $f_\text{mix}$ is the input to CSNet, along with the clicks to generate the masks. The annotation task itself is outsourced to an external \emph{annotation service}, which only receives $\mathbf{x}_\text{img}$ and $f_\text{mix}$, and gives us the high-quality result masks $\mathbf{m}_\text{Res}$. As $f_\text{mix}$ already contains all non-RGB information, we can avoid giving the non-RGB modalities $\mathbf{x}_{\text{mod}, i}$ to the annotation service. This is beneficial, as we consider  $\mathbf{x}_{\text{mod}, i}$ valuable data. A detailed discussion of our network and modules such as MMFuser and CSNet can be found in the main paper's method section. }
    \label{fig:saas_scenario}
\end{figure*}

In our scenario, the RGB backbone $\text{FM}_\text{RGB}$ for image feature extraction is such an external foundation model. 
The external party providing the foundation model service will be called \emph{foundation model provider}. The scenario is depicted in Figure \ref{fig:saas_scenario}. Therein, we are the \emph{operating institution}, and will act on behalf of our interest.
We can send RGB images $\mathbf{x}_\text{img}$ to the foundation model provider and receive a set of image features $\mathcal{F}_\text{img}$. 
Most multi-modal fusion strategies for segmentation assume the backbone to be either trainable or at least allow backpropagation through the network as a pre-condition. 
As this is not possible in our scenario, we will devise an asymmetric multi-modal fusion strategy that does not depend on the trainability and internal workings of the RGB backbone. We are thus able to effectively treat the RGB backbone as a black-box whilst still profiting from the additional knowledge from other modalities.

Since our method is designed to be capable of dealing with a black-box, using an external foundation model as an RGB backbone is no obstacle during training or during usage: 
\begin{itemize}
    \item \textbf{During training}, we send all RGB images in our training dataset to the foundation model provider for feature extraction. The \emph{foundation model provider} responds with the extracted features $\mathcal{F}_\text{img}$ for our entire dataset. This feature extraction can be executed once in bulk before the training even starts. We thus do not have to attend it or wait for a response during training. On our side, we can effectively train all other components of the network on the extracted features $\mathcal{F}_\text{img}$. 
    \item \textbf{During usage}, we usually use interactive segmentation systems to annotate ground truth masks for entirely new datasets. We can send the images of this new dataset to the foundation model provider, which will in turn send the extracted features $\mathcal{F}_\text{img}$ to us. Similar to the training, this feature extraction can be carried out in bulk without any user attending to it. In case we also have non-RGB modalities, we then use the MMFuser module to generate the mixed feature tensor $f_\text{mix}$. Both, $\mathcal{F}_\text{img}$ and $f_\text{mix}$, only need to be created once per image. The interactive annotation itself can then be repeatedly carried out using CSNet. 
\end{itemize}

It should be noted that our framework based on a black-box RGB backbone does not actually necessitate the use of an external foundation model service. In only offers the possibility to do so. The RGB backbone may just as well be present locally. In fact, during our experiments, we only use local RGB backbones $\text{FM}_\text{RGB}$. 
We will simulate $\text{FM}_\text{RGB}$ being an external black-box by freezing the model and refraining from any architectural alterations. 

Since we want to annotate data in an interactive segmentation setting, we also assume the existence of a third party, the \emph{annotation service}. Once we receive the extracted RGB features from the foundation model provider, we will enrich those features with additional modalities using a multi-modal fusion strategy. 
These enriched features are then sent to the \emph{annotation service}. The external \emph{annotation service} will only receive the RGB image, the enriched features, and a small interactive segmentation network that operates on the features and user clicks (see Fig. \ref{fig:saas_scenario}). 

\subsubsection{Notes on used Network Bandwidth} 
First, it should be mentioned that the size of tensors fabricated by foundation model providers, such as AWS Nova or LeewayHertz, strongly depends on what the user demands. In fact, the handbook for AWS Nova, explicitly mentions this \cite{amazonnova} and lists examples for image resolution and resulting token numbers. For example, an image with a resolution of $900 \times 900$ results in approximately 1300 tokens according to the handbook. From this, we can infer their approximate patch size as 
\begin{equation}
    \frac{900}{\sqrt{1300}} \approx 25. 
\end{equation}
Assuming that the images have a resolution of $448 \times 448$, we would arrive at a spatial token resolution of $18 \times 18$. 
With an embedding dimension of 1280, a float32 encoding, and 4 extracted tensor per image, this implies the transfer of 
\begin{equation}
    18 \cdot 18 \cdot 1280 \cdot 4 \cdot 32 \text{  bits} = 53084160 \text{ bits}
\end{equation} per image. With a consumer connection of 500 mbit per second, a single image only needs 0.1061 seconds for transfer, which is much faster than any user could annotate an image. On top of this, the operation can be carried out in bulk without the user attending to it. We thus do not assume network bandwidth to be a problem.

\subsection{Differences to Other Work Related to Interactive Segmentation for Multiple Surfaces in an Image}
The main paper does not present the very first work on interactive segmentation with multiple surface in the same image. However, we deem all previous work to be too different from our work to allow for a direct comparison. On top of this, we come to find their interaction mechanisms quite arbitrary. 
In \cite{agustsson2019interactive}, the authors propose a mechanism where an initial segmentation is generated using specifically chosen border points. Corrections can only be carried out as an extension mechanism for existing regions, making corrections on non-exising regions impossible. 
The authors of MagicPaint \cite{andriluka2020efficient} provide a vast variety of input tools such as freeform scribbles, variable stroke thickness and filling mechanisms. In contrast these methods our mechanism only works with clicks. 
DynaMITe \cite{rana2023dynamite} is the only purely click-based system for multiple surfaces in the same image. While our system allows for a sequential creation of masks, DynaMITe enforces a direct manipulation of the multi-surface mask. This direct manipulation requires the user to reconsider their class selection after every click. Additionally, the multi-surface aspect and the accompanying interaction mechanism are deeply connected to the architecture of DynaMITe. Thus, the interaction mechanism and the metrics within the multi-surface case of the DynaMITe system are too different from our scenario to allow for a meaningful performance comparison. 
However, as the DynaMITE provides results for single-surface interactive segmentation, we can make a comparison for the single-surface case. The comparison can be found in Table \ref{tab:dynamite_comparison}. Nonetheless, it should be noted that DynaMITe uses the considerably different Swin backbone \cite{liu2021swin}. 
\begin{table}[]
    \centering
    \caption{A comparison of our baseline model with DynaMITe for the single-surface interactive segmentation task. In all cases the metric is the NoC@90, for which a lower value indicates better performance. It should be noted that DynaMITe uses a different backbone than us. }
    \label{tab:dynamite_comparison}
    \begin{tabular}{c|c|c|c|c}
    \toprule
    \textbf{Model} & GrabCut & Berkeley & SBD & DAVIS \\ 
    \midrule
    DynaMITe (Swin-L) & 1.72 & 1.90 & 5.64 & 5.09 \\
    DynaMITe (Swin-T) & 1.78 & 1.96 & 6.32 & 5.23 \\
    MMMS (ours, ViT-B) & 1.54 & 2.75 & 6.10 & 5.14 \\
    \bottomrule
    \end{tabular}
\end{table}

In addition to the aforementioned methods, there is also work that is marginally related to our paper. iCMFormer++ \cite{li2024learning} is specialized for cases in which multiple object instances of the same class are present in the same image. The segmentation is carried out in two rounds. In the first round, a segmentation mask for a single object instance of the targeted class is created. Using the created mask, this object is then cropped from the image and given to the network in the second round. In the second round, the network is used to create a semantic mask covering all objects of that class as a single surface. 

There is also work dealing with target ambiguity in interactive segmentation, such as DISNet \cite{hu2024interactive} and PiClick \cite{yan2024piclick}. Target ambiguity deals with the problem that a single click does not clearly tell the network which surface to segment. \emph{E.g. if I click on the tire of a car, do I want to to only segment the tire, or the complete car?} In contrast to this, multi-surface segmentation deals with the overlap of multiple annotated masks per image. Although somewhat related, these papers tackle different challenges within interactive segmentation.

\subsection{Further Notes on Computational Efficiency}
In the main paper, we already include some remarks on the response time. We use an Nvidia V100 as the GPU and a Intel\textregistered Xeon\textregistered CPU E5-2697 v4 as the CPU. 
If we amortize the duration of the feature extraction over all clicks, our model takes 42 ms per click on average when tested on the MFNet dataset using the GPU. If we only consider the response time in isolation, we measure 12 ms when averaged over 30 clicks. On the CPU, we arrive at 206 ms per click. We thus consider our model to be real-time-capable.

Here, we extend our discussion regarding the computational efficiency of our model. We compare our response time with response times of existing models using the measurements from \cite{huang2023interformer}. It should be noted that the authors of \cite{huang2023interformer} did not disclose which CPU they used. The comparison can be found in Table \ref{tab:responsetime}. 

\begin{table}
	\centering
	\caption{We compare the response time or our model with measurements from other models from \cite{huang2023interformer}. }
	\label{tab:responsetime}
	\begin{tabular}{c|cc}
		\toprule 
		\textbf{Model} & \textbf{Device} & \textbf{Response Time} \\
		\midrule
		Interformer-Light (ViT-B) & CPU & 0.19 \\
		Interformer-Tiny (ViT-L) & CPU & 0.32 \\ 
		SimpleClick (ViT-B) & CPU & 1.51 \\
		SimpleClick (ViT-L) & CPU & 3.33 \\
		SimpleClick (ViT-H) & CPU & 7.76 \\
		MMMS (ours, ViT-B) & CPU & 0.206 \\
		MMMS (ours, ViT-B) & GPU & 0.012 \\
		\bottomrule
	\end{tabular}
\end{table} 

A second aspect of computational efficiency lies in the memory usage. We measure the occupied VRAM during evaluation on our NVIDIA V100 and count the number of parameters in Table \ref{tab:memory_usage}. We do this for the RGB-only case and progressively add depth maps, lidar and event camera images. We choose the DINOv2-ViT-B based model as it delivered the best performance. 

\begin{table}
	\centering
	\caption{We measure the occupied VRAM during evaluation and count the number of parameters. The first row only uses RGB as input. In the rows beneath, we progressively add \underline{D}epth, \underline{L}idar, and \underline{E}vent camera images.}
	\label{tab:memory_usage}
	\begin{tabular}{c|cc}
		\toprule 
		\textbf{Modalities} & \textbf{VRAM (MiB)} & \textbf{\# params (mio.)} \\
		\midrule
		RGB & 1474 & 125.2 \\
		RGB+D & 1582 & 144.1 \\
		RGB+D+L & 1686 & 159.7 \\ 
		RGB+D+L+E & 1772 & 175.3 \\
		\bottomrule
	\end{tabular}
\end{table}

\subsection{Combining the Surface-Specific Masks into a Joint Mask}
Whenever we segment multiple surfaces in the same image, we are able to combine them to a joint mask. This joint mask assigns each pixel as either belonging to one of the surfaces or to non of them (background). In this section, we will describe how the multi-surface interactive segmentation problem from the main paper relates to the construction of such a joint mask. We will first describe how the joint mask is constructed in context of the classical interactive segmentation problem. Afterwards, we extend the procedure of constructing a joint mask to the multi-surface evaluation framework where some masks may be revisited. We will also discuss how the surface-specific masks are extracted from the joint mask. 

\subsubsection{Joint Masks in Classical Interactive Segmentation}
We now discuss how a joint mask is constructed, when we work with the assumptions of classical interactive segmentation. Let $L$ be the number of surfaces that we want to segment. As previously stated in the main paper, we want to create a set 
\begin{equation} 
	\mathcal{S}_{\mathbf{m}} = \left\{ \mathbf{m}^{(1)},  ..., \mathbf{m}^{(L)} \right\}. 
\end{equation}
of segmentation masks. In the setting of classical interactive segmentation, each of these masks is considered in isolation. To create a joint mask $\mathbf{m}^\text{joint} \in \{0, 1, ..., L\}^{H \times W}$, the surface-specific masks are created sequentially and pasted onto the joint mask. We start out with an initial joint mask $\mathbf{m}^{\text{joint}, 0}$ that is completely filled with 0s. In our case a pixel-value of 0 indicates that the pixel belongs to the background, whereas a pixel-value $l = 1, ..., L$ indicates the pixel belongs to the $l$-th surface. As soon as the $l$-th mask is created, we update the pixels $(i, j)$ in the previous joint mask with the rule
\begin{equation}
	m^{\text{joint}, l}_{i, j} = \begin{cases}
		l &, \text{ if } m^{(l)}_{i, j} = 1. \\
		m^{\text{joint}, l-1}_{i, j} &, \text{ if } m^{(l)}_{i, j} = 0. 
	\end{cases}
\end{equation}
In this way, the masks are each constructed and added to the joint mask. As each mask is considered in isolation, no mask is revisited. We eventually end up with the final joint mask $\mathbf{m}^\text{joint} = \mathbf{m}^{\text{joint}, L}$. 

\subsubsection{Joint Masks in Multi-Surface Interactive Segmentation}
In our extended multi-surface evaluation mechanism, the masks for some surfaces may be revisited for correction. 
Instead of simply adding masks in a linear order $l = 1, ..., L$ we continuously pick the worst mask beloning to surface $k$ and improve it. Afterwards, the improved mask can be used to update the joint mask $\mathbf{m}^{\text{joint}, \lambda - 1}$ to $\mathbf{m}^{\text{joint}, \lambda}$. Note that we change the index notation from $l$ to $\lambda$ to avoid confusion. 
We also have to extend our update rule. Once we consider that some surfaces may be revisited, we also have the possibility of decreasing the size of a surface. Assume we currently correct the mask for surface $k$ and by doing so remove a few pixels that previously belonged to that surface in the joint mask. In this case, the removed pixels are set to the background class in the updated joint mask. 
Our extended update rule for pixels $(i, j)$ is 
\begin{equation}
	\label{c6:eq:ms_joint_mask}
	m^{\text{joint}, \lambda + 1}_{i, j} = \begin{cases}
		k &, \text{ if } m^{(k)}_{i, j} = 1. \\
		0 &, \text{ if } (m^{(k)}_{i, j} = 0) \wedge (m^{\text{joint}, \gamma}_{i, j} = k).  \\
		m^{\text{joint}, \lambda}_{i, j} &, \text{ if } (m^{(k)}_{i, j} = 0) \wedge (m^{\text{joint}, \gamma}_{i, j} \neq k).   
	\end{cases}
\end{equation}

Finally, we mention how we extract a binary mask $\mathbf{m}^{(k)}$ for the $k$-th surface of the joint mask. We take all pixel belonging to that surface as foreground pixels and the rest as background pixels. We extract pixels $(i, j)$ according to the rule.
\begin{equation}
    m^{(k)}_{i, j} = \begin{cases}
        1 &, \text{ if } m^\text{joint}_{i, j} = k. \\
        0 &, \text{ otherwise.} 
    \end{cases}
\end{equation}

\subsection{Further Implementation Details}
As we want to show the general efficacy of our multi-modal fusion strategy, we use datasets that offer different non-RGB modalities. 
The first dataset we use in the main paper is the synthetic DeLiVER dataset \cite{zhang2023cmx, zhang2023delivering} with the split from \cite{zhang2023delivering}. In addition to the RGB images, DeLiVER contains depth maps, lidar and event camera images as additional modalities. We use the lidar representation from \cite{zhang2023delivering}. 
The second dataset in the main paper is MFNet \cite{ha2017mfnet}, which provides thermal images as an additional modality. For MFNet we use the splits provided with the dataset. 
Furthermore, we test our method on FMB \cite{liu2023segmif} and MCubeS \cite{Liang_2022_CVPR} for which the results can be found in the supplementary material.

We train for 100 epochs on DeLiVER and MFNet, while training for 400 epochs on FMB and MCubeS. We found these training durations promising in preliminary experiments.
We use the Adam optimizer \cite{kingma2017adammethodstochasticoptimization} with a batch size of 8, a learning rate $\mu = 5 \cdot 10^{-5}$ and $\beta_1 = 0.9, \beta_2=0.999$.
If we train for 100 epochs, a scheduler reduces the learning by $\frac{1}{10}$ at epochs 95 and 100. 
If we train for 400 epochs, this reduction happens at epochs 390 and 400. During training and testing, we use a resolution of $448 \times 448$ unless said otherwise. During training, this resolution is imposed by random crops. During testing, we follow common practice \cite{Liu_2023_ICCV, ritm2022} and rescale to this size.

Our multi-modal fusion strategy works on a variety of vision foundation models. We test it on four different backbones as our RGB foundation model $\text{FM}_\text{RGB}$. All of these models are vision transformers \cite{dosovitskiyimage}. 
The first model is a ViT-B that has been pretrained with DINOv2 \cite{oquab2024dinov2, darcet2023vitneedreg}, and has a patch size of 14. 
The second model is a ViT-B pretrained with the masked auto-encoder (MAE) \cite{MaskedAutoencoders2021} framework. 
We also use the ViT-B encoder that has been trained as part of the Segment Anything Model (SAM) \cite{kirillov2023segment}, since we assume this backbone to be capable of producing features that are beneficial for segmentation tasks. The SAM backbone only operates on a resolution of $1024 \times 1024$. For this reason, we upscale the input images before feeding them to the backbone, and downscale the resulting feature tensors afterwards. 
On top of this, we use a ViT-H pretrained with IJEPA \cite{assran2023self}. Here, we use the \emph{huge (H)} model as no smaller size was available. 
In order to simulate $\text{FM}_\text{RGB}$ being a black-box, we do not train or backpropagate through the RGB backbones at any time.
Our RGB-only baseline is almost identical to our multi-modal model, apart from the MMFuser being removed completely. In all cases, the models receive the RGB images as input. 
We use a SegFormer-B1 encoder inside the MMFuser and a SegFormer-B0 \cite{xie2021segformer} as the basis for CSNet.

\subsection{Further Results}

\subsubsection{Ablation of the CrossBlock}
In Table \ref{tab:ablation_mfnet}, we can see the results of our ablation study of our CrossBlock on the MFNet dataset. In the first line, neither the efficient cross-attention (EffCA) nor the subsequent MLP are used. In this case, the feature tensors from different modalities are just added. 
Only adding a single one of the two submodules has different effects depending on how demanding the metric is. Using the efficient cross-attention or the MLP on its own, slightly degrades the performance on the easier NoC@80 metric (8.81 vs 8.93 and 8.88, respectively). 
Once we use the NoC@90 metric, which demands a higher degree of detail, we also see improvements using a single submodule. 
This phenomenon can be attributed to the fact that the NoC metric is likely to not increase linearly over different IoU thresholds, even when measured on the same model.
Using both, EffCA and the MLP, incurs improvements in all metrics.

\begin{table}[]
    \centering
    \caption{Ablation study of the CrossBlock on MFNet. In all cases, we use the thermal images and the backbone is a DINOv2-B14. }
    \begin{tabular}{cc|cc|c}
    \toprule
    EffCA & MLP & NoC@80 $\downarrow$ & NoC@90 $\downarrow$ & $\text{NoCMS}@(80,70)$ $\downarrow$ \\
    \midrule
    &  & 8.81 & 15.05 & 10.11 \\
    \checkmark &  & 8.93 & 15.04 & 10.23 \\
    & \checkmark & 8.88 & 14.93 & 10.19 \\
    \checkmark & \checkmark & \textbf{8.66} & \textbf{14.88} & \textbf{9.88} \\
    \bottomrule
    \end{tabular}
    \label{tab:ablation_mfnet}
\end{table}

\subsubsection{Results on FMBNet} 
We also evaluate the effectiveness of our multi-modal fusion strategy on the FMB dataset \cite{liu2023segmif}. In addition to RGB images, FMB offers infrared images as a supplementary modality. The results can be found in Table \ref{tab:fmb_results}.
If we compare different backbones, we see similar results as with other datasets. When taking a look at the NoC@90 metric in the RBG-only setting the ranking is as follows: DINOv2-B14 is best with 12.97, followed by SAM-B16 with 13.80, MAE-B16 with 14.00 and IJEPA-H16 with 14.52. 
Using the infrared images leads to improvements in all cases. The strongest improvement can be observed on the NoC@90 when using the IJEPA-based backbone. There, the number of clicks is reduced by 1.57 on average. We attribute this to IJEPA as it is generally worse than the other models in its role as $\text{FM}_\text{RGB}$. 

As the DINOv2-based model already produces the best RGB features, the effects of including the information from infrared images are comparatively smaller. On the NoC@90 metric we see an average improvement of 1.05 clicks. The NoCMS@(80, 70) metric is reduced by 0.91. 
Another aspect that we can observe is that the NoCMS@(80, 70) is always higher than the regular NoC@80. We attribute this to the NoCMS metric accounting for competition between multiple surfaces in the same image. As the NoCMS@(80, 70) metric requires the revisiting of previous surfaces, we assume this metric to be more demanding.

\begin{table}
\centering
\caption{Results on the FMB dataset \cite{liu2023segmif}.  The leftmost column ($\text{FM}_\text{RGB}$) indicates the backbone we used. The usage of infrared images is indicated by a checkmark. In RGB-only cases the MMFuser is not present.}
\resizebox{\linewidth}{!}{
\begin{tabular}{c|c|cc|c}
    \toprule
    $\text{FM}_\text{RGB}$ & Infrared & NoC@80 $\downarrow$ & NoC@90 $\downarrow$ & $\text{NoCMS}@(80,70)$ $\downarrow$ \\
    \midrule
    \multirow{2}{*}{DINOv2-B14} &  & 7.86 & 12.97 & 8.28 \\
    & \checkmark & \textbf{7.07} & \textbf{11.92} & \textbf{7.37} \\
    \hline
    \multirow{2}{*}{MAE-B16} &  & 8.46 & 14.00 & 8.98 \\
    & \checkmark & \underline{7.56} & 12.64 & 8.12 \\
    \hline
    \multirow{2}{*}{SAM-B16} &  & 8.35 & 13.80 & 8.89 \\
    & \checkmark & 7.55 & \underline{12.53} & \underline{8.08} \\
    \hline
    \multirow{2}{*}{IJEPA-H16} &  & 9.02 & 14.52 & 9.80 \\
    & \checkmark & 7.72 & 12.95 & 8.30 \\
    \bottomrule
\end{tabular}
}
\label{tab:fmb_results}
\end{table}

\subsubsection{Results on MCubeS}
The results on the MCubeS dataset \cite{Liang_2022_CVPR} can be found in Table \ref{tab:mcubes_results}. In addition to RGB images, this dataset offers three non-RGB modalities: The \emph{Angle of Linear Polarization (AoLP)} and the \emph{Degree of Linear Polarization (DoLP)} as measured by a camera with a polarizaion filter, and \emph{Near Infrared (NIR)} images. RGB images are given to the model in all cases. 

For the model with DINOv2-B14 in the role of $\text{FM}_\text{RGB}$, we first test the effect of using each modality on its own. Our multi-modal fusion mechanism performs best when using the DoLP, which allows the model to reduce the NoC@90 by 0.42, the NoC@80 by 0.63 and the NoCMS@(80, 70) by 0.46. Our model successfully employs all non-RGB modalities when used in isolation. 
Using all of them at once improves the performance even further, reducing the NoCMS@(80, 70) from 16.77 to 16.13. The failure rate FRMS@(80, 70) is reduced from 65.15 to 61.98. 
As using all modalities at the same time is the most promising strategy, we also test this for other versions of $\text{FM}_\text{RGB}$. For MAE the NoC@90 improves from 18.66 to 17.83. When looking at the multi-surface segmentation metric NoCMS@(80, 70), the largest improvement can be observed for IJEPA with a reduction from 17.68 to 16.29. Here, we also see the largest improvement in the FRMS@(80, 70) from 70.95 to 61.90.

\begin{table*}
    \centering
    \caption{Results on the MCubeS dataset \cite{Liang_2022_CVPR}. The leftmost column ($\text{FM}_\text{RGB}$) indicates the backbone we used. The usage of non-RGB modalities is indicated by a checkmark. In RGB-only cases the MMFuser is not present.}
    \resizebox{0.85\linewidth}{!}{
    \begin{tabular}{c|ccc|cccc|c|c}
    \toprule
    $\text{FM}_\text{RGB}$ & AoLP & DoLP & NIR & NoC@60 $\downarrow$ & NoC@70 $\downarrow$ & NoC@80 $\downarrow$ & NoC@90 $\downarrow$ & $\text{NoCMS}@(80,70)$ $\downarrow$ & $\text{FRMS}@(80,70)$ $\downarrow$ \\
    \midrule
    \multirow{5}{*}{DINOv2-B14} &  &  &  & 11.94 & 14.09 & 16.29 & 18.35 & 16.77 & 65.15 \\
    & \checkmark &  &  & 11.55 & 13.75 & 15.93 & 18.07 & 16.50 & 64.36 \\
    &  & \checkmark &  & 11.14 & 13.38 & 15.66 & 17.93 & 16.31 & 62.93 \\
    &  &  & \checkmark & 11.10 & 13.42 & 15.72 & 18.01 & 16.42 & 64.28 \\
    & \checkmark & \checkmark & \checkmark & \textbf{10.80} & \textbf{13.12} & \textbf{15.47} & \textbf{17.79} & \textbf{16.13} & \underline{61.98} \\
    \hline 
    \multirow{2}{*}{MAE-B16} &  &  &  & 12.84 & 15.02 & 17.01 & 18.66 & 17.36 & 69.28 \\
    & \checkmark & \checkmark & \checkmark & 11.14 & \underline{13.27} & 15.63 & \underline{17.83} & \underline{16.18} & 63.57 \\
    \hline 
    \multirow{2}{*}{SAM-B16} &  &  &  & 12.60 & 14.72 & 16.87 & 18.65 & 17.21 & 68.01 \\
    & \checkmark & \checkmark & \checkmark & \underline{11.05} & 13.38 & \underline{15.61} & 17.89 & 16.30 & 62.93 \\
    \hline 
    \multirow{2}{*}{IJEPA-H16} &  &  &  & 13.33 & 15.57 & 17.41 & 18.84 & 17.68 & 70.95 \\
    & \checkmark & \checkmark & \checkmark & 11.24 & 13.39 & 15.65 & 17.90 & 16.29 & \textbf{61.90} \\
    \bottomrule
    \end{tabular}
    }
    \label{tab:mcubes_results}
\end{table*}

\subsection{Integrating MMFuser into SkipClick}
\label{sec:mm_skipclick}
\begin{figure*}
    \centering
    \includegraphics[width=0.85\linewidth]{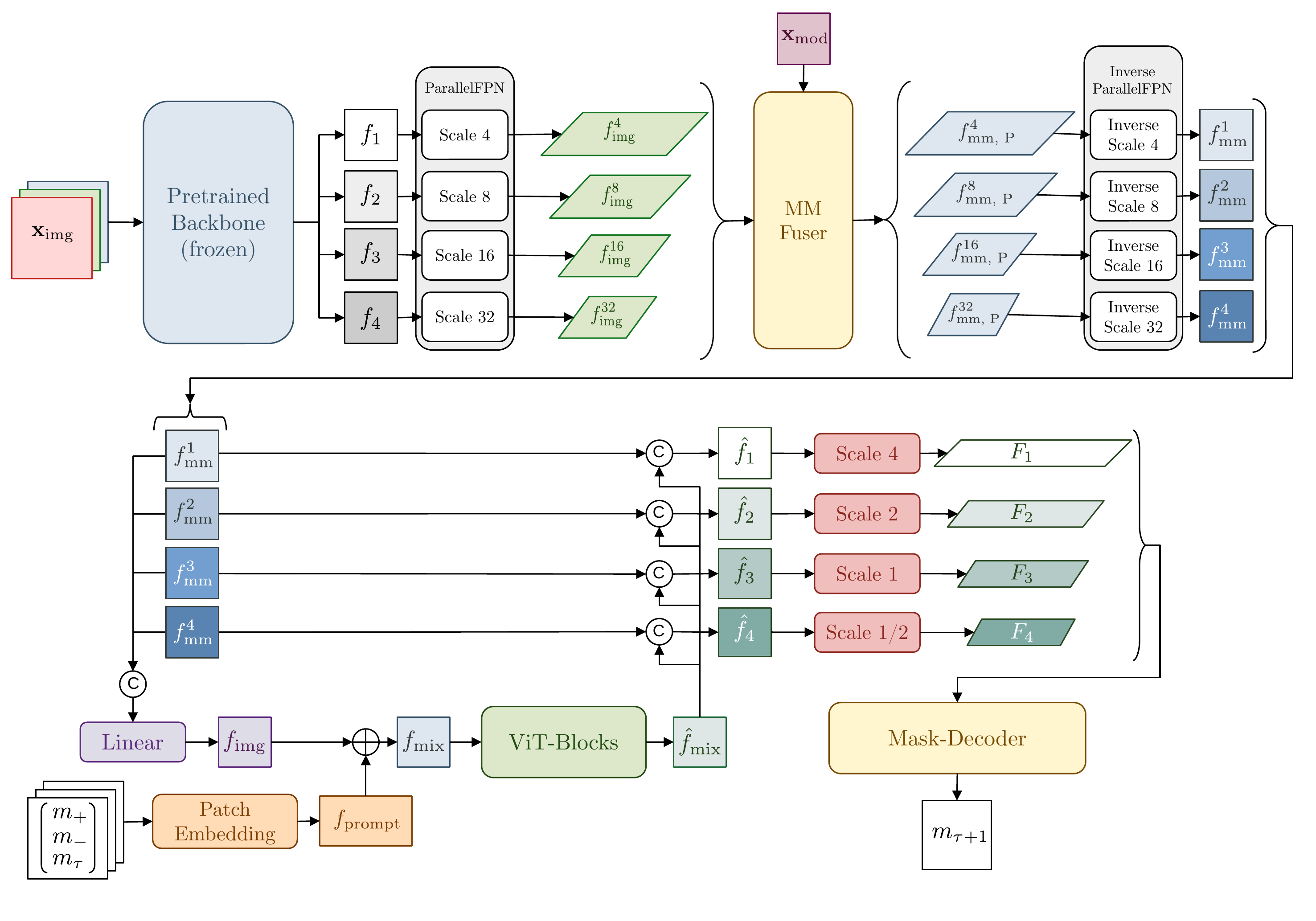}
    \caption{An extension of the SkipClick architecture to also make use of the MMFuser module. This figure is an extended version of Figure 2b from \cite{schoen2025skipclick}. }
    \label{fig:skipclick_mmfuser}
\end{figure*}

In this section, we briefly describe how we integrate MMFuser into the SkipClick architecture. As a complete description of SkipClick would be far beyond the scope of this text, we refer the interested reader to \cite{schoen2025skipclick} for further details. However, we do illustrate the altered version of SkipClick in Figure \ref{fig:skipclick_mmfuser}. 

The MMFuser is integrated after the backbone of the SkipClick architecture. As SkipClick uses a DINOv2-B14 as a backbone, the patch size is $14 \times 14$ and $d_\text{model} = 768$. Let  
\begin{equation}
    f_1, f_2, f_3, f_4 \in \mathbb{R}^{\frac{H}{14}\times \frac{W}{14} \times 768}
\end{equation}
be the feature tensors extracted from the intermediate layers of the backbone. 
We first use a ParallelFPN to convert the features $(f_1, f_2, f_3, f_4)$ to a feature pyramid  
\begin{equation}
    \left( f^{4}_\text{img}, f^{8}_\text{img}, f^{16}_\text{img}, f^{32}_\text{img} \right) = \text{ParallelFPN}(f_1, f_2, f_3, f_4). 
\end{equation}
We now have a feature pyramid to which we can apply the MMFuser. We give $\mathbf{x}_\text{mod}$ and the feature pyramid to MMFuser and obtain
\begin{equation}
\begin{split}
    \left( f^{4}_\text{mm, P}, f^{8}_\text{mm, P}, f^{16}_\text{mm, P}, f^{32}_\text{mm, P} \right) = \\ 
    \text{MMFuser} \left(\left( f^{4}_\text{img}, f^{8}_\text{img}, f^{16}_\text{img}, f^{32}_\text{img} \right), \mathbf{x}_\text{mod} \right).
\end{split}
\end{equation}
The remaining parts of the SkipClick architecture are designed to operate on tensors in $\mathbb{R}^{\frac{H}{14}\times \frac{W}{14} \times 768}$. Thus, we need to reverse the changes in the tensor shape again. To do so, we apply a second ParallelFPN to which we refer to as \emph{InverseParallelFPN}. 
The network does not actually constitute the mathematically inverse operation of the regular ParallelFPN. 
Instead, it consists of multiple scaling modules that each change the shape of a particular stage in our feature pyramid to the original shape of the tensor that has been extracted from the backbone. Similar to the ParallelFPN, each scaling module is based on convolutions, up-/downsampling mechanisms, layer normalizations and GELU activations. A detailed description can be found in the supplementary material in section \ref{sec:parallel_fpn}. We have 
\begin{equation}
\begin{split}
    \left( f^1_\text{mm}, f^2_\text{mm}, f^3_\text{mm}, f^4_\text{mm} \right) = \\
    \text{InverseParallelFPN}\left( f^{4}_\text{mm, P}, f^{8}_\text{mm, P},  f^{16}_\text{mm, P}, f^{32}_\text{mm, P} \right).
\end{split}
\end{equation}
As the shape has been restored to that of the backbones output, we have 
\begin{equation}
    f^1_\text{mm}, f^2_\text{mm}, f^3_\text{mm}, f^4_\text{mm} \in \mathbb{R}^{\frac{H}{14}\times \frac{W}{14} \times 768}. 
\end{equation}
From this stage onward, the remaining part of the architecture stays identical to SkipClick. While Figure \ref{fig:skipclick_mmfuser} provides an overview of the overall architecture, we refrain from a detailed description and refer the interested reader to \cite{schoen2025skipclick}. 
Finally, it might be interesting to note that the reason we do not have an extra SkipClick-based backbone in our experiments is that SkipClick uses an unaltered DINOv2-B14 backbone as well.

\subsection{ParallelFPN and InverseParallelFPN}
\label{sec:parallel_fpn}
In this section, we describe the \emph{ParallelFPN} used in the main paper. We also describe the \emph{InverseParallelFPN} that is exclusively used when integrating our MMFuser module into SkipClick. It should be noted that our ParallelFPN is inspired by the SimpleFPN used in \cite{Liu_2023_ICCV}. 
Each of the two models consists of four different scaling modules, which are called \emph{Scale $i$} and \emph{Inverse Scale $i$} for $i = 4, 8, 16, 32$. 
Let $d_\text{FM}$ and $P_\text{FM}$ be the internal dimensionality of the ViT-based foundation model and its patch size, respectively. If the original images have the shape $H \times W$, the various tensors of our feature pyramid will have the shape $\frac{H}{i} \times \frac{W}{i}$, and $d_\text{embed}^i$ will be the respective channel dimension. In our case $d_\text{embed}^{4} = 64, d_\text{embed}^{8} = 128, d_\text{embed}^{16} = 320, d_\text{embed}^{32} = 512$. 

The shape of the scaling modules' input and output tensors changes in the following way: 
\begin{itemize}
    \item The module \emph{Scale $i$} transforms a feature tensor from the ViT-like representation shape $\mathbb{R}^{\frac{H}{P_\text{FM}} \times \frac{W}{P_\text{FM}} \times d_\text{FM}}$ to a feature pyramid shape $\mathbb{R}^{\frac{H}{i} \times \frac{W}{i} \times d_\text{embed}^i}$. 
    \item The module \emph{Inverse Scale $i$} has the inverse effect on the shape. It transforms a feature tensor from $\mathbb{R}^{\frac{H}{i} \times \frac{W}{i} \times d_\text{embed}^i}$ to $\mathbb{R}^{\frac{H}{P_\text{FM}} \times \frac{W}{P_\text{FM}} \times d_\text{FM}}$.   
\end{itemize}
Each of the modules is a feed-forward network, i.e. all of its layers are executed in succession. The architecture of the used modules can be found in Table \ref{tab:parallel_fpn}. Each of the table cells contains the submodules of a particular scaling module. 
Within each scaling module the  submodules are executed in the order going from top of the table cell to its bottom. It should be noted that the LayerNormalizations have been implemented as GroupNormalizations with a single group to allow for convenient usage of the PyTorch library.
The description in the table also makes use of hidden dimensions, which can be computed as follows: 
\begin{equation}
    d_\text{hidden}^4 = \max \left\{2 \cdot d_\text{embed}^4, \frac{d_\text{FM}}{2} \right\} 
\end{equation}
\begin{equation}
    d_\text{hidden}^8 = \max \left\{d_\text{embed}^8, \frac{d_\text{FM}}{2} \right\}
\end{equation}
\begin{equation}
    d_\text{hidden}^{16} = \max \left\{d_\text{embed}^{16}, d_\text{FM} \right\}
\end{equation}
\begin{equation}
    d_\text{hidden}^{32} = \max \left\{d_\text{embed}^{32}, 2 \cdot d_\text{FM} \right\}
\end{equation}

The hyperparameters given in Table \ref{tab:parallel_fpn} are to be interpreted as follows: 
\begin{itemize}
    \item A convolutional layer $\text{Conv}(d_\text{in}, d_\text{out}, k, s)$ receives $d_\text{in}$ input channels, produces $d_\text{out}$ output channels, has a kernel size of $k \times k$ and a stride of $s$. 
    \item An interpolation layer $\text{Interpolate}(h, w)$ resizes the height and width axes of a tensor to $(h, w)$ by using bilinear interpolation.
    \item $\text{LayerNormalization}(d)$ carries out a layer normalization on a tensor with $d$ feature channels. 
    \item The activation function GELU \cite{hendrycks2016gelu} has no hyperparameters.
\end{itemize}

\begin{table*}[]
    \centering
    \caption{The submodules of the scaling modules in the ParallelFPN and the InverseParallelFPN. Each of the table cells contains the submodules of a particular scaling module. Within each scaling module the  submodules are executed consecutively in the order going from top of the cell to its bottom. Any hyperparameters a submodule might need are written in the parentheses next to it. }
    \begin{tabular}{c|c|c}
        \toprule
        \textbf{Submodule} & \textbf{ParallelFPN} & \textbf{InverseParallelFPN} \\
        \midrule
        \multirow{10}{*}{Scale 4} & $\text{Interpolate}\left(\frac{H}{8}, \frac{W}{8} \right)$ & $\text{Interpolate}\left(\frac{H}{8}, \frac{W}{8} \right)$ \\
        & $\text{Conv}\left(d_\text{FM}, d_\text{hidden}^4, k=3, s=1 \right)$ & $\text{Conv}\left(d_\text{embed}^1, \frac{d_\text{hidden}^4}{2}, k=3, s=1 \right)$ \\
        & $\text{LayerNormalization}\left( d_\text{hidden}^4 \right)$ & $\text{LayerNormalization}\left( \frac{d_\text{hidden}^4}{2} \right)$ \\ 
        & GELU & GELU \\ 
        & $\text{Interpolate}\left(\frac{H}{4}, \frac{W}{4} \right)$ & $\text{Interpolate}\left(\frac{H}{P_\text{FM}}, \frac{W}{P_\text{FM}} \right)$ \\ 
        & $\text{Conv}\left(d_\text{hidden}^4, \frac{d_\text{hidden}^4}{2}, k=3, s=1 \right)$ & $\text{Conv}\left(\frac{d_\text{hidden}^4}{2}, d_\text{hidden}^4, k=3, s=1 \right)$  \\ 
        & $\text{LayerNormalization}\left( \frac{d_\text{hidden}^4}{2} \right)$ & $\text{LayerNormalization}\left( d_\text{hidden}^4 \right)$ \\ 
        & $\text{Conv}\left(d_\text{hidden}^4, d_\text{embed}^1, k=1, s=1 \right)$ & $\text{Conv}\left(d_\text{hidden}^4, d_\text{FM}, k=1, s=1 \right)$  \\ 
        & $\text{LayerNormalization}\left( d_\text{embed}^4 \right)$ & $\text{LayerNormalization}\left( d_\text{FM} \right)$ \\ 
        & GELU & GELU \\ 
        \hline

        \multirow{6}{*}{Scale 8} & $\text{Interpolate}\left(\frac{H}{8}, \frac{W}{8} \right)$  &  $\text{Interpolate}\left(\frac{H}{P_\text{FM}}, \frac{W}{P_\text{FM}} \right)$ \\ 
        & $\text{Conv}\left(d_\text{FM}, d_\text{hidden}^8, k=3, s=1 \right)$ & $\text{Conv}\left(d_\text{embed}^8, d_\text{hidden}^8, k=3, s=1 \right)$ \\ 
        & $\text{LayerNormalization}\left( d_\text{hidden}^8 \right)$ & $\text{LayerNormalization}\left( d_\text{hidden}^8 \right)$ \\ 
        & $\text{Conv}\left(d_\text{hidden}^8, d_\text{embed}^8, k=1, s=1 \right)$ & $\text{Conv}\left(d_\text{hidden}^8, d_\text{FM}, k=1, s=1 \right)$ \\ 
        & $\text{LayerNormalization}\left( d_\text{embed}^8 \right)$  & $\text{LayerNormalization}\left( d_\text{FM} \right)$ \\ 
        & GELU & GELU \\ 
        \hline

        \multirow{6}{*}{Scale 16} & $\text{Conv}\left(d_\text{FM}, d_\text{hidden}^{16}, k=3, s=1 \right)$ & $\text{Conv}\left(d_\text{embed}^{16}, d_\text{hidden}^{16}, k=3, s=1 \right)$ \\ 
        & $\text{Interpolate}\left(\frac{H}{16}, \frac{W}{16} \right)$  &  $\text{Interpolate}\left(\frac{H}{P_\text{FM}}, \frac{W}{P_\text{FM}} \right)$ \\ 
        & $\text{LayerNormalization}\left( d_\text{hidden}^{16} \right)$ & $\text{LayerNormalization}\left( d_\text{hidden}^{16} \right)$ \\ 
        & $\text{Conv}\left(d_\text{hidden}^{16}, d_\text{embed}^{16}, k=1, s=1 \right)$ & $\text{Conv}\left(d_\text{hidden}^{16}, d_\text{FM}, k=1, s=1 \right)$ \\ 
        & $\text{LayerNormalization}\left( d_\text{embed}^{16} \right)$  & $\text{LayerNormalization}\left( d_\text{FM} \right)$ \\ 
        & GELU & GELU \\ 
        \hline
        
        \multirow{6}{*}{Scale 32} & $\text{Conv}\left(d_\text{FM}, d_\text{hidden}^{32}, k=3, s=1 \right)$ &  $\text{Interpolate}\left(\frac{H}{P_\text{FM}}, \frac{W}{P_\text{FM}} \right)$\\ 
        & $\text{Interpolate}\left(\frac{H}{32}, \frac{W}{32} \right)$  & $\text{Conv}\left(d_\text{embed}^{32}, d_\text{hidden}^{32}, k=3, s=1 \right)$  \\ 
        & $\text{LayerNormalization}\left( d_\text{hidden}^{32} \right)$ & $\text{LayerNormalization}\left( d_\text{hidden}^{32} \right)$ \\ 
        & $\text{Conv}\left(d_\text{hidden}^{32}, d_\text{embed}^{32}, k=1, s=1 \right)$ & $\text{Conv}\left(d_\text{hidden}^{32}, d_\text{FM}, k=1, s=1 \right)$ \\ 
        & $\text{LayerNormalization}\left( d_\text{embed}^{32} \right)$  & $\text{LayerNormalization}\left( d_\text{FM} \right)$ \\ 
        & GELU & GELU \\ 
        \bottomrule

    \end{tabular}
    \label{tab:parallel_fpn}
\end{table*}

\subsection{Qualitative Results}
\begin{figure*}
\centering
\newcommand{\fivemodwidth}{0.17} 
\begin{tabular}{ccccc}
\textbf{Groud Truth} & \textbf{Depth} & \textbf{Lidar} & \textbf{Event Camera} & \textbf{Prediction} \\
\includegraphics[width=\fivemodwidth\linewidth]{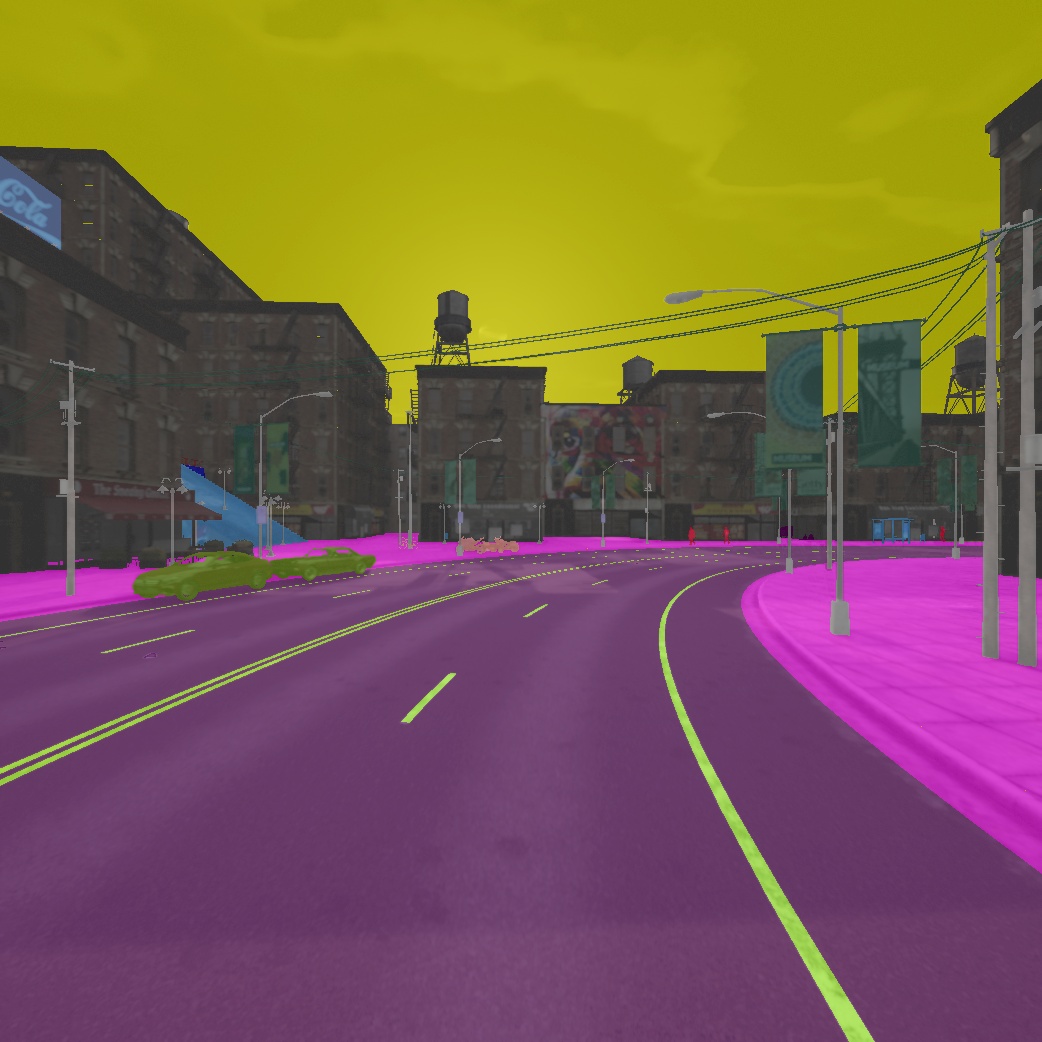} & 
\includegraphics[width=\fivemodwidth\linewidth]{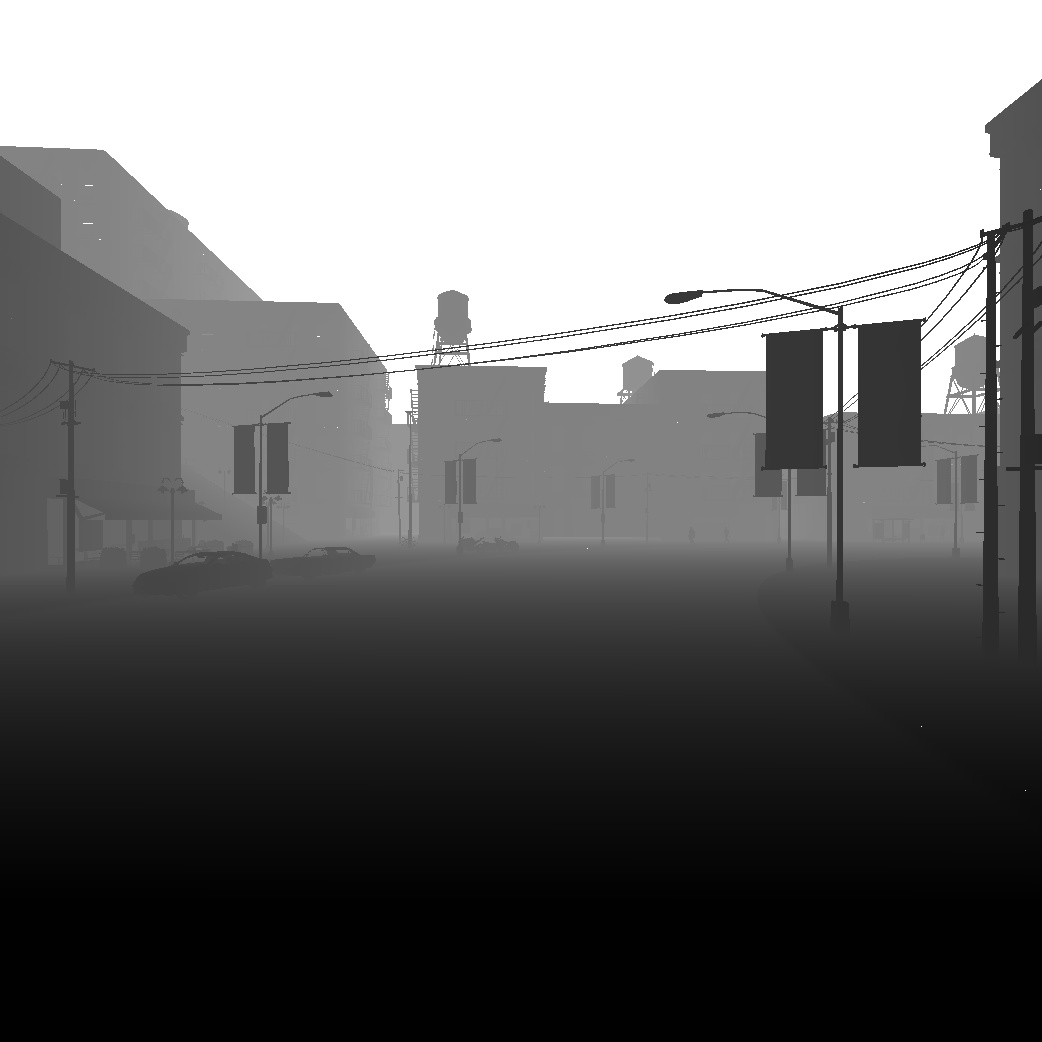} & 
\includegraphics[width=\fivemodwidth\linewidth]{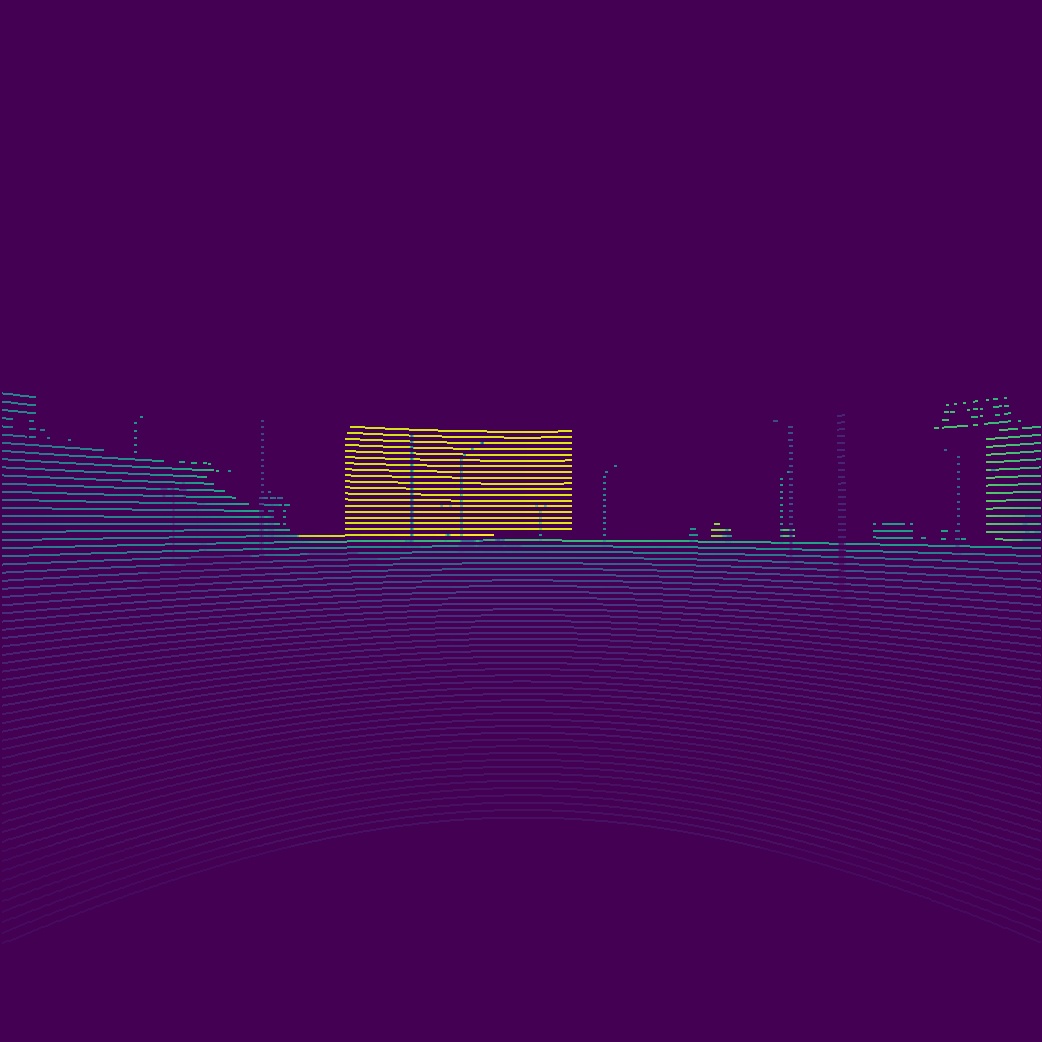} & 
\includegraphics[width=\fivemodwidth\linewidth]{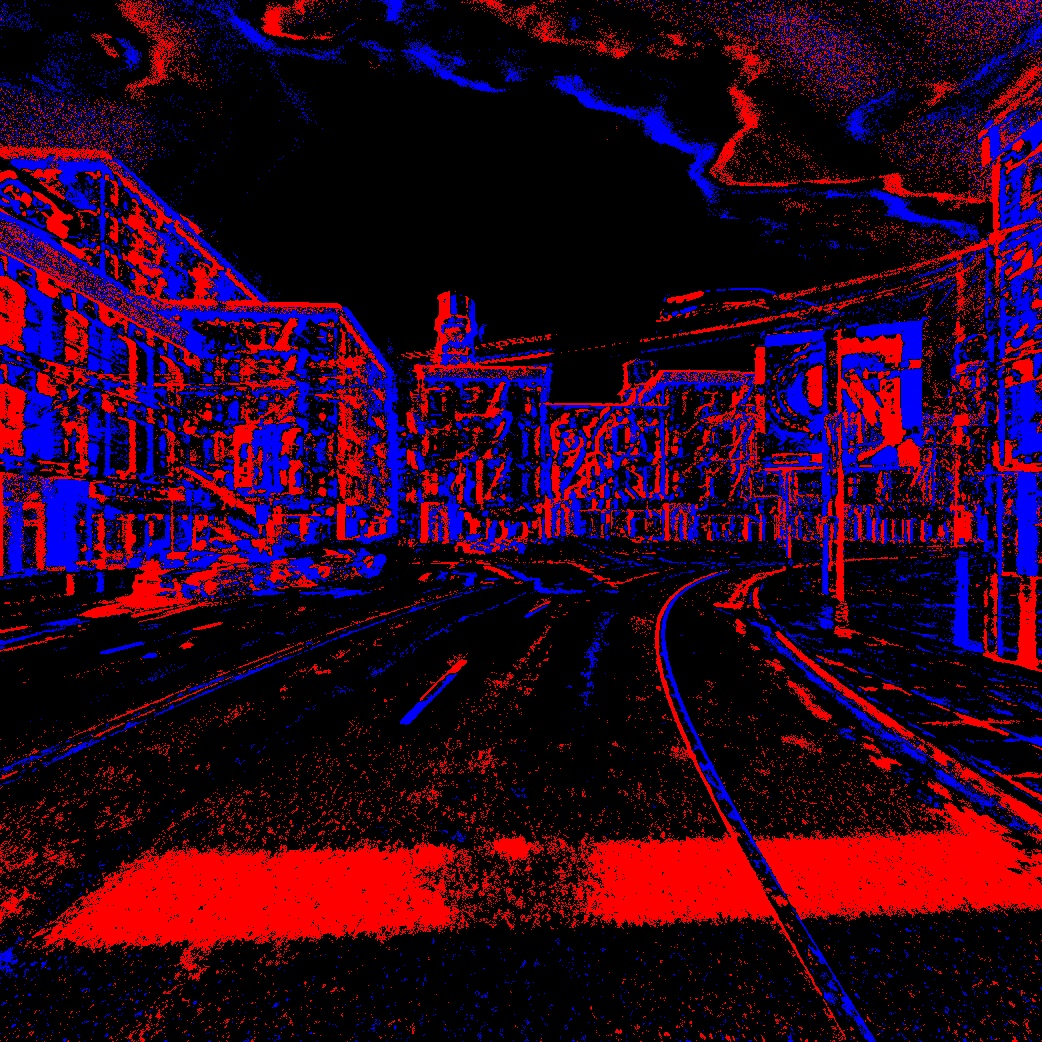} & 
\includegraphics[width=\fivemodwidth\linewidth]{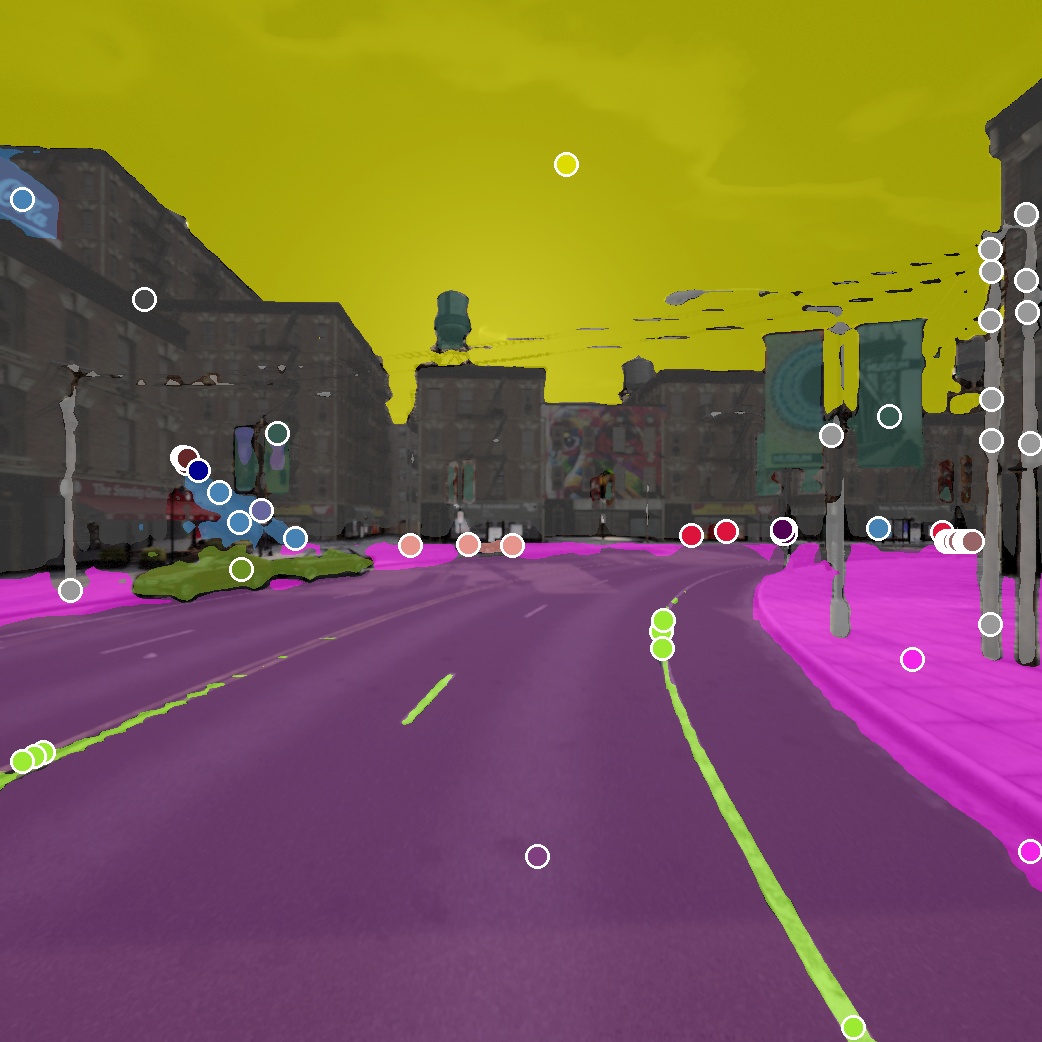} \\
\includegraphics[width=\fivemodwidth\linewidth]{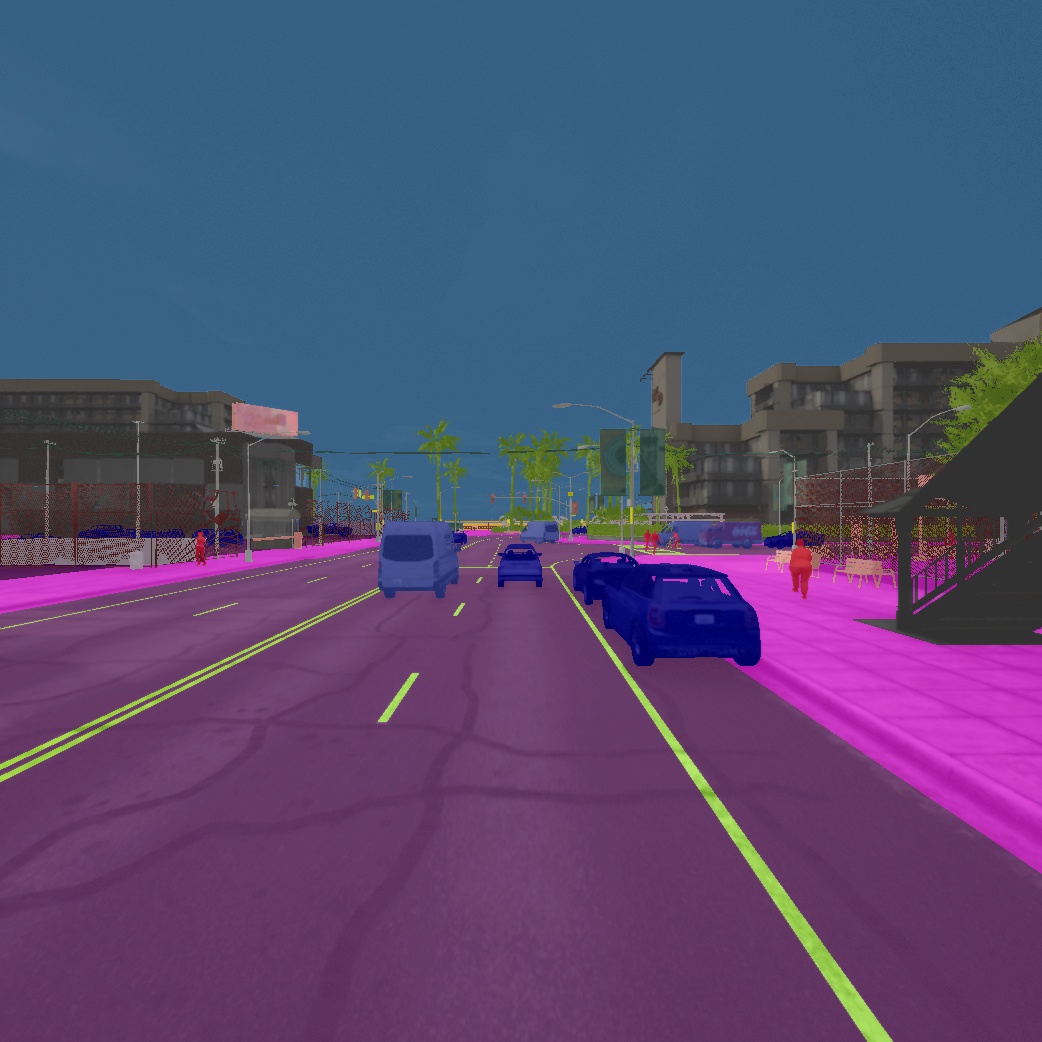} & 
\includegraphics[width=\fivemodwidth\linewidth]{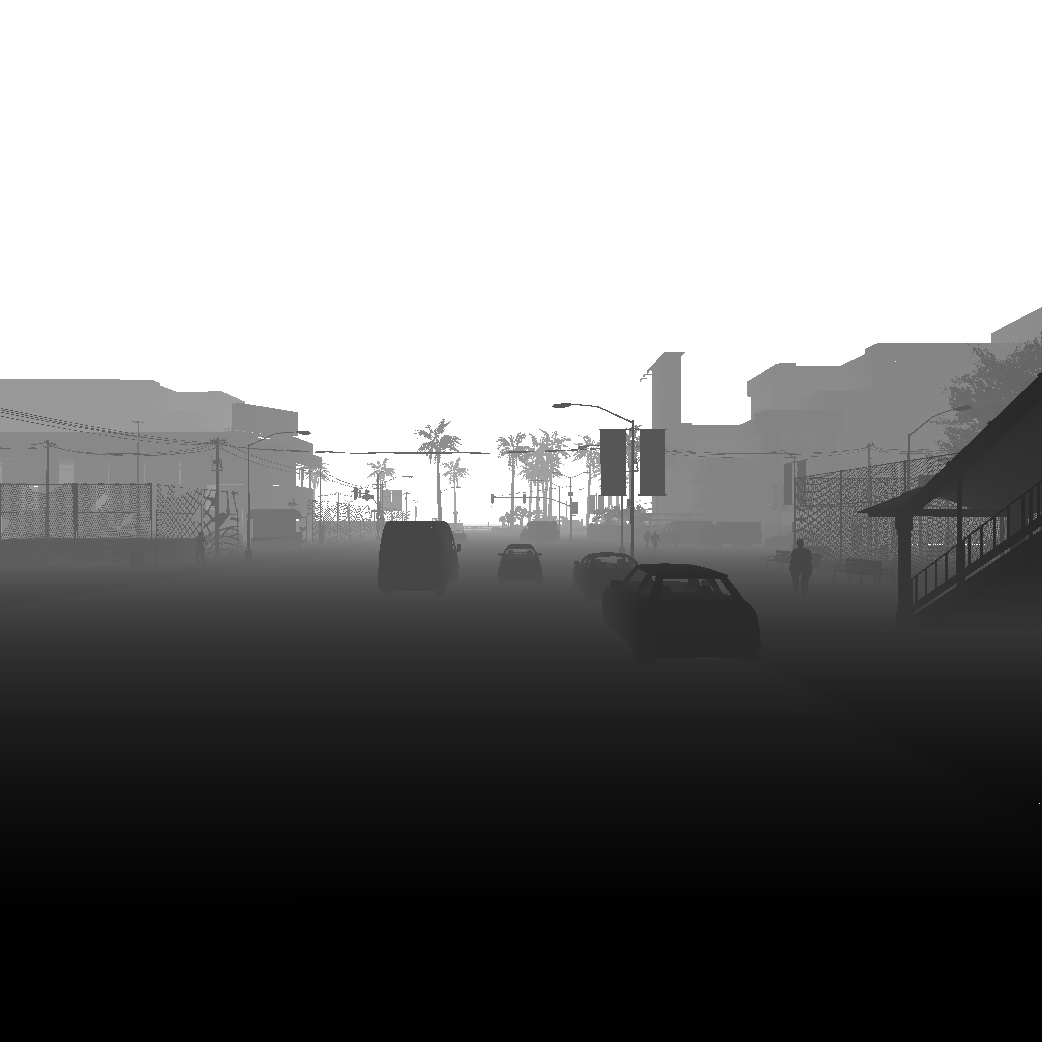} & 
\includegraphics[width=\fivemodwidth\linewidth]{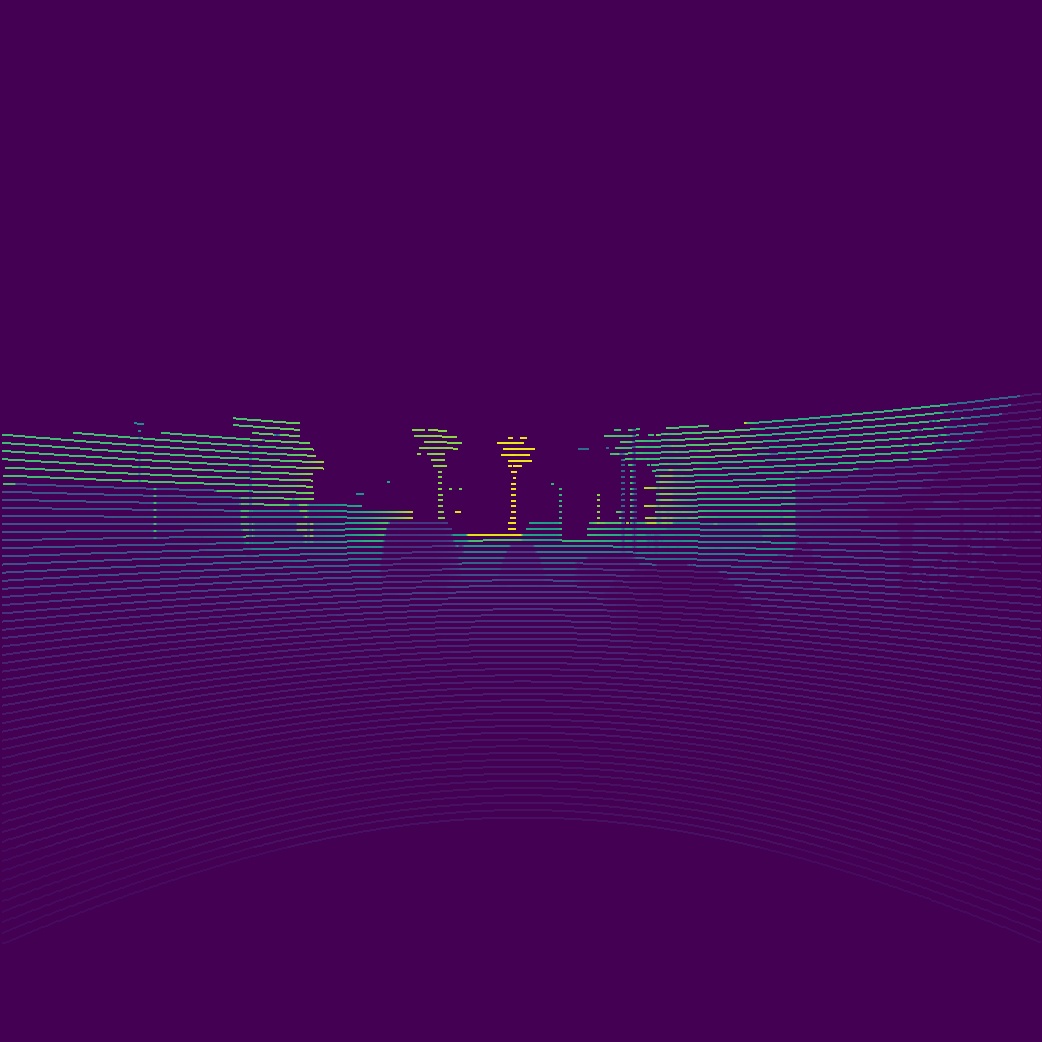} & 
\includegraphics[width=\fivemodwidth\linewidth]{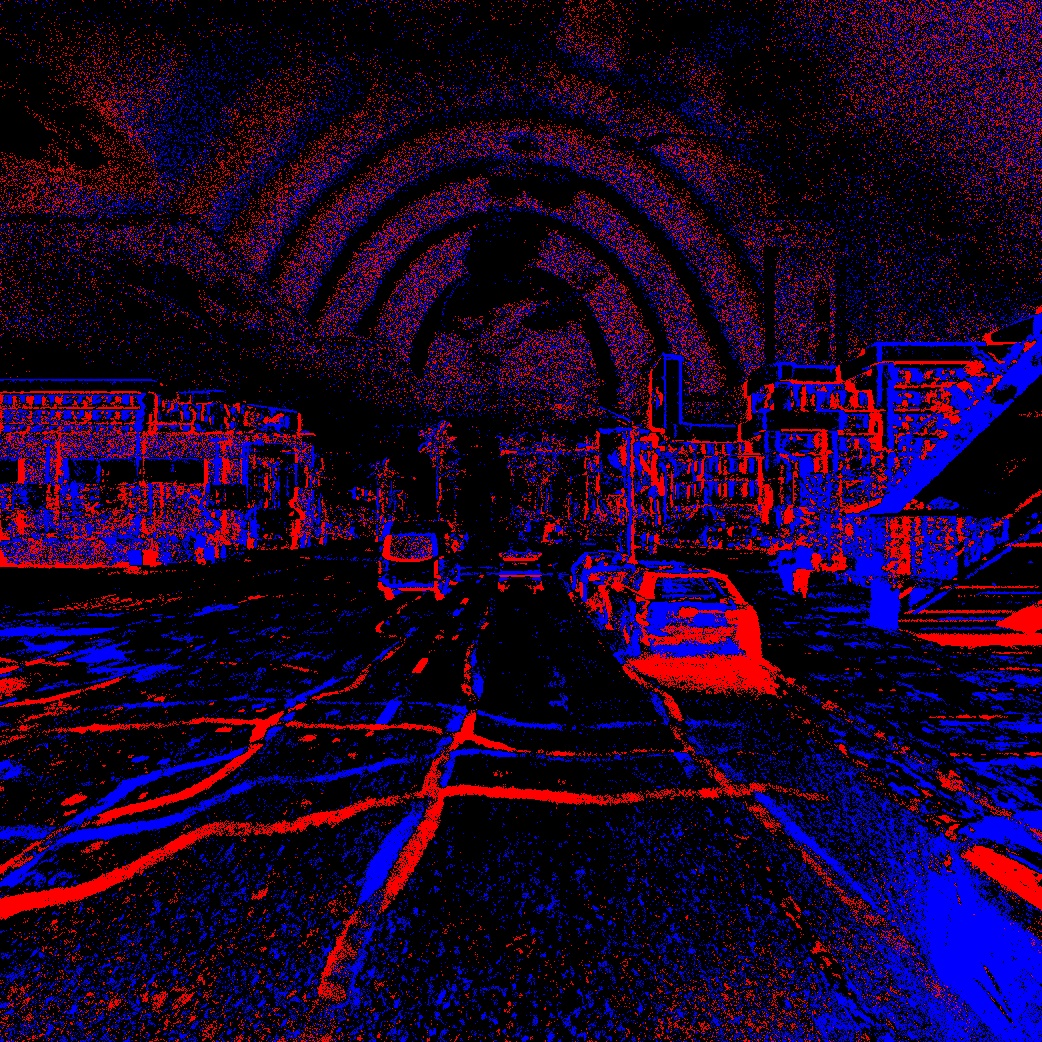} & 
\includegraphics[width=\fivemodwidth\linewidth]{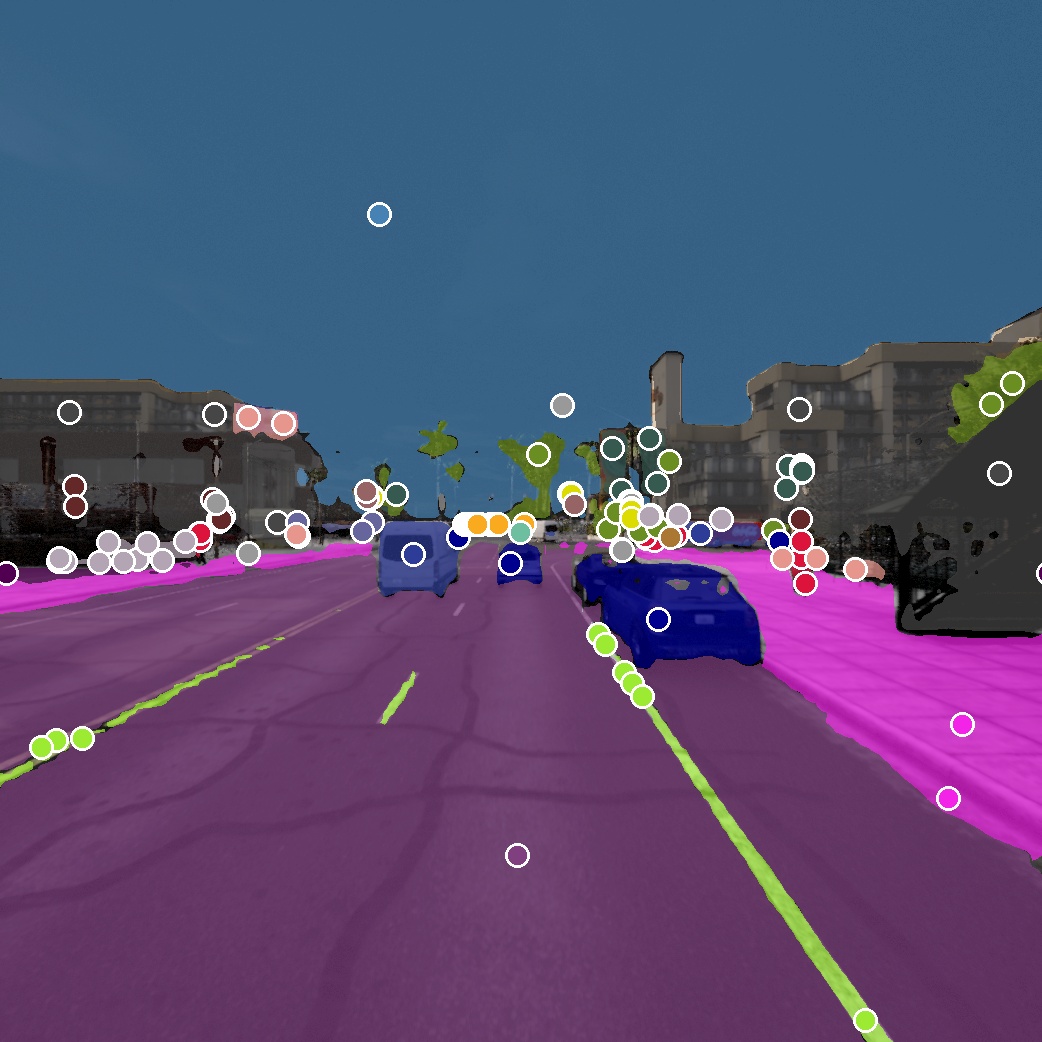} \\
\end{tabular}
\caption{Qualitative results on the DeLiVER dataset \cite{zhang2023delivering}. The model that generated the results is based DINOv2-B14 and has been given all three additional modalities: Depth, lidar, and event camera images. The lidar has been normalized and converted to the viridis color map to allow for better visualization. The first and last image in each row are the RGB image with the ground truth and the prediction plotted onto the image, respectively. } 

\label{fig:deliver_qualitative_app}
\end{figure*}

\begin{figure*}
\centering
\newcommand{\fivemodwidth}{0.17} 
\begin{tabular}{ccccc}
\textbf{Ground Truth} & \textbf{AoLP} & \textbf{DoLP} & \textbf{NIR} & \textbf{Prediction} \\
\includegraphics[width=\fivemodwidth\linewidth]{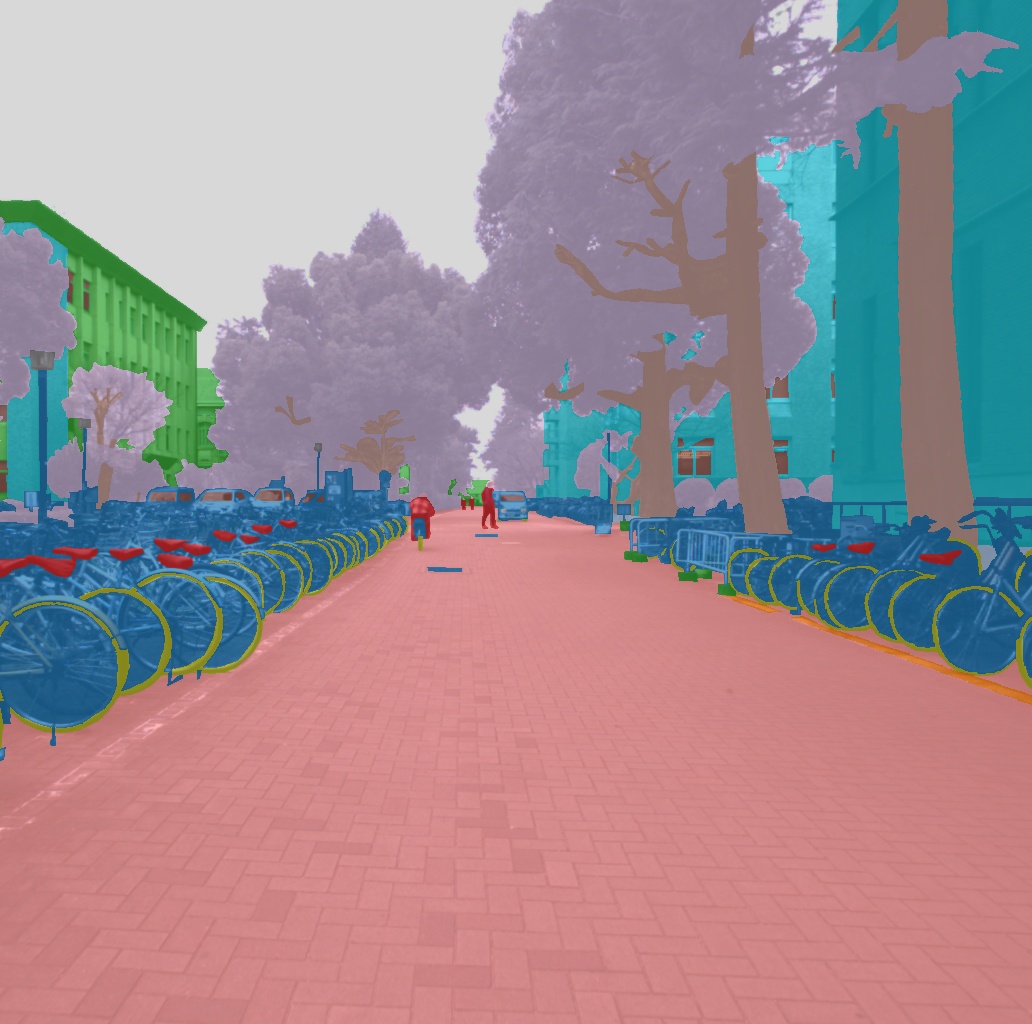} & 
\includegraphics[width=\fivemodwidth\linewidth]{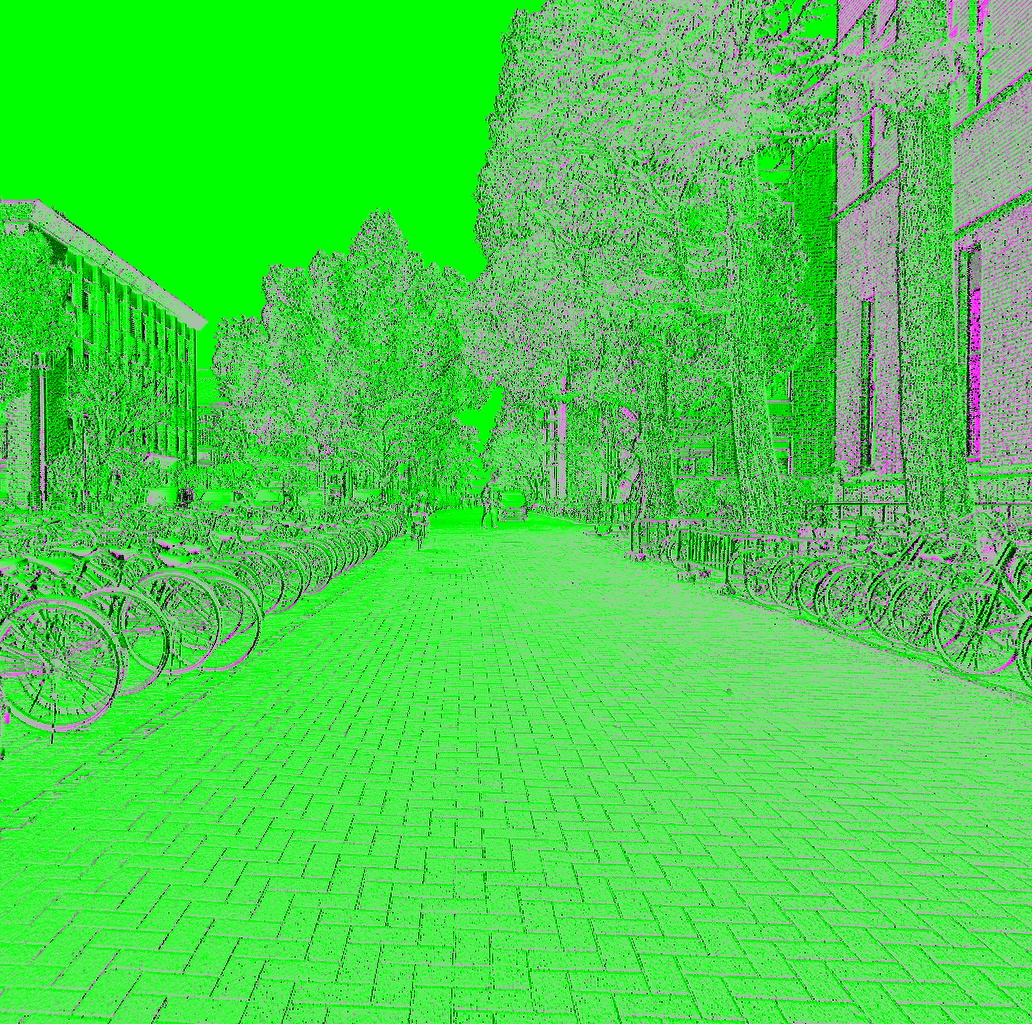} & 
\includegraphics[width=\fivemodwidth\linewidth]{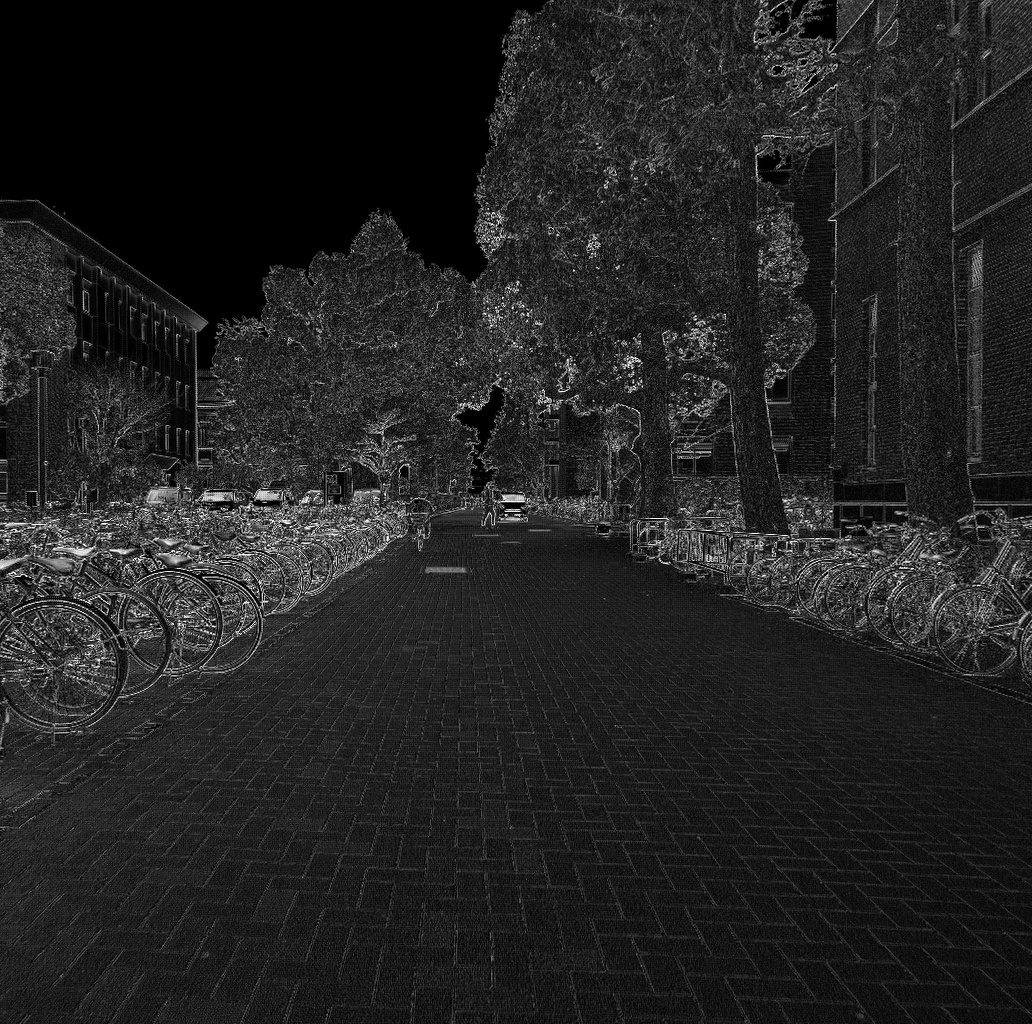} & 
\includegraphics[width=\fivemodwidth\linewidth]{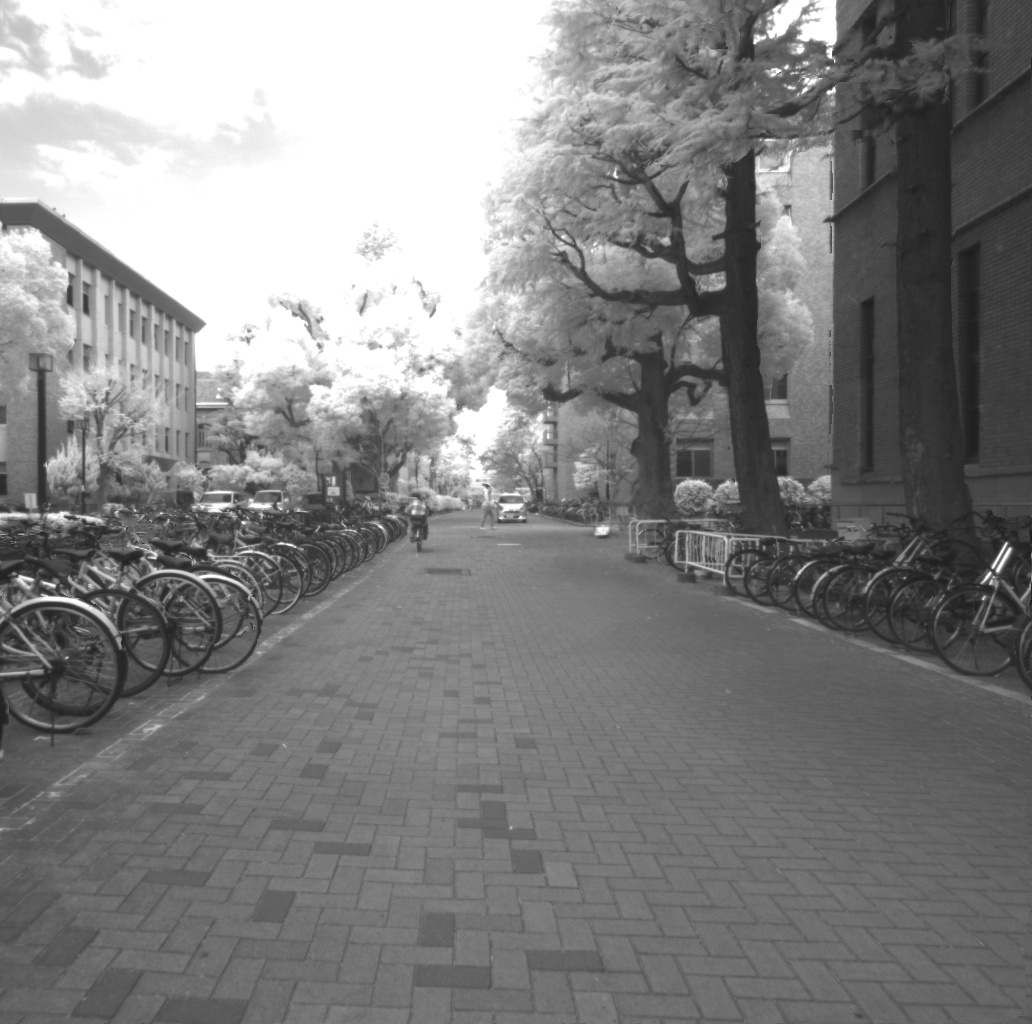} & 
\includegraphics[width=\fivemodwidth\linewidth]{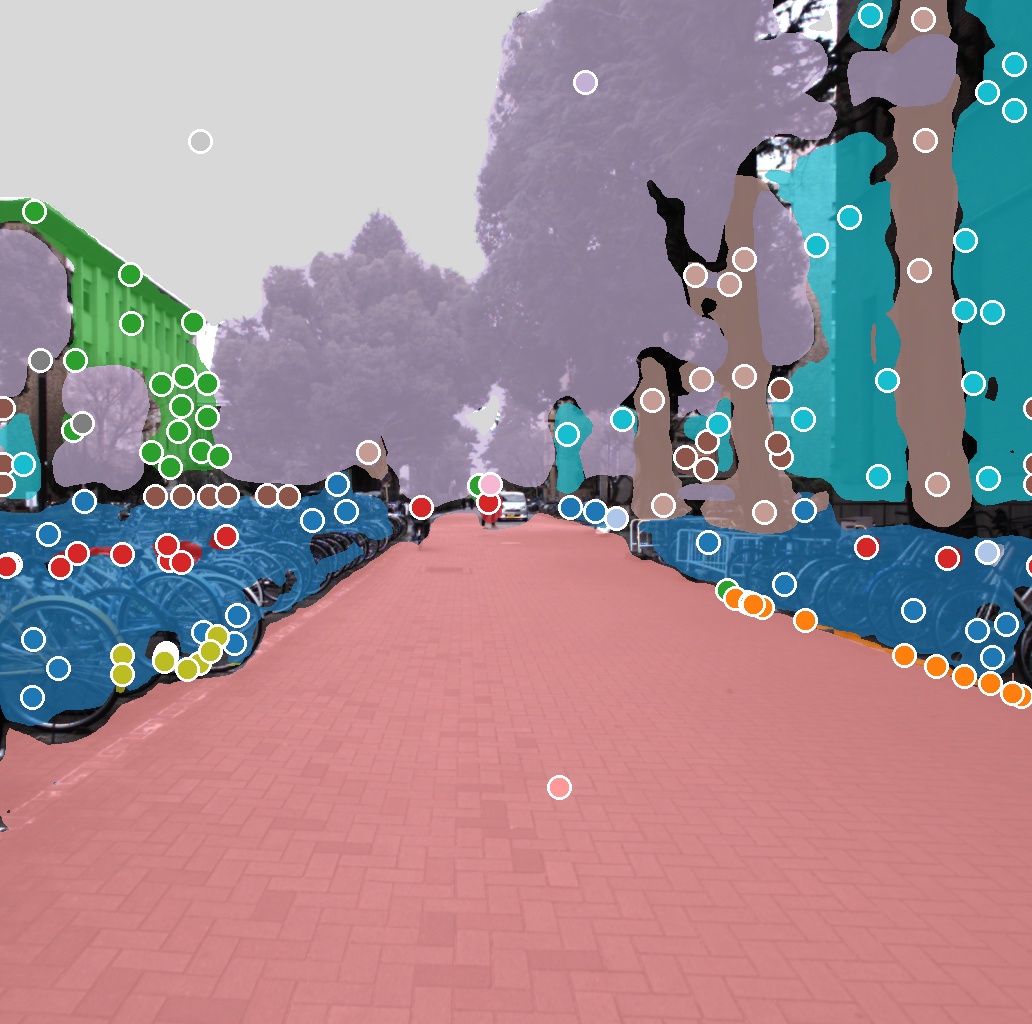} \\
\includegraphics[width=\fivemodwidth\linewidth]{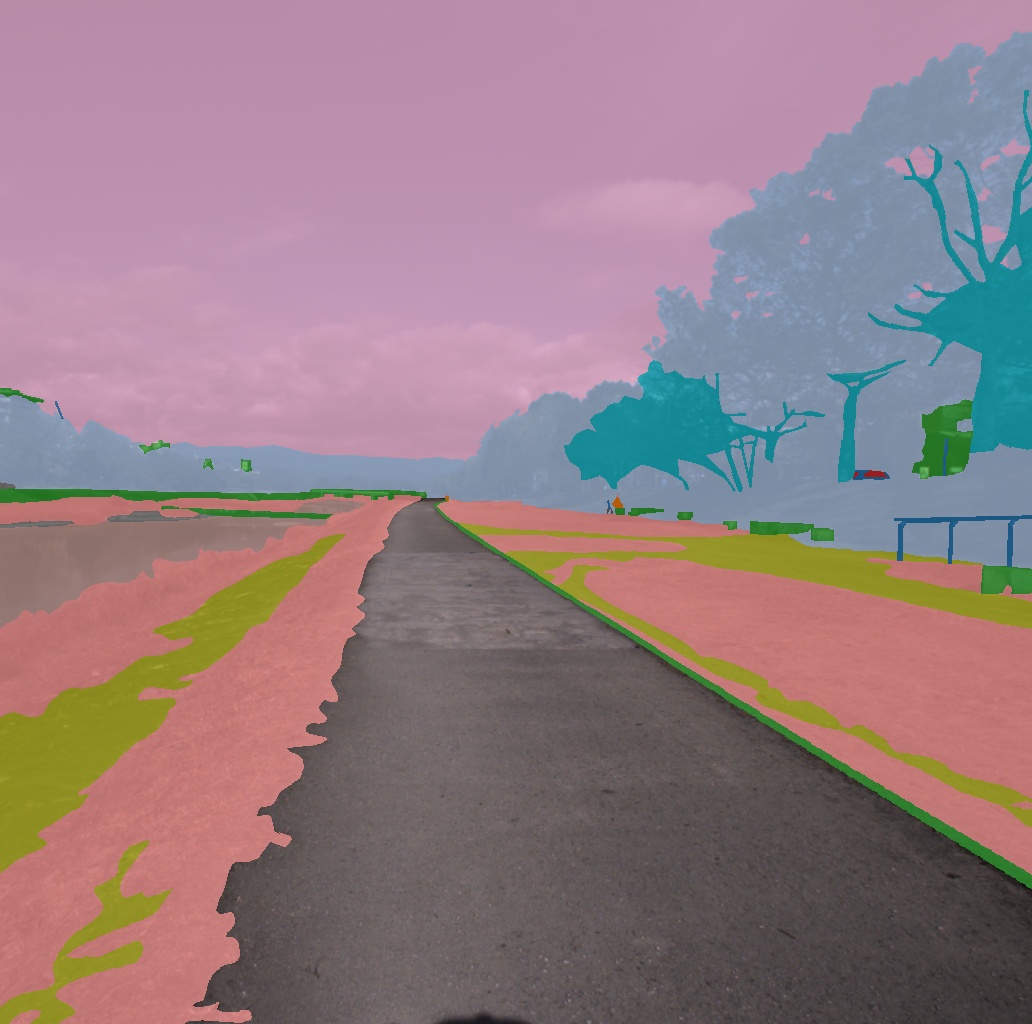} & 
\includegraphics[width=\fivemodwidth\linewidth]{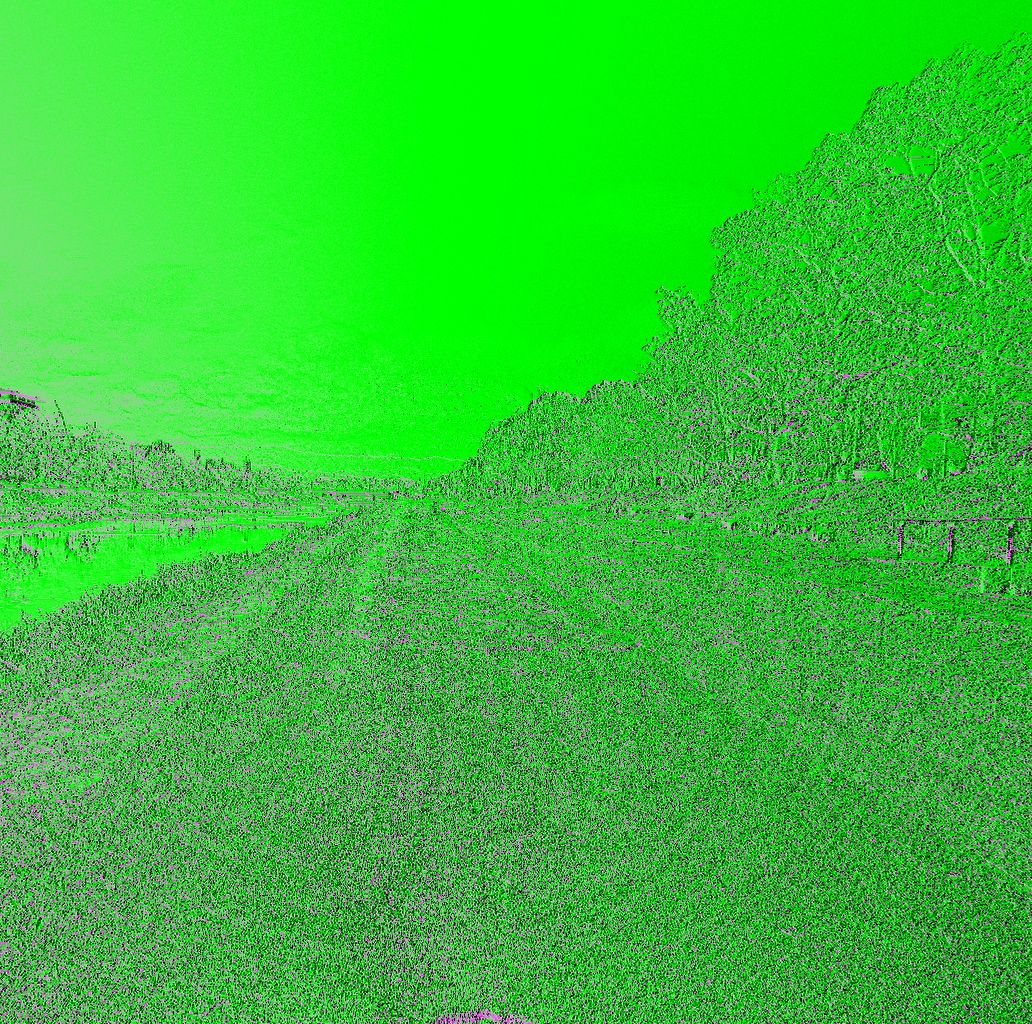} & 
\includegraphics[width=\fivemodwidth\linewidth]{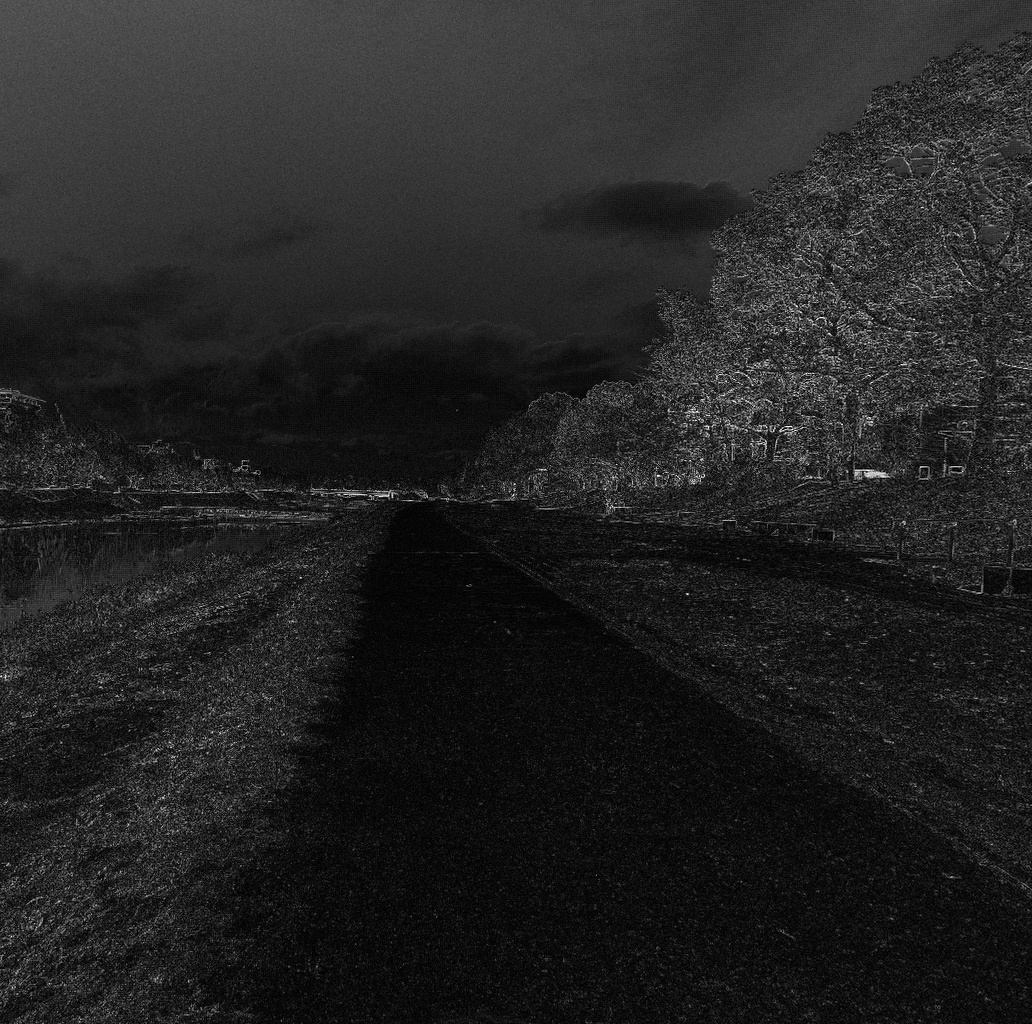} & 
\includegraphics[width=\fivemodwidth\linewidth]{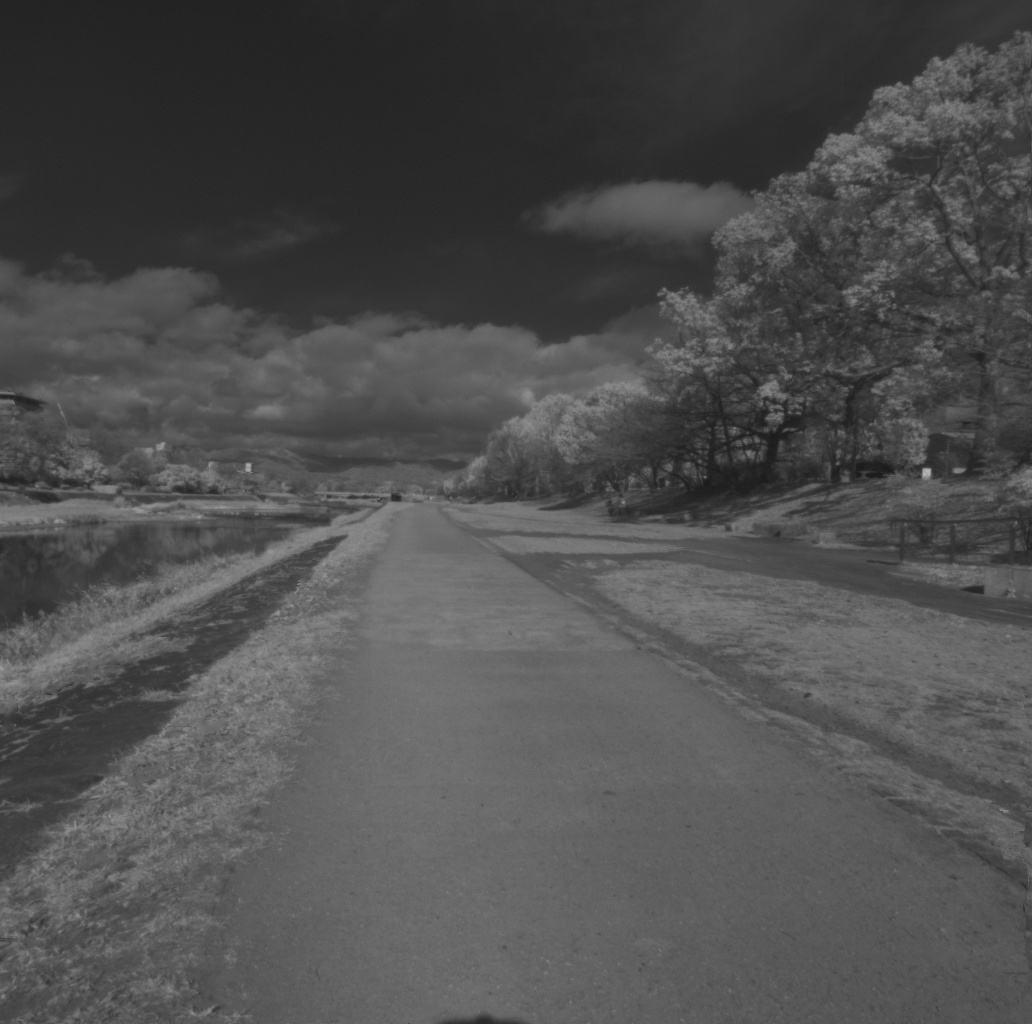} & 
\includegraphics[width=\fivemodwidth\linewidth]{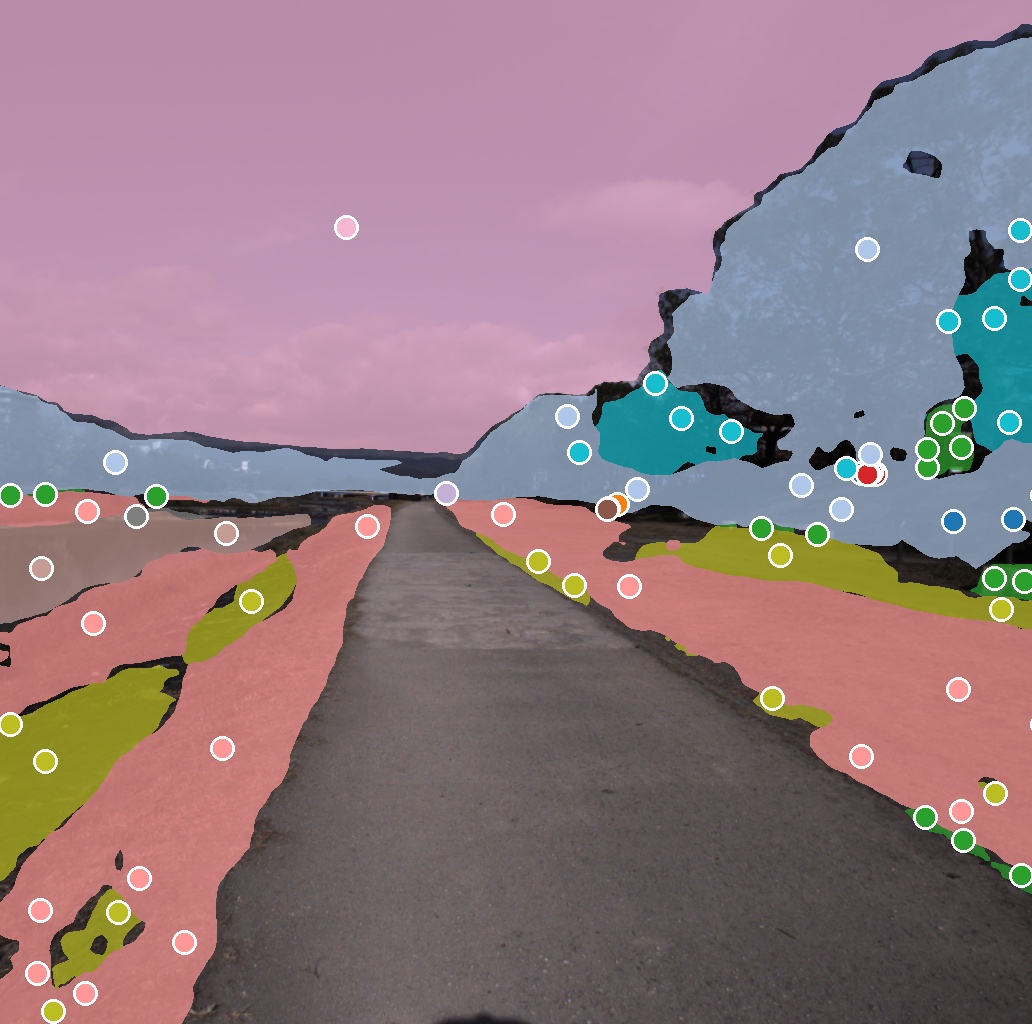} \\
\end{tabular}
\caption{Qualitative results on the MCubeS dataset \cite{Liang_2022_CVPR}. The model that generated the results is based DINOv2-B14. We use \emph{Angle of Linear Polarization (AoLP)}, \emph{Degree of Linear Polarization (DoLP)} and \emph{Near Infrared (NIR)} images as additional modalities. The first and last image in each row are the RGB image with the ground truth and the prediction plotted onto the image, respectively. }

\label{fig:mcubes_qualitative_app}
\end{figure*}

\begin{figure*}
\centering
\newcommand{\threemodwidth}{0.3} 
\begin{tabular}{ccc}
\textbf{Ground Truth} & \textbf{Infrared} & \textbf{Prediction} \\
\includegraphics[width=\threemodwidth\linewidth]{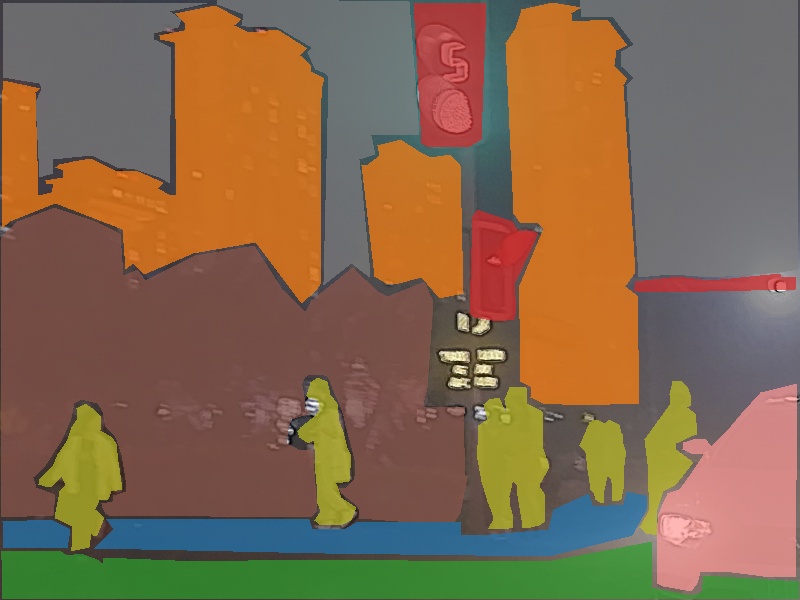} & 
\includegraphics[width=\threemodwidth\linewidth]{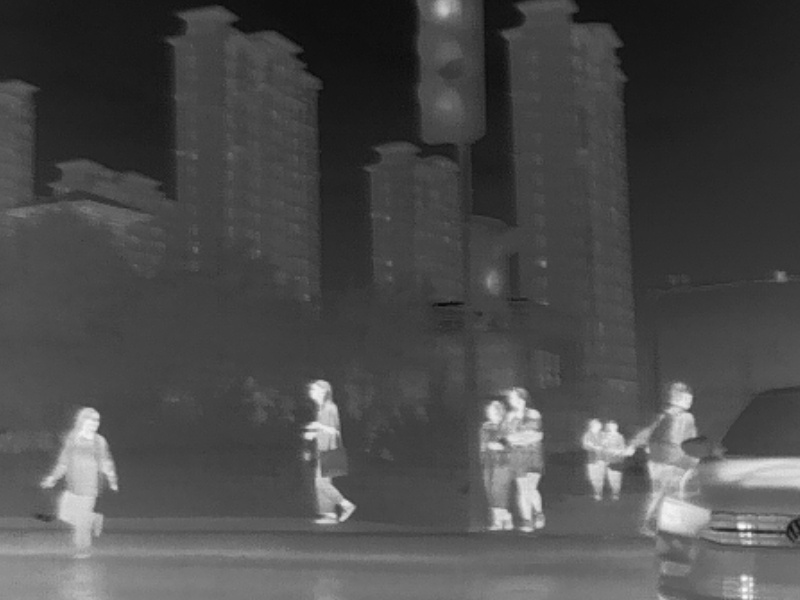} & 
\includegraphics[width=\threemodwidth\linewidth]{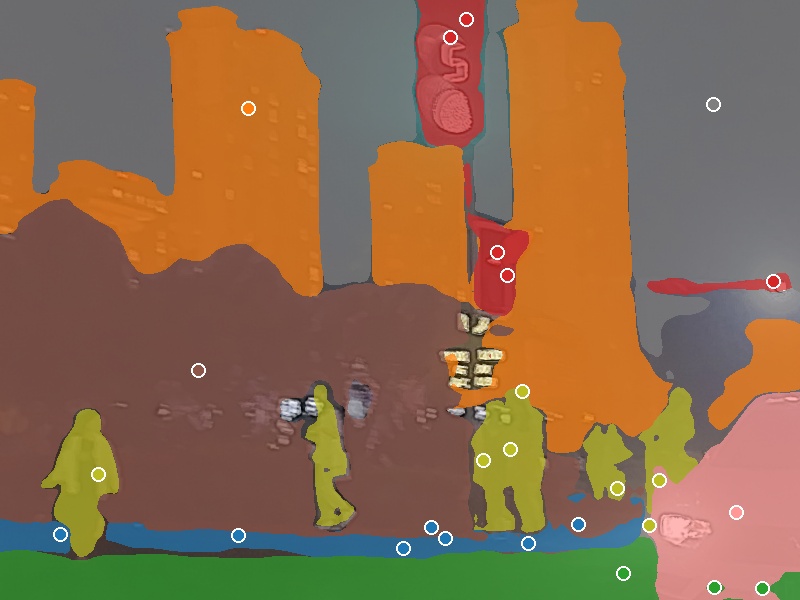} \\
\includegraphics[width=\threemodwidth\linewidth]{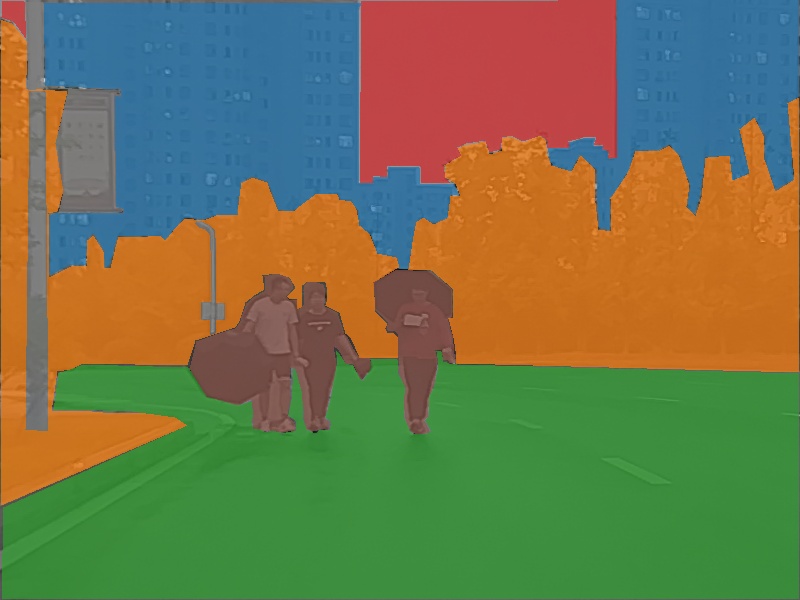} & 
\includegraphics[width=\threemodwidth\linewidth]{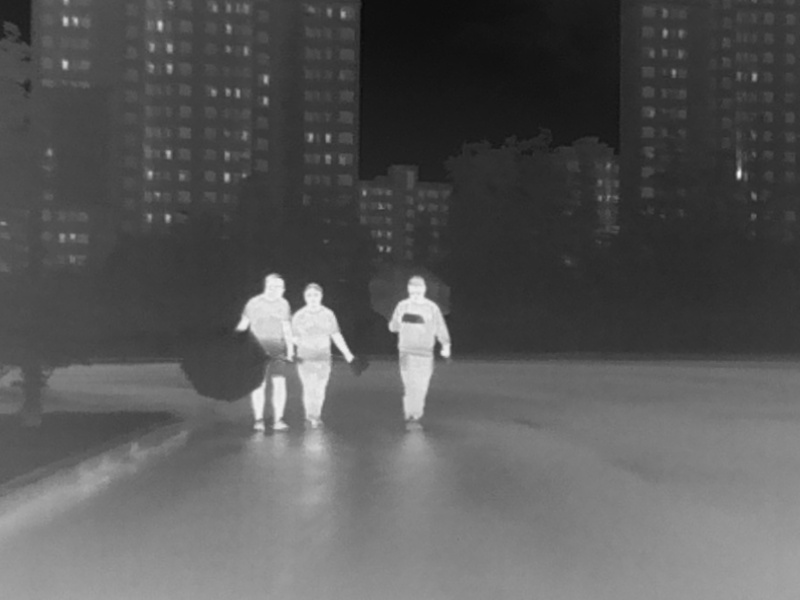} & 
\includegraphics[width=\threemodwidth\linewidth]{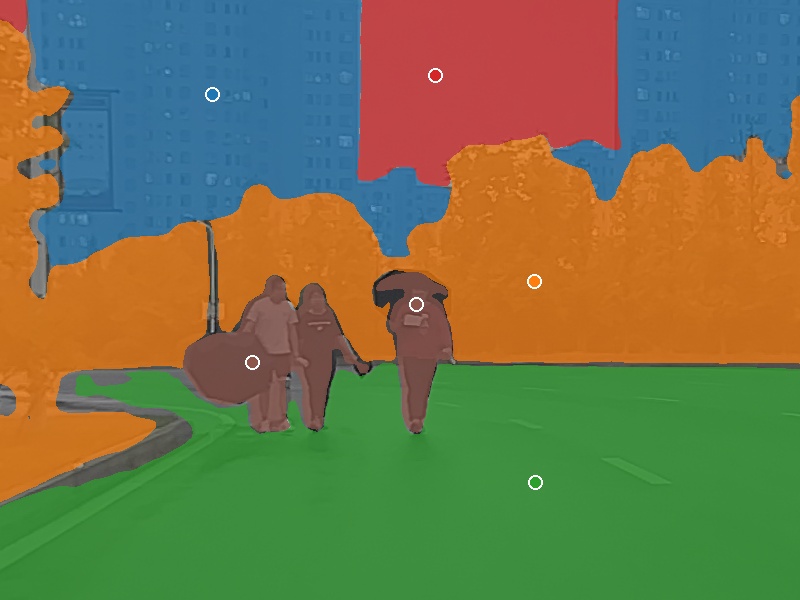} \\
\end{tabular}
\caption{Qualitative results on the FMB dataset \cite{liu2023segmif}. The model that generated the results is based DINOv2-B14. This dataset offers infrared images as an additional modality. The first and third image in each row are the RGB image with the ground truth and the prediction plotted onto the image, respectively. }

\label{fig:fmb_qualitative_app}
\end{figure*}

\begin{figure*}
\centering
\newcommand{\threemodwidth}{0.3} 
\begin{tabular}{ccc}
\textbf{Ground Truth} & \textbf{Thermal} & \textbf{Prediction} \\
\includegraphics[width=\threemodwidth\linewidth]{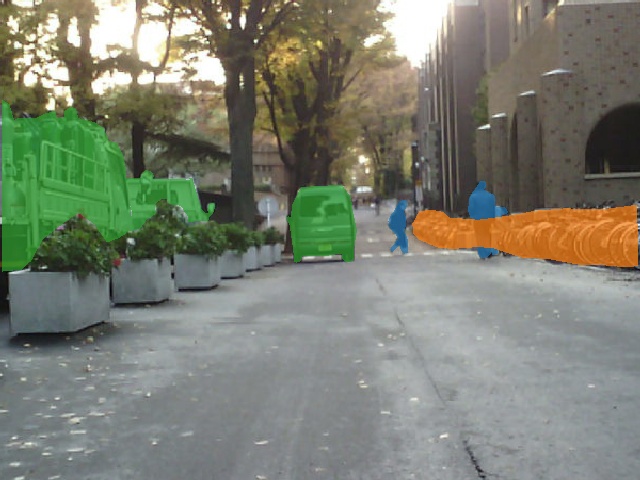} & 
\includegraphics[width=\threemodwidth\linewidth]{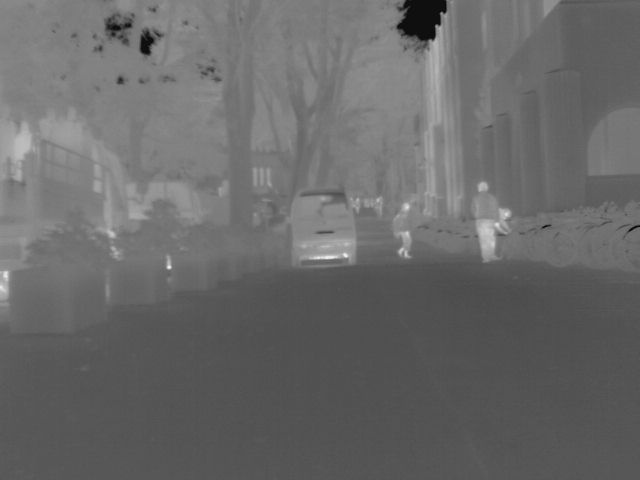} & 
\includegraphics[width=\threemodwidth\linewidth]{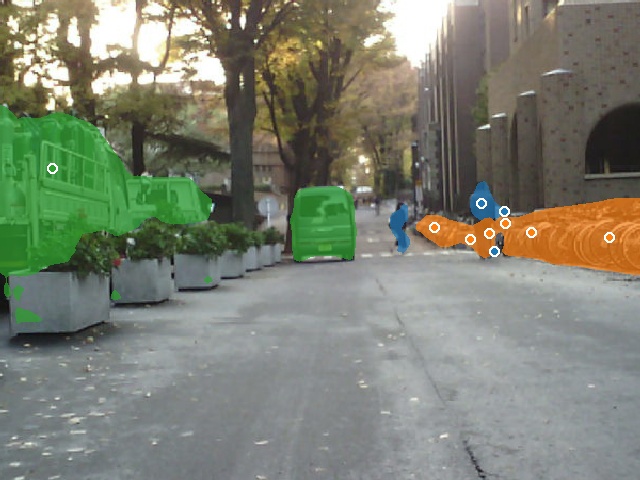} \\
\includegraphics[width=\threemodwidth\linewidth]{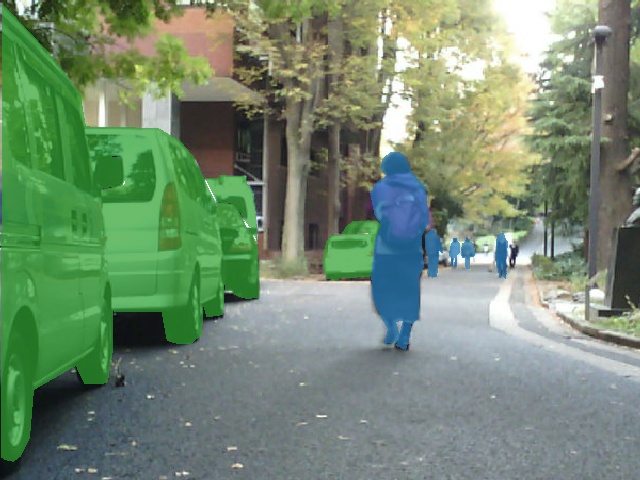} & 
\includegraphics[width=\threemodwidth\linewidth]{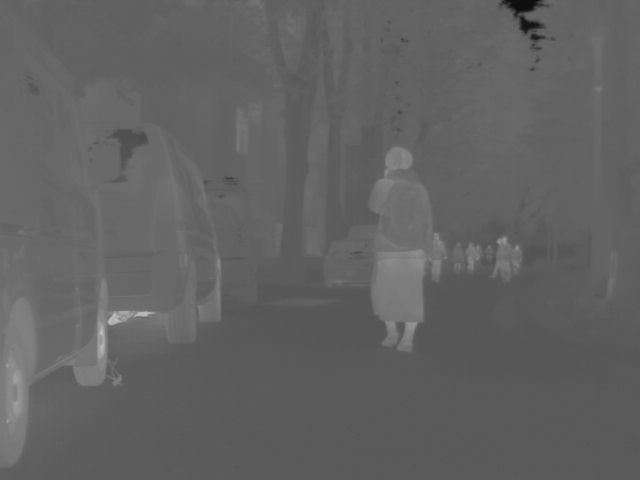} & 
\includegraphics[width=\threemodwidth\linewidth]{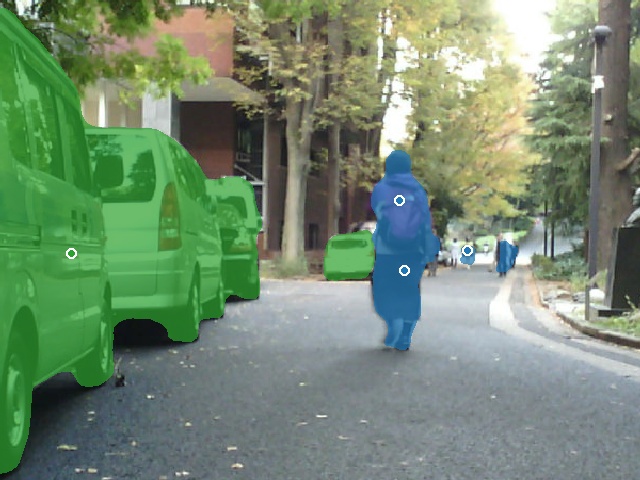} \\
\end{tabular}
\caption{Qualitative results on the MFNet dataset \cite{ha2017mfnet}. The model that generated the results is based DINOv2-B14. This dataset offers thermal images as an additional modality. The first and third image in each row are the RGB image with the ground truth and the prediction plotted onto the image, respectively. }

\label{fig:mfnet_qualitative_app}
\end{figure*}

We also display some qualitative results for DeLiVER \cite{zhang2023delivering} in Figure \ref{fig:deliver_qualitative_app}, MCubeS \cite{Liang_2022_CVPR} in Figure \ref{fig:mcubes_qualitative_app}, the FMB dataset \cite{liu2023segmif} in Figure \ref{fig:fmb_qualitative_app}, and MFNet \cite{ha2017mfnet} in Figure \ref{fig:mfnet_qualitative_app}. 
In all cases, the first column of each figure always shows the RGB input image with the ground truth plotted onto the image. The last column always shows the RGB image with the prediction and the positive clicks for each surface plotted onto it. We refrain from plotting the negative clicks to avoid clutter. 

\begin{figure*}
    \centering
    \newcommand{\threemodwidth}{0.3} 

    \begin{tabular}{ccc}
    \textbf{Ground Truth} & \textbf{Before} & \textbf{After} \\
    \includegraphics[width=\threemodwidth\linewidth]{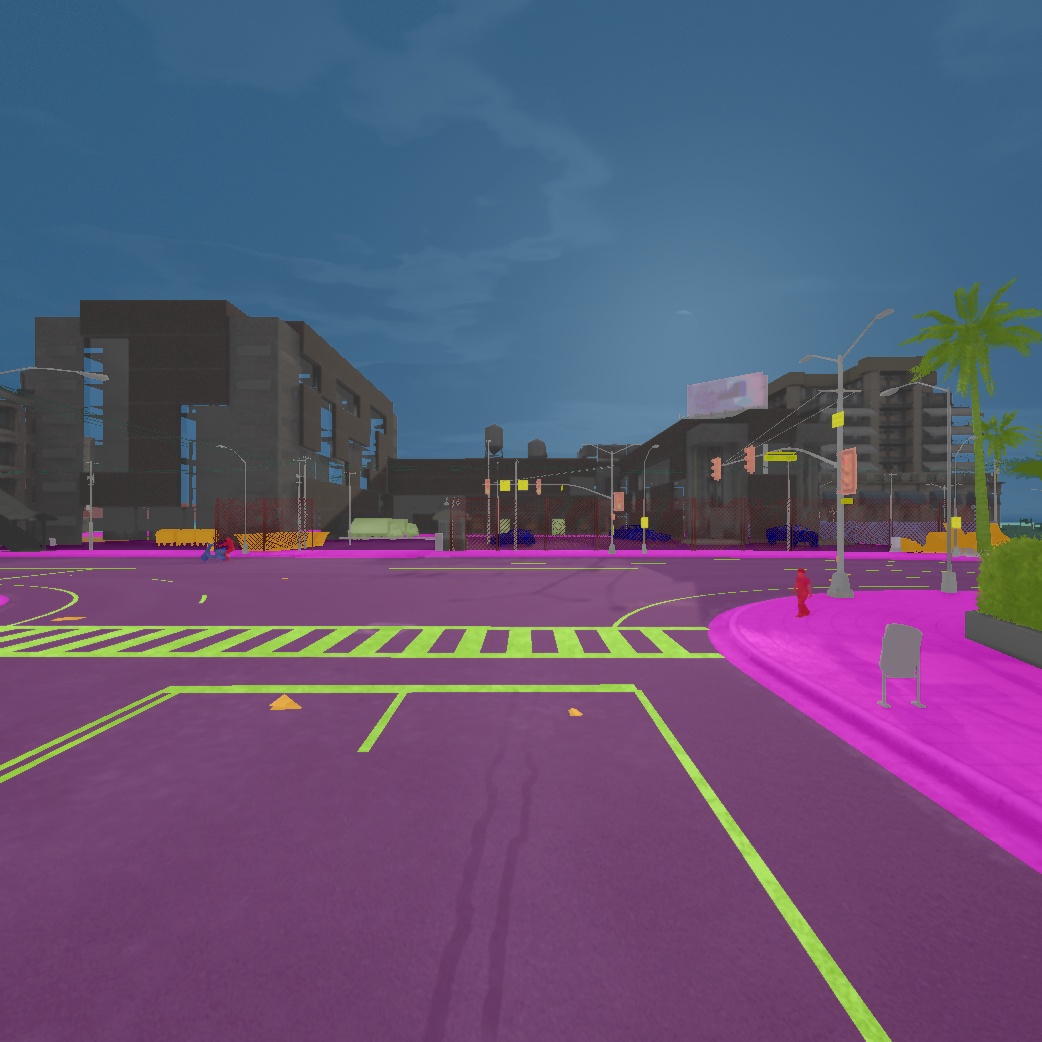} & 
    \includegraphics[width=\threemodwidth\linewidth]{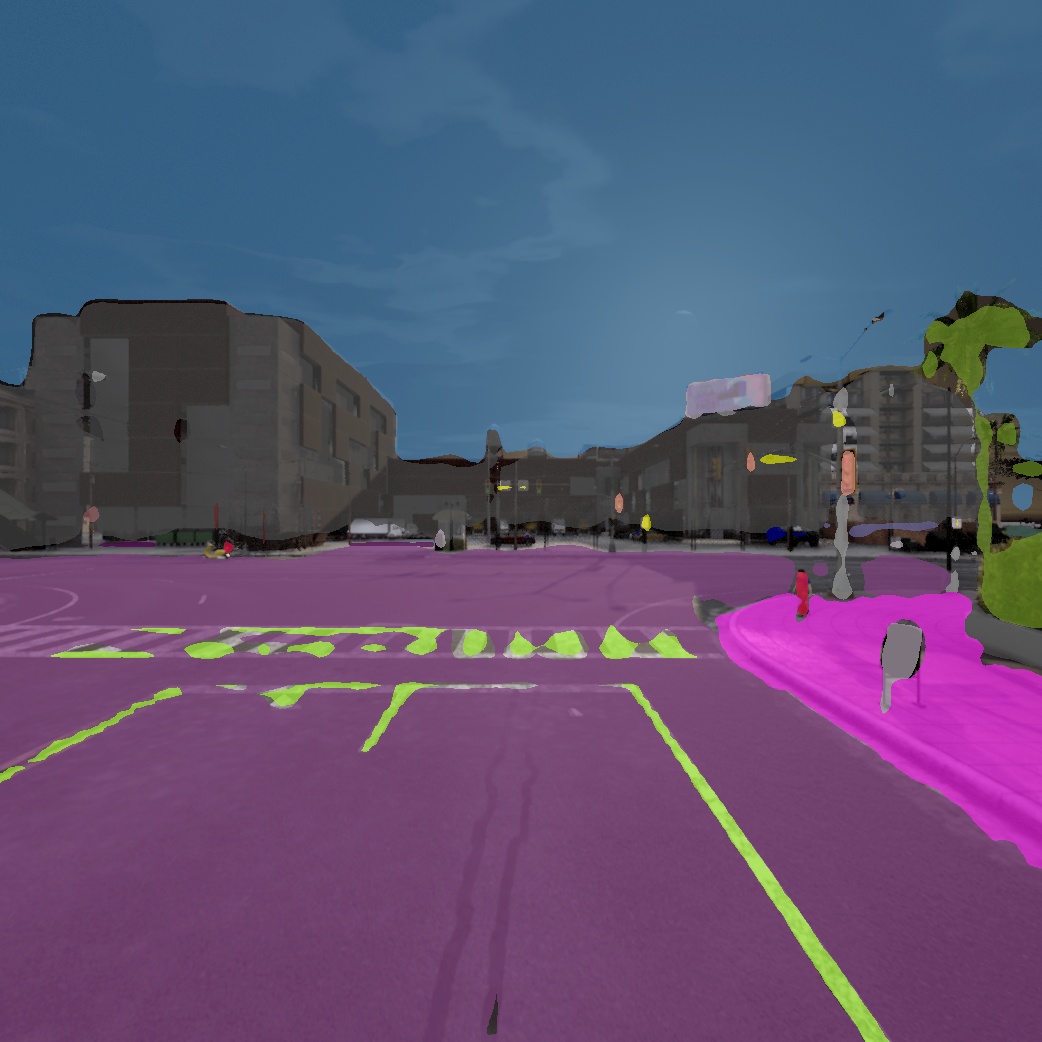} & 
    \includegraphics[width=\threemodwidth\linewidth]{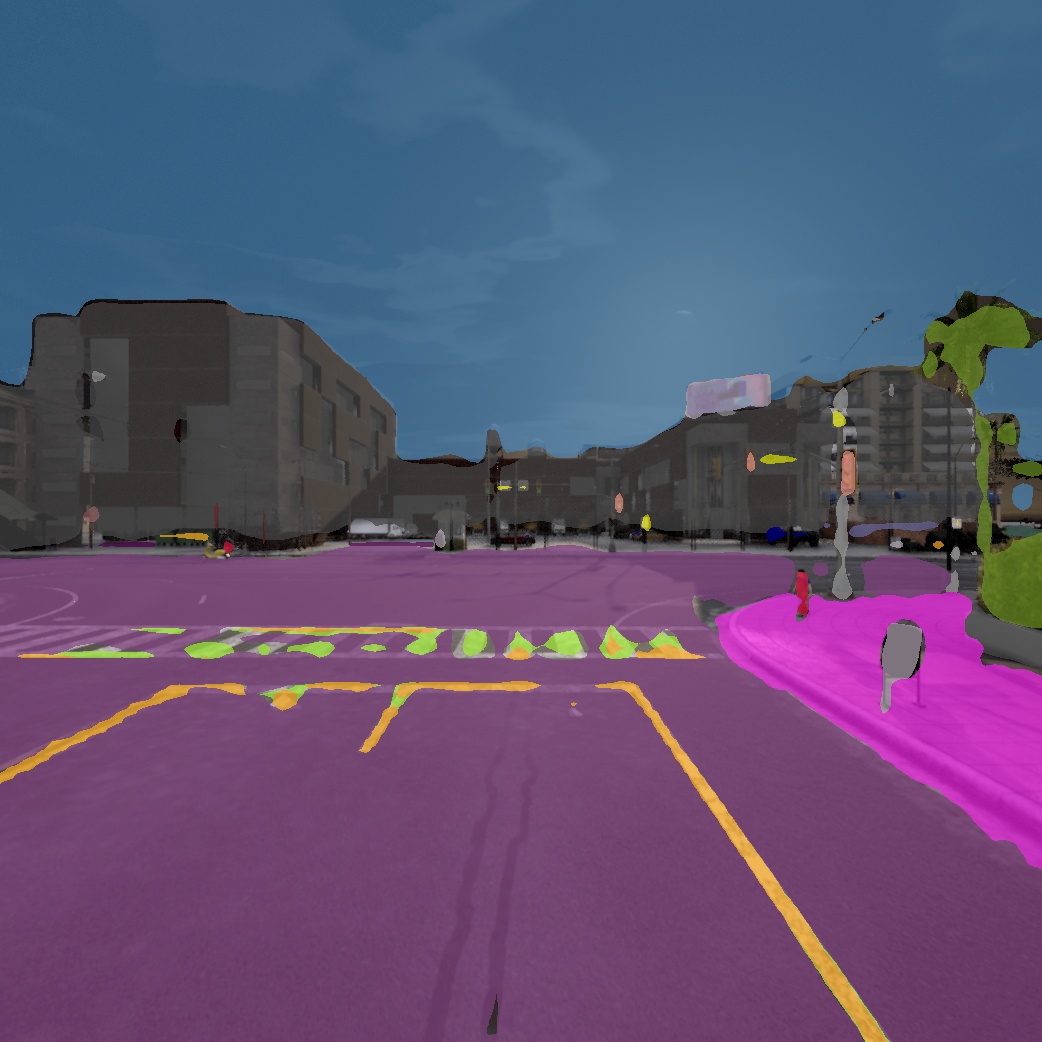} \\
    \includegraphics[width=\threemodwidth\linewidth]{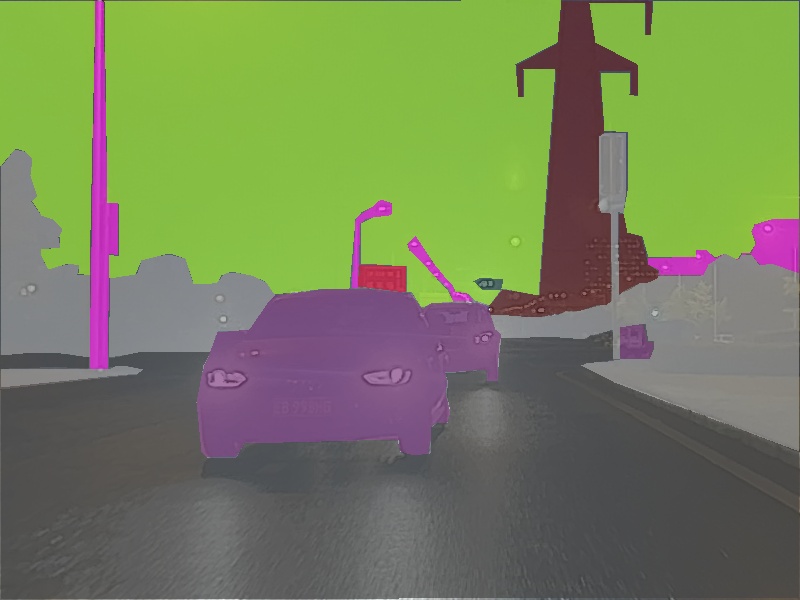} & 
    \includegraphics[width=\threemodwidth\linewidth]{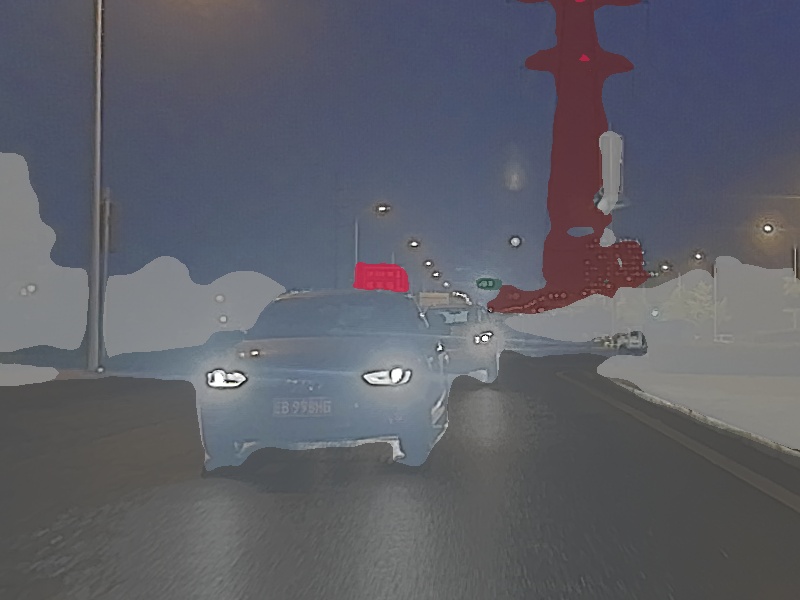} & 
    \includegraphics[width=\threemodwidth\linewidth]{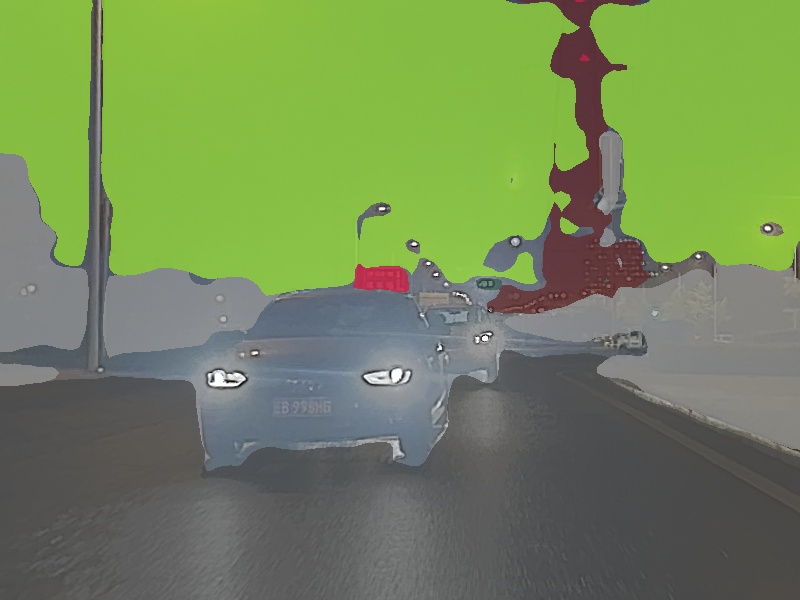} \\
    \includegraphics[width=\threemodwidth\linewidth]{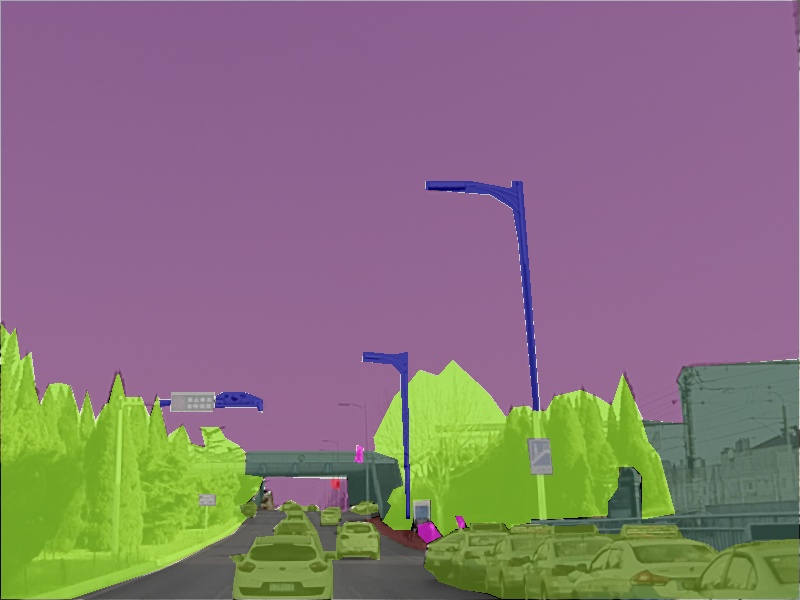} & 
    \includegraphics[width=\threemodwidth\linewidth]{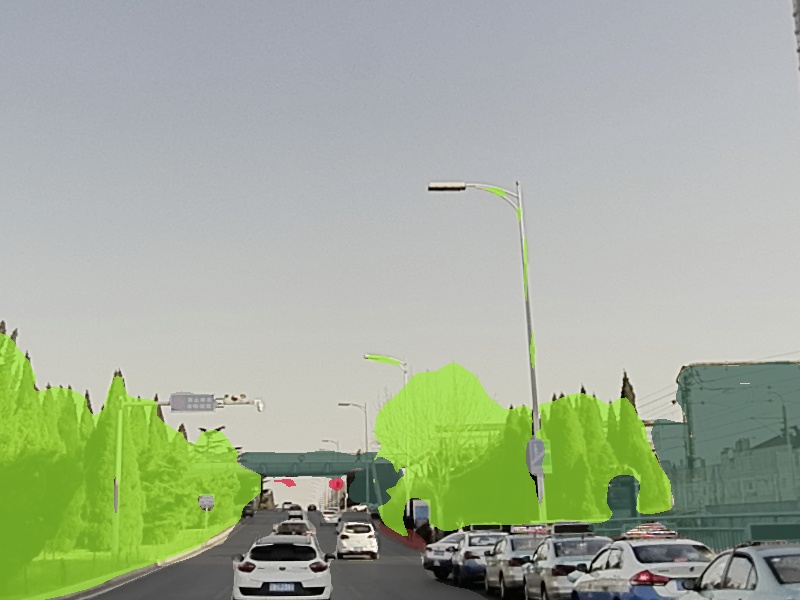} & 
    \includegraphics[width=\threemodwidth\linewidth]{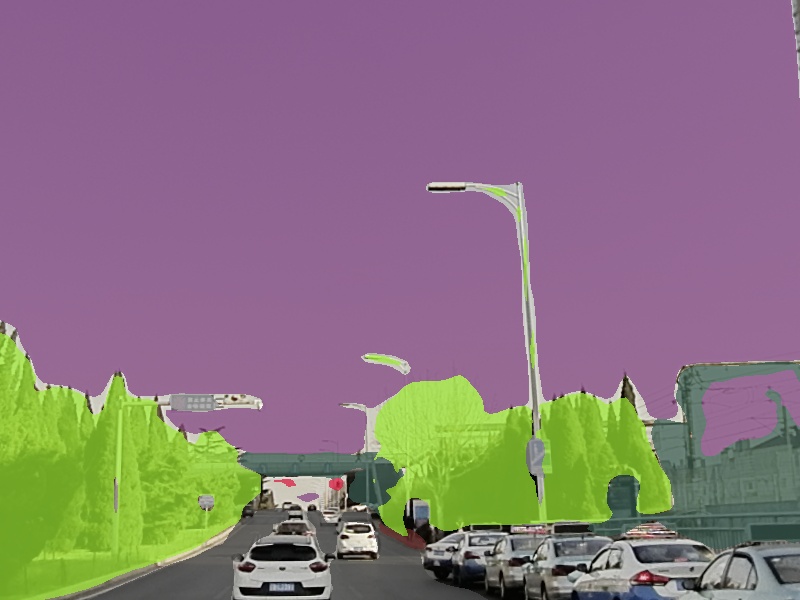} \\
    \end{tabular}

    \caption{Examples for conflicts between masks which are caused by false positives in the most recently edited mask. The first row contains examples from DeLiVER \cite{zhang2023delivering}, while the second and third row are from FMB \cite{liu2023segmif}. The left column displays the ground truth. The middle column displays the joint mask before the overlapping mask has been inserted, and the right column after the overlapping mask has been inserted}
    \label{fig:mask_conflicts}
\end{figure*}

For each of the figures, the model used to generate the prediction is based on DINOv2-B14 as the RGB foundation backbone. We always choose the version with the maximum number of available modalities. Figure \ref{fig:mask_conflicts} displays examples for conflicts between overlapping masks.

\end{document}